\documentclass{article}

\PassOptionsToPackage{compress}{natbib}
\PassOptionsToPackage{dvipsnames,table}{xcolor}

\usepackage[final]{neurips_2024}

\usepackage[utf8]{inputenc} 
\usepackage[T1]{fontenc}    
\usepackage{hyperref}       
\usepackage{url}            
\usepackage{booktabs}       
\usepackage{amsfonts}       
\usepackage{nicefrac}       
\usepackage{microtype}      
\usepackage{graphicx}
\usepackage{listings}
\usepackage{multirow}
\usepackage{pifont}
\usepackage{amsmath}
\usepackage{subcaption}
\usepackage{caption}
\usepackage{float}
\usepackage{makecell}
\usepackage{multicol}
\usepackage{amsthm}        
\theoremstyle{definition}
\usepackage{amssymb}  
\usepackage{enumitem}
\usepackage{listings}
\usepackage{array,pifont,booktabs}
\usepackage{csquotes}
\usepackage{tabularx}
\usepackage{placeins}
\usepackage{siunitx}
\usepackage{tcolorbox}
\usepackage{tikz}
\usepackage[dvipsnames]{xcolor}
\usetikzlibrary{positioning, arrows, calc, patterns}
\usepackage{subcaption} 

\usepackage{tikz}
\usetikzlibrary{positioning,arrows,shapes}

\usepackage{pgfplots}
\usepackage{tikz}
\pgfplotsset{compat=1.18}

\definecolor{llamacolor}{RGB}{231, 76, 60}    
\definecolor{qwencolor}{RGB}{52, 152, 219}    
\definecolor{deepseekcolor}{RGB}{46, 204, 113} 

\usepackage{pgfplots}
\pgfplotsset{compat=1.17}

\newtheorem{definition}{Definition}

\title{\EWEB: 24T tokens of organized web data}

\newcommand{\EWEB}{\textsc{Essential-Web v1.0}}
\newcommand{\ETAX}{\textsc{EAI-Taxonomy}}
\newcommand{\EMODELBOLD}{\textbf{EAI-Distill-0.5b}}
\newcommand{\EMODEL}{\texttt{EAI-Distill-0.5b}}
\newcommand{\oneref}[1]{%
  \hyperref[#1]{\nameref*{#1} (Section~\ref*{#1}, p.~\pageref*{#1})}%
}

\newcommand{\cmark}{\ding{51}} 
\newcommand{\xmark}{\ding{55}} 
\newcolumntype{C}[1]{>{\centering\arraybackslash}m{#1}} 

\author{%
  Essential AI \\
  San Francisco, CA \\
  \href{mailto:research@essential.ai}{\texttt{research@essential.ai}} \\
}

\begin{document}
\renewcommand{\lstlistingname}{Algorithm}

\maketitle

\begin{abstract}

Data plays the most prominent role in how language models acquire skills and knowledge. The lack of massive, well-organized pre-training datasets results in costly and inaccessible data pipelines. We present \EWEB, a 24-trillion-token dataset in which every document is annotated with a twelve-category taxonomy covering topic, format, content complexity, and quality. Taxonomy labels are produced by \EMODEL, a fine-tuned 0.5b-parameter model that achieves an annotator agreement within \(3\%\) of \texttt{Qwen2.5-32B-Instruct}. With nothing more than SQL‑style filters, we obtain competitive web-curated datasets in math (-8.0\% relative to SOTA), web code (+14.3\%), STEM (+24.5\%) and medical (+8.6\%). \EWEB \ is available on HuggingFace: \href{https://huggingface.co/datasets/EssentialAI/essential-web-v1.0}{EssentialAI/essential-web-v1.0}. 
\end{abstract}

\vspace{4\baselineskip}

\begin{figure}[h]
\centering
\resizebox{0.98\columnwidth}{!}{%
\begin{tikzpicture}[
    box/.style={rectangle, rounded corners, draw=black, text centered, minimum width=2.5cm, minimum height=0.5cm, font=\footnotesize, align=center},
    blueBox/.style={box, fill=ForestGreen!15},
    arrow/.style={->, >=stealth},
    node distance=0.4cm and 0.8cm
]
\node[box] (dedupL) {Deduplicated, Filtered CC};
\node[left=0.1cm of dedupL, font=\scriptsize] {100TB};
\node[above=0.5cm of dedupL, text centered, align=center] (leftTitle) {\textbf{Training High-Recall Classifier}\\[0.05cm]{\footnotesize\mdseries curation timeline: weeks to months}};
\node[box, below=of dedupL] (trainMath) {train base math classifier};
\node[box, below=of trainMath] (runInf) {run inference};
\node[left=0.2cm of runInf, font=\scriptsize] {100TB};
\node[box, below=of runInf] (ingest) {manually inspect output};
\node[box, below=of ingest] (newData) {curate new training data};
\node[box, below=of newData] (retrain) {retrain};
\node[box, below=of retrain,fill=ForestGreen!15] (highRecall) {\(>\)100B high recall math};
\draw[arrow] (dedupL) -- (trainMath);
\draw[arrow] (trainMath) -- (runInf);
\draw[arrow] (runInf) -- (ingest);
\draw[arrow] (ingest) -- (newData);
\draw[arrow] (newData) -- (retrain);
\draw[arrow] (retrain) -- (highRecall);
\draw[arrow, red, dotted, thick] (retrain.west) to[bend left=60] (runInf.west);
\node[box, right=4cm of dedupL, fill=Fuchsia!20] (dedupR) {\EWEB\\ {\tiny\itshape >1B distinct document labels} };
\node[left=0.1cm of dedupR, font=\scriptsize] {100TB};
\node[text centered, align=center] at (dedupR |- leftTitle) {\textbf{\EWEB \ Approach}\\[0.05cm]{\footnotesize\mdseries curation timeline: hours to days}};
\node[box, below=0.8cm of dedupR, xshift=-2cm] (mathFilter) {\texttt{subject == math}};
\node[box, below=0.8cm of dedupR, xshift=2cm] (arbitraryFilter) {arbitrary subject};
\node[box, below= of mathFilter, fill=ForestGreen!15] (eaiMath) {\(>\)100B high recall math};
\node[box, below= of arbitraryFilter, fill=ForestGreen!15] (eaiArbitrary) {high recall dataset};
\draw[] (dedupR.south) -- ++(0,-0.3) coordinate (split);
\draw[arrow] (split) -| (mathFilter.north);
\draw[arrow] (split) -| (arbitraryFilter.north);
\draw[arrow] (mathFilter) -- (eaiMath);
\draw[arrow] (arbitraryFilter) -- (eaiArbitrary);
\end{tikzpicture}
}
\end{figure}

\newpage

{\tableofcontents}

\newpage

\section{Introduction}

Among innovations that can accelerate the path to more intelligent AI models, data innovations frequently outpace others, such as architectures or optimizers~\citep{kaplan2020scalinglawsneurallanguage, beck2024xlstmextendedlongshortterm, Bahri_2024, sorscher2023neuralscalinglawsbeating, shen2024scalinglawslinearcomplexity, liu2025muonscalablellmtraining}. A careful curation of the bytes consumed by large language models (LLMs) enables greater control over the skills they acquire. With pre-training datasets scaling to trillions of tokens, a detailed examination of their content can be intimidating. Moreover, open-weight models rarely disclose the composition of their datasets, and the broader ecosystem is trending toward reduced transparency as pre-training datasets continue to grow. This introduces challenges in reproducibility and auditing models -- challenges that will grow as models become more autonomous. An accessible and interpretable open-data ecosystem is therefore essential for training competitive models in the open. 

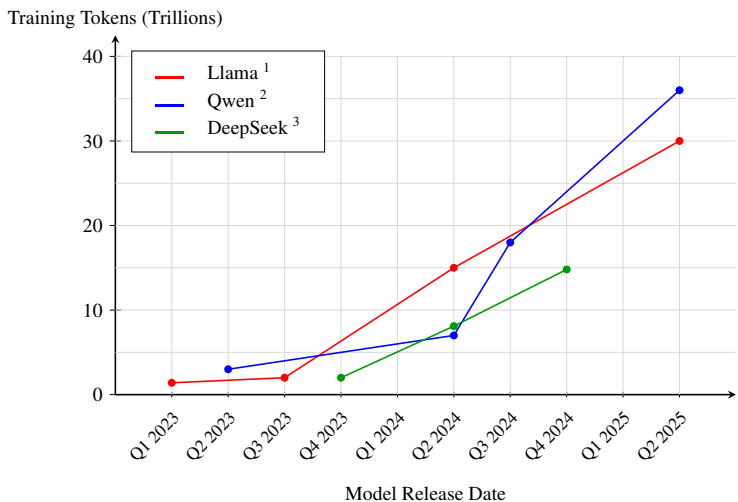
\begin{figure}[ht]
\centering
\scalebox{0.75}{ 
\begin{tikzpicture}[
    axis/.style={->, >=stealth, thick},
    label/.style={font=\normalsize},
    smol/.style={font=\footnotesize},
    title/.style={font=\normalsize},
    legend/.style={font=\normalsize}
]

\draw[axis] (0,0) -- (11,0);
\draw[axis] (0,0) -- (0,6.375)
      node[above, label] {Training Tokens (Trillions)};

\node[below, label] at (5.5,-1.5) {Model Release Date};

\foreach \y/\ylabel in {0/0, 1.5/10, 3/20, 4.5/30, 6/40} {
    \draw (-0.1,\y) -- (0.1,\y);
    \node[left, label] at (-0.1,\y) {\ylabel};
}

\foreach \x/\xlabel in {1/Q1 2023, 2/Q2 2023, 3/Q3 2023, 4/Q4 2023, 5/Q1 2024, 6/Q2 2024, 7/Q3 2024, 8/Q4 2024, 9/Q1 2025, 10/Q2 2025} {
    \draw (\x,-0.05) -- (\x,0.05);
    \node[below, smol, rotate=45, anchor=north east] at (\x,-0.1) {\xlabel};
}

\foreach \y in {0.75,1.5,2.25,3,3.75,4.5,5.25,6} {
    \draw[gray!30, thin] (0,\y) -- (11,\y);
}
\foreach \x in {1,2,3,4,5,6,7,8,9,10} {
    \draw[gray!30, thin] (\x,0) -- (\x,6);
}

\draw[thick, red] (1,0.21) -- (3,0.30) -- (6,2.25) -- (10,4.50);
\foreach \x/\y in {1/0.21, 3/0.30, 6/2.25, 10/4.50} {
    \fill[red] (\x,\y) circle (2pt);
}

\draw[thick, blue] (2,0.45) -- (6,1.05) -- (7,2.70) -- (10,5.40);
\foreach \x/\y in {2/0.45, 6/1.05, 7/2.70, 10/5.40} {
    \fill[blue] (\x,\y) circle (2pt);
}

\draw[thick, green!60!black] (4,0.30) -- (6,1.215) -- (8,2.22);
\foreach \x/\y in {4/0.30, 6/1.215, 8/2.22} {
    \fill[green!60!black] (\x,\y) circle (2pt);
}

\node[legend, draw, fill=white, inner sep=6pt] at (2,5.2) {
    \begin{tabular}{ll}
        \textcolor{red}{\rule{0.5cm}{1.5pt}} & Llama \footnotemark \\[2pt]
        \textcolor{blue}{\rule{0.5cm}{1.5pt}} & Qwen \footnotemark \\[2pt]
        \textcolor{green!60!black}{\rule{0.5cm}{1.5pt}} & DeepSeek \footnotemark \\
    \end{tabular}
};

\end{tikzpicture}
}
\caption{LLM pre-training dataset sizes over time.}
\end{figure}

\footnotetext{Llama dataset sizes sourced from \cite{touvron2023llamaopenefficientfoundation}, \cite{touvron2023llama2openfoundation}, \cite{grattafiori2024llama3herdmodels}, \cite{meta2025llama4}}
\footnotetext{Qwen dataset sizes sourced from \cite{bai2023qwentechnicalreport}, \cite{yang2024qwen2technicalreport}, \cite{qwen2025qwen25technicalreport}, \cite{yang2025qwen3technicalreport}} 
\footnotetext{DeepSeek dataset sizes sourced from \cite{deepseekai2024deepseekllmscalingopensource}, \cite{deepseekai2024deepseekv2strongeconomicalefficient}, and \cite{deepseekai2025deepseekv3technicalreport}}

Open-source pre-training datasets are divided into: (1) enormous, general-purpose datasets sorted by uninterpretable quality classifiers (2) smaller, domain-specific datasets curated with bespoke, complex pipelines. Both general and domain-specific datasets are unstructured, difficult to explore, and difficult to iteratively improve. Constructing these datasets requires significant computational resources given the scale and complexity of the data pipelines. Once the datasets are publicly released, their metadata rarely enables tuning beyond modifying classifier thresholds. 

We release \EWEB, a 24-trillion-token dataset with expressive and extensive metadata at a document-level. This metadata includes subject matter, web page type, content complexity, and document quality. Practitioners can now rapidly and inexpensively curate new datasets by writing SQL-like filters that utilize these metadata columns. Suppose a researcher wants to prepare a multi-billion-token chemistry corpus using publicly-available web data. Today, the researcher must first train a high‑recall chemistry classifier, a task hindered by scarce labeled data. Then, the classifier is run across hundreds of millions of documents to recall sufficient data. With \EWEB, a researcher can filter for chemistry, skip low-quality web pages (ads, product listings), and surface reasoning-dense documents — all with a query that takes under 15 minutes to write.

To construct \EWEB, we take advantage of powerful open-weight LLMs to synthetically label web documents with a 12-category taxonomy. We utilize these labels to train a more efficient classifier, \EMODEL, and run inference on 23.6B documents. Inference at this scale requires $\approx$90k AMD MI300x GPU-hours.\footnote{The inference job ran on 512 AMD MI300x for about 1 week.} We expect this one-off inference cost to be amortized as the community iterates on datasets and methods utilizing \EWEB.

To demonstrate the utility of \EWEB, we construct simple filters to curate high-performing datasets in math, web code, STEM, and medical domains. Our math dataset performs within 8.0\% of SOTA and our web code, STEM, and medical datasets outperform SOTA by 14.3\%, 24.5\%, 8.6\% respectively.

\section{Related Work}

Most web‑scale datasets rely on Common Crawl, an  18‑year crawl of more than 250 billion web pages \citep{commoncrawl}. Research on Common Crawl data can be naturally grouped into (1) heuristic filtering and de-duplication (2) monolithic, model-based filtering (3) domain-focused, model-based filtering, and (4) taxonomies for data curation.

\paragraph{Heuristic filtering and de-duplication.} C4 (160B tokens, 2019) pioneered heuristic filtering for Common Crawl. C4 proposes a processing pipeline to deduplicate and filter low quality text using document-level statistical heuristics \citep{raffel2023exploringlimitstransferlearning}. It was followed by an explosion of large-scale, open-source datasets and pipelines such as RefinedWeb (600B tokens), SlimPajama (600B tokens), Dolma-CC (3T tokens), and RedPajama (30T tokens)~\citep{penedo2023refinedwebdatasetfalconllm, weber2024redpajamaopendatasettraining, shen2024slimpajamadcunderstandingdatacombinations, soldaini2024dolmaopencorpustrillion}. 

\paragraph{Monolithic, model-based filtering.} In 2024, we see the introduction of monolithic model-based classifiers applied after well-tuned heuristic filtering and de-duplication. FineWeb‑Edu (1.3T tokens) filters documents with an education‑density classifier whose labels are bootstrapped from Llama‑3‑70B‑Instruct \citep{penedo2024finewebdatasetsdecantingweb}. Similarly, DCLM-baseline (3.6T tokens) filters documents using a fastText classifier trained to detect high instruction density  \citep{li2025datacomplmsearchgenerationtraining}. Nemotron‑CC ensembles several quality classifiers into one score before sampling Common Crawl \citep{su2025nemotroncctransformingcommoncrawl}. While classifier‑ranked datasets excel on broad language benchmarks, they lag on math and code evaluations. Their minimal metadata, often just the URL and a handful of quality scores, makes post‑hoc domain filtering with existing metadata almost impossible. Researchers therefore build domain‑specific processing pipelines to compensate.

\paragraph{Domain-focused, model-based filtering.}  Math, for instance, is rare (less than \(0.5\%\) of Common Crawl)\footnote{Calculated using our subject matter label, FDC, to filter for Mathematics.} yet beneficial for reasoning skills. OpenWebMath (12B tokens) builds a LaTeX detection pipeline, develops a custom HTML-to-text extractor to preserve mathematical formatting, and a math-detecting classifier \citep{paster2023openwebmathopendatasethighquality}. DeepSeek Math (120B tokens, not publicly released) iteratively trains a classifier to maximize recall of math documents resulting in a significantly larger dataset \citep{shao2024deepseekmathpushinglimitsmathematical}. MegaMath (264B web-based math tokens) builds multi-step classification, extraction, and processing pipelines for math and code \citep{zhou2025megamathpushinglimitsopen}. FineMath trains an initial classifier to recall math documents, re-extracts billions of pages with a math-specific extractor, and re-classifies these documents using a second, more powerful math classifier \citep{allal2025smollm2smolgoesbig}. Math is not the only domain that has been targeted: OpenCoder iteratively trains a classifier to detect web code and TheBlueScrubs-v1 trains a classifier to detect medical documents~\citep{huang2025opencoderopencookbooktoptier, felipe2025thebluescrubsv1comprehensivecuratedmedical}.

\paragraph{Taxonomies for curating training data.} \cite{wettig2025organizewebconstructingdomains} introduce taxonomies to structure web data. The authors develop a two-category taxonomy that measures topic and format, train a classifier using synthetic labels generated by Llama-3.1-Instruct models, label 200B tokens of Common Crawl, and demonstrate the utility of a combining model-based quality filters and a taxonomy. The authors introduce normalized mutual information to measure label and category redundancy. Using normalized mutual information, they show that embedding‑based clusters mostly overlap with topical labels, not format ones.

Unlike prior multi-trillion-token datasets, our work, \EWEB \ attaches a 12‑field taxonomy to every document, enabling both general and domain‑specific filtering through a single interface.

\subsection{Contributions}

We make four contributions:

\begin{enumerate}[leftmargin=*]
    \item \textbf{\EWEB} a deduplicated, 23.6B‑document (24T token) corpus drawn from Common Crawl and tagged with \ETAX, a 12‑field taxonomy, covering topic, web page type, content complexity, and quality.
    \item \textbf{Downstream validation.}  Simple SQL filters over the taxonomy produce datasets competitive with the best open-source, web-based baselines on math (-8.0\% relative to SOTA), code (+14.3\%), STEM (+24.5\%), and medical (+8.6\%). Our results demonstrate that taxonomy-curated datasets are competitive with complex, domain-specific pipelines with no domain-specific training.
    \item \textbf{Taxonomy evaluation toolkit.}  We introduce a metric suite of normalized mutual information (NMI) to gauge category independence \citep{wettig2025organizewebconstructingdomains}, annotator agreement using a variant of Cohen’s \(\kappa\) to ensure clear decision boundaries between category labels, and domain-recall to measure how well we recall high-value domains.
    \item \textbf{Efficient annotator model.}  We release \EMODEL \, a 0.5B‑parameter classifier produced by fine-tuning \texttt{Qwen2.5-0.5b-Instruct} using labels from \texttt{Qwen‑2.5‑32B‑Instruct} \citep{qwen2025qwen25technicalreport}. It labels the entire dataset in \(\approx\)90k MI300x GPU‑hours while remaining within 3\%, 14\%, 1\% of the teacher on annotator agreement, NMI, and domain-recall.
\end{enumerate}

\section{Taxonomy}
\label{sec:taxonomy}

\subsection{Formal Definition}
\label{sec:tax_definition}

\begin{definition}
\label{def:taxonomy}
A taxonomy is a finite set \(T=\{ C_{1},\dots,C_{k}\}\) of categories. Each category \(C_{i}\) has a non-empty, finite label set \(L_{i}\).

For a document \(d\), the taxonomic annotation is:
\[
T(d)=\bigl((\lambda_{1},\mu_{1}),\dots,(\lambda_{k},\mu_{k})\bigr),
\qquad
\lambda_{i}\in L_{i},\;
\mu_{i}\in\!\bigl(L_{i}\setminus\{\lambda_{i}\}\bigr)\cup\{\bot\},
\]
where \(\lambda_{i}\) is the primary label for category \(C_{i}\), \(\mu_{i}\) is an optional secondary label distinct from \(\lambda_{i}\) (useful when a document legitimately fits two labels), and \(\bot\) denotes abstention. All categories and label sets are fixed a priori, allowing us to train a single static classifier.
\end{definition}


\subsection{Desiderata} \label{sec:taxonomy-desiderata}
A useful metaphor for approaching taxonomy development is to think of the categories as axes in a high-dimensional grid space. Each document lives on a coordinate in that space, namely the category/label pairs assigned to it. We argue that a well-designed, web-scale taxonomy should satisfy five desiderata:

\begin{enumerate}
    \item \textbf{Orthogonality.} Every category should contribute information that is largely independent of the others — analogous to orthogonal axes in Euclidean space.
    \item \textbf{Expressivity.} The coordinate system must be sufficiently fine-grained that unions of category/label pairs can assemble a wide variety of targeted subsets.
    \item \textbf{Correctness.} Expressive, orthogonal axes are useless if labels are arbitrary. Annotators, human or LLM, must consistently assign the same labels to a document far more often than by chance. 
    \item \textbf{Efficiency.} Scaling to Common Crawl demands a fast and high-performing classifier.
    \item \textbf{Effectiveness.} Ultimately, a taxonomy matters only if it delivers in downstream performance. 
\end{enumerate}







\subsection{Development Methodology}

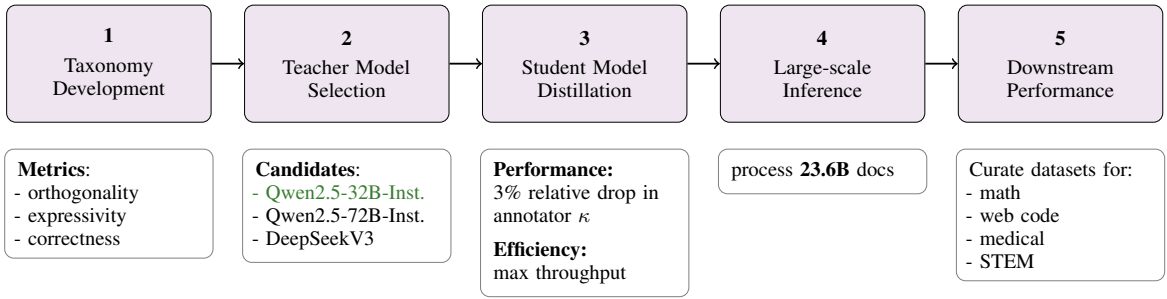
\begin{figure}[h]
\centering
\resizebox{0.98\textwidth}{!}{%
\begin{tikzpicture}[
    mainbox/.style={rectangle, rounded corners, draw=black, text centered, text width=3cm, minimum height=1.8cm, fill=Fuchsia!10, align=center, inner sep=2pt, font=\small},
    subbox/.style={rectangle, rounded corners, draw=gray, text width=2.8cm, minimum height=0.5cm, fill=white, align=left, font=\footnotesize, inner sep=.2cm},
    node distance=0.4cm
]

\node[mainbox] (taxonomy) at (0,0) {\footnotesize\textbf{1}\\[0.1cm] Taxonomy\\ Development};
\node[mainbox, right= 0.5cm of taxonomy] (parent_model) {\footnotesize\textbf{2}\\[0.1cm] Teacher Model Selection};
\node[mainbox, right=0.5cm of parent_model] (distill) {\footnotesize\textbf{3}\\[0.1cm] Student Model \\ Distillation};
\node[mainbox, right=0.5cm of distill] (inference) {\footnotesize\textbf{4}\\[0.1cm] Large-scale\\ Inference};
\node[mainbox, right=0.5cm of inference] (evaluate) {\footnotesize\textbf{5}\\[0.1cm] Downstream Performance};

\node[subbox, below=0.4cm of parent_model] (models) {\textbf{Candidates}:  \\ \textcolor{OliveGreen}{- Qwen2.5-32B-Inst.}\\ - Qwen2.5-72B-Inst. \\ - DeepSeekV3};
\node[subbox, below=0.4cm of taxonomy] (tax1) {\textbf{Metrics}: \\- orthogonality \\ - expressivity\\ - correctness};

\node[subbox, below=0.4cm of distill] (dist1) {\textbf{Performance:} \\ 3\% relative drop in annotator \(\kappa\) \\ \vspace{0.5\baselineskip}\textbf{Efficiency:} \\ max throughput};

\node[subbox, below=0.4cm of inference] (inf1) {process \textbf{23.6B} docs};

\node[subbox, below=0.4cm of evaluate] (eval1) {Curate datasets for: \\ - math \\ - web code \\ - medical \\ - STEM};

\draw[->, thick] (taxonomy.east) -- (parent_model.west);
\draw[->, thick] (parent_model.east) -- (distill.west);
\draw[->, thick] (distill.east) -- (inference.west);
\draw[->, thick] (inference.east) -- (evaluate.west);

\end{tikzpicture}%
}
\vspace{0.5\baselineskip}
\caption{Overview of our five-stage methodology for developing and deploying \ETAX.}
\label{fig:methodology}
\end{figure}

The development of \ETAX \ and \EMODEL \ (Figure~\ref{fig:methodology}) are discussed in the following sections:
\begin{itemize}
    \item \oneref{sec:downstream-results} compares datasets for math, web code, medical, and STEM curated using \ETAX \ against top-performing, open-source datasets in each domain. 
    \item \oneref{sec:taxonomy_metrics} defines the metrics used to evaluate orthogonality, correctness, and expressivity and introduces held-out evaluation sets.
    \item \oneref{sec:parent_model_selection} evaluates top performing open-source LLMs and motivates the selection of \texttt{Qwen2.5-32b-Instruct} as a teacher model. 
    \item \oneref{sec:running-at-scale} discusses the inference optimizations, training recipe, and performance evaluation of \EMODEL, the student model. 
\end{itemize}

\subsection{Selected Categories}\label{sec:taxonomy-overview}

Each web page receives labels for 12 categories across 5 logical groupings, which we refer to as \ETAX. In Table~\ref{tab:domains-categories}, we briefly introduce our categories and explain what each tries to capture, see Appendix \ref{sec:in-depth-taxonomy} for more details. Many of these categories are inspired by or directly taken from existing work in the open-source data curation community.\footnote{The Free Decimal Correspondance was designed by \cite{ockerbloom2010fdc}. The Bloom taxonomy was designed by \cite{bloom2001}. Document Type V2 was designed by \cite{wettig2025organizewebconstructingdomains}. Reasoning Depth and Technical Correctness were designed by \cite{yuan2025naturalreasoningreasoningwild28m}. Education Level was designed by \cite{penedo2024finewebdatasetsdecantingweb}.} \EWEB\ has 14.1M unique combinations of primary labels and 1.2B unique combinations of primary / secondary labels across all 23.6B docs.

\newcounter{dds}

\begingroup
\renewcommand{\arraystretch}{1.3}
\begin{table}[ht]
\centering
\begin{tabularx}{\textwidth}{@{}X X@{}}
\toprule
\textbf{Description} & \textbf{Categories} \\ \midrule
\textbf{FDC.} Free Decimal Correspondence is a public-domain analogue of the Dewey Decimal System.\footnotemark\setcounter{dds}{\value{footnote}} Labels subject matter hierarchically. &
\textbf{Level 1}: broad topic (ex: \texttt{5 - Science}) \newline
\textbf{Level 2}: fine topic (ex: \texttt{51 - Mathematics})\newline
\textbf{Level 3}: very fine topic (ex: \texttt{512 - Algebra})\\ \cmidrule{1-2}
\textbf{Bloom.} Educational-objective taxonomy. &
\textbf{Cognitive Process}: mental effort required \newline
\textbf{Knowledge Domain}: content abstraction \\ \cmidrule{1-2}
\textbf{Document Type.} Web page types. &
\textbf{V1}: broad types \newline
\textbf{V2}: fine types \\ \cmidrule{1-2}
\textbf{Content Quality.} Measures rigor and target audience. &
\textbf{Reasoning Depth}: thinking steps \newline
\textbf{Educational Level}: reader background  \newline
\textbf{Technical Correctness}: correctness \\ \cmidrule{1-2}
\textbf{Extraction.} Crawl artifacts and text extraction errors. &
\textbf{Extraction Artifacts}: HTML extraction errors \newline
\textbf{Missing Content}: from scrape or extraction \\ \bottomrule
\end{tabularx}
\vspace{0.5\baselineskip}
\caption{Taxonomy categories and their descriptions.}
\label{tab:domains-categories}
\end{table}
\endgroup

\footnotetext[\value{dds}]{"Dewey," "Dewey Decimal," "Dewey Decimal Classification," and "DDC" are trademarks of OCLC.}


\section{Downstream Results}
\label{sec:downstream-results}

We show that taxonomy‑based datasets with no domain‑specific training are competitive with bespoke pipelines built to target math, web code, medical and STEM data. 

\subsection{Experimental Protocol}
\label{sec:downstream-experimental-protocol}
In line with recent work such as DeepSeekMath, SmolLM2, and MegaMath, we anneal "pre-trained" models with domain-specific datasets to evaluate performance \citep{shao2024deepseekmathpushinglimitsmathematical, allal2025smollm2smolgoesbig, zhou2025megamathpushinglimitsopen}. We train two 2.3B parameters transformer models for 320B tokens, \(8 \times\) the Chinchilla compute-optimal ratio \citep{hoffmann2022chinchilla}.\footnote{When calculating Pareto, we ignore the size of the tied embedding/un-embedding matrix, which is 295M parameters.} To evaluate domain-specific datasets, we decay the learning rate of one of the base models to zero while training on 80B tokens of the new dataset. We also anneal on the original data mix to provide a baseline. The rationale behind starting with a base model is that it improves signal on difficult benchmarks such as MMLU and GSM8K, as opposed to training on each dataset from scratch for 80B tokens. After annealing, each model has seen 400B tokens, \(10\times\) Chinchilla Pareto.  

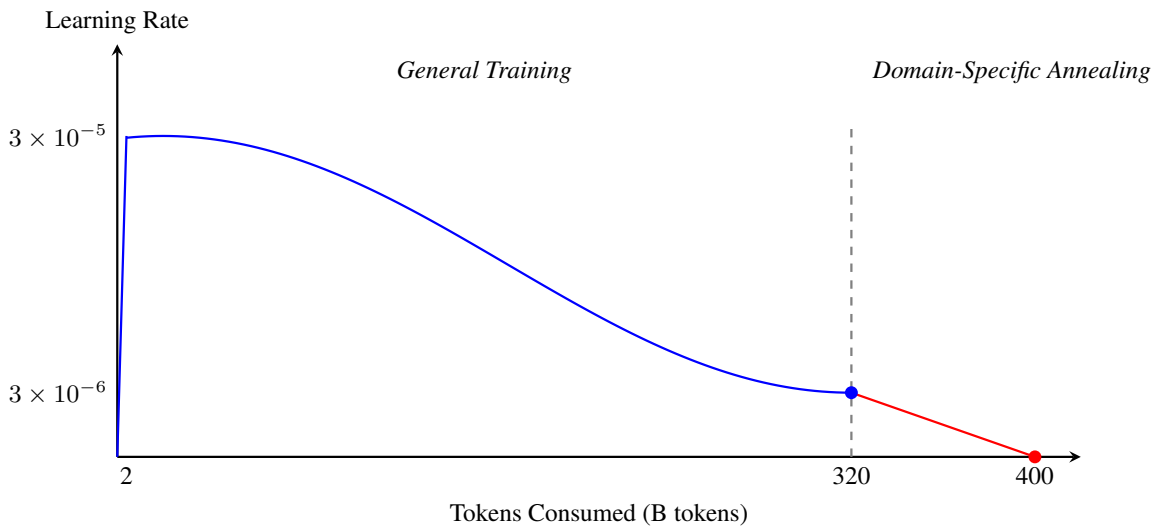
\begin{figure}[t]
\centering
\resizebox{0.98\textwidth}{!}{%
\begin{tikzpicture}[
    scale=1.15,
    axis/.style={->, >=stealth, thick},
    label/.style={font=\small},
    phase/.style={font=\small}
]
\draw[axis] (0,0) -- (10.5,0);
\draw[axis] (0,0) -- (0,4.5) node[above, label] {Learning Rate};

\node[below, label] at (5.25,-0.4) {Tokens Consumed (B tokens)};

\node[left, label] at (0,3.5) {$3 \times 10^{-5}$};
\node[left, label] at (0,0.7) {$3 \times 10^{-6}$};

\node[below, label] at (8,0) {320};
\node[below, label] at (10,0) {400};
\node[below, label] at (0.1,0) {2};

\draw[dashed, thick, gray] (8,0) -- (8,3.6);

\node[phase] at (4,4.2) {\textit{General Training}};
\node[phase] at (9.75,4.2) {\textit{Domain-Specific Annealing}};

\draw[thick, blue] (0,0) -- (0.1,3.5);

\draw[thick, blue, domain=0.1:8, samples=100] plot ({\x}, {0.7 + 1.4 * (1 + cos(((\x-0.5)/7.5)*180))});

\draw[thick, red] (8,0.7) -- (10,0);

\fill[blue] (8,0.7) circle (2pt);
\fill[red] (10,0) circle (2pt);

\end{tikzpicture}%
}
\caption{We train two base models: (1) \textbf{General Base} (100\% web data) and (2) \textbf{Code Base} (50\% web data, 50\% code data) for 320B tokens. We then evaluate all domain-specific \ETAX \ and top-performing, web-curated datasets by annealing one of the two base models for 80B tokens.}
\end{figure}

We vary the data composition of the two base models, while keeping all other training hyper-parameters fixed. The model architecture, configuration, and training hyper-parameters can be found in Appendix~\ref{sec:model_config_and_hparam}.
\begin{itemize}
    \item \textbf{General-Base}:  DCLM-baseline \citep{li2025datacomplmsearchgenerationtraining}
    \item \textbf{Code-Base}: 50\% DCLM-baseline + 50\% Python from Stack v2 Dedup \citep{lozhkov2024starcoder2stackv2} 
\end{itemize}

\paragraph{DCLM-baseline.} DCLM-baseline is a 3.6T token pre-training dataset based on Common Crawl. It is de-duplicated, heuristically-filtered, and labeled using a model-based classifier. For classification, the authors train a fastText classifier on instruction-formatted data from OpenHermes 2.5 and \texttt{r/ExplainLikeImFive} subreddit \citep{OpenHermes2.5}. DCLM-baseline is curated by selecting the top 10\% of documents after de-duplication and heuristic filters based on this classifier score \citep{li2025datacomplmsearchgenerationtraining}.\footnote{DCLM-baseline can be found on HuggingFace: \href{https://huggingface.co/datasets/mlfoundations/dclm-baseline-1.0}{mlfoundations/dclm-baseline-1.0}.} The HuggingFace dataset card notes the dataset is not intended for domains such as code and math. 

\paragraph{Stack v2.} The Stack v2 is a dataset of 104.2M github repositories collected from Software Heritage~\citep{software_heritage}. We use the Minhash LSH deduplicated version~\citep{lozhkov2024starcoder2stackv2}.\footnote{Stack v2 dedup can be found on HuggingFace: \href{https://huggingface.co/datasets/bigcode/the-stack-v2-dedup}{bigcode/the-stack-v2-dedup}.} We restrict training to the Python subsect because HumanEval\(^+\) and MBPP\(^+\) are Python only \citep{chen2021codex, austin2021programsynthesislargelanguage, liu2023codegeneratedchatgptreally}. 

Table~\ref{tab:eai-datasets-table} and Table~\ref{tab:public-datasets-table} in the Appendix contain HuggingFace links for our datasets and public datasets used in this section, respectively.

\paragraph{Decontamination.} All datasets were decontaminated with a 13-gram Bloom filter that normalizes text and punctuation before white-space tokenization. The following evals were decontaminated against: GSM8K, MATH, HumanEval, MBPP, MMLU, CareQA, MedMCQA, MedQA-USMLE, and PubMedQA.

\paragraph{Evaluation framework.} We run evaluations using \texttt{lm-eval-harness} \citep{eval-harness}. See Appendix~\ref{sec:detailed_evals} for details.

\subsection{Math}

There has been extensive work in the open source community to develop high-quality math datasets \citep{paster2023openwebmathopendatasethighquality, allal2025smollm2smolgoesbig, zhou2025megamathpushinglimitsopen, shao2024deepseekmathpushinglimitsmathematical}. We compare the performance of taxonomy-based math datasets with all top performing, public math datasets that are curated from Common Crawl. We only select math datasets that are filtered web data and do not report performance of synthetic math datasets or datasets where an LLM was used to rewrite documents.\footnote{We only report filtered web data with no document-level LLM changes to the isolate performance of different data filtering methods. These results focus on the quality of the natural web data selected by different filtering methods.} A list of datasets tested and the domain-specific steps used during preparation can be found in Table~\ref{tab:math_datasets}.\footnote{We acknowledge that the total dataset preparation cost of the taxonomy is much higher than any of these individual datasets because we run a 500M parameter transformer to classify 23.6B documents. However, we only focus on domain-specific curation costs in Table~\ref{tab:math_datasets}.} We use \textbf{General-Base} for the annealing experiments. To evaluate performance, we run 8-shot GSM8K, 4-shot MATH, and 5-shot MMLU-Math, which consists of all MMLU subtasks related to math \citep{hendrycks2021measuringmassivemultitasklanguage}.\footnote{MMLU-Math: MMLU Abstract Algebra, MMLU College Mathematics, MMLU Elementary Mathematics, MMLU High School Mathematics, MMLU High School Statistics}

\begin{table}[htbp]
\small
\centering
\setlength{\tabcolsep}{6pt}            
\begin{tabular}{l c c c p{4.25cm}}      
\toprule
\textbf{Dataset} & \textbf{Size (B tok)} & \textbf{Math extract} & \textbf{Math classifier} & \textbf{Domain–specific curation notes} \\
\midrule
FineMath 3+ & 32 & \cmark & \cmark & 118M math classifer on 7.1B docs \\
OpenWebMath & 12 & \cmark & \cmark & math fastText  \\
MegaMath Web (Top 10\%)\footnotemark & 30 & \cmark & \cmark & iteratively trained math fastText \\
\ETAX \ Top Math & 29 & \xmark & \xmark & \ETAX \ filter \\
\ETAX \ Math w/ FM & 34 & \xmark & \cmark & \ETAX \ filter; \newline FineMath classifier on 116M docs \\
\bottomrule
\end{tabular}
\vspace{0.5\baselineskip}
\caption{Overview of top-performing, open-source math datasets and taxonomy-based math datasets along with brief summary of effort to curate. "Math extract" denotes a domain‐specific HTML-to-plaintext extraction was used. "Math classifier" indicates a domain-specific classifier was used.}
\label{tab:math_datasets}
\end{table}

\footnotetext[13]{We prepared MegaMath Web (Top 10\%) by selecting the top 10\% of documents based on the \texttt{math\_score} column.}

\subsubsection{Taxonomy-Based Math Datasets}

We prepare two taxonony-based math datasets. \ETAX \ Top Math (29B tokens) targets high-quality math documents that exhibit reasoning and are technically correct, the full filter can be found in Algorithm~\ref{lst:taxonomy-top-math-filter} in Appendix~\ref{sec:taxonomy_filters_appendix}. \ETAX \ Math w/ FM (34B tokens) filters for any document labeled as \texttt{51 - Mathematics}. We then label all 116M recalled documents with the FineMath Classifier\footnote{A 118M parameter transformer-based classifier trained on synthetic labels generated by \texttt{LLama3-70B-Instruct}. More details can be found on HuggingFace:  \href{https://huggingface.co/HuggingFaceTB/finemath-classifier}{HuggingFaceTB/finemath-classifier}.} and filter for the top 34B tokens so that the size is comparable to FineMath 3+ (Algorithm~\ref{lst:taxonomy-math-w-fm-filter}). 

\subsubsection{Downstream Math Results}

\begin{table}[h]
\centering
\begin{tabular}{lccc}
\toprule
\textbf{Dataset} & \textbf{GSM8K} & \textbf{MATH} & \textbf{MMLU–Math} \\
\midrule
FineMath 3+              & \textbf{26.4\%}$\scriptstyle\pm1.4$ & \textbf{11.7\%}$\scriptstyle\pm0.4$ & \textbf{32.3\%}$\scriptstyle\pm1.5$ \\
OpenWebMath              & 14.6\%$\scriptstyle\pm1.1$          & 9.3\%$\scriptstyle\pm0.4$           & 29.9\%$\scriptstyle\pm1.5$ \\
MegaMath Web (Top 10\%)  & 9.8\%$\scriptstyle\pm0.9$           & 7.9\%$\scriptstyle\pm0.3$           & 29.9\%$\scriptstyle\pm1.5$ \\
DCLM-baseline            & 4.8\%$\scriptstyle\pm0.7$           & 4.4\%$\scriptstyle\pm0.3$           & 27.0\%$\scriptstyle\pm1.4$ \\
\cmidrule{1-4}
\ETAX\ Top Math          & 21.3\%$\scriptstyle\pm1.3$          & 11.0\%$\scriptstyle\pm0.4$          & 30.5\%$\scriptstyle\pm1.5$ \\
\ETAX\ Math w/ FM        & 22.4\%$\scriptstyle\pm1.3$          & 11.5\%$\scriptstyle\pm0.4$          & 30.9\%$\scriptstyle\pm1.5$ \\
\bottomrule
\end{tabular}
\vspace{0.5\baselineskip}
\caption{Model performance (mean $\pm$ standard error) on GSM8K, Hendrycks Math, and MMLU-Math benchmarks. All datasets reported are filtered web data. In Appendix~\ref{sec:additional-math-results}, we include MegaMath Web Pro, which filters MegaMath Web using the FineMath classifier and rewrites the top documents using an LLM.}
\label{tab:downstream-math-results}
\end{table}

FineMath 3+ achieves the highest score in GSM8K, with \ETAX \ Top Math and \ETAX\ Math w/ FM datasets trailing by -4.0pp (15.2\%) and -5.1pp (19.3\%) respective. On MATH and MMLU-Math, \ETAX \ Math w/ FM performs within standard error of FineMath 3+. Other curated sets (MegaMath Web (Top 10\%), OpenWebMath) lag both FineMath and our taxonomy splits on GSM8K and MATH. The full results to the experiments can be found in Table~\ref{tab:downstream-math-results}. FineMath, MegaMath Web (Top 10\%), and OpenWebMath are domain-specific datasets with complex pipelines to maximize performance on mathematics benchmarks. The FineMath classifier targets \textquote{high school and early undergraduate levels} of mathematics, which directly caters to GSM8K and MATH. The performance of \ETAX \ Top Math comes from a simple semantic filter. \ETAX \ Math w/ FM takes advantage of the FineMath classifier by recalling a small, high-density subset of Common Crawl to reclassify. 

To understand the composition of FineMath 3+, we annotate it using the taxonomy. We find that only \(61.9\%\) of FineMath 3+ documents are labeled with FDC \texttt{51 - Mathematics}, as primary or secondary label. By inspecting the FDC categories of the remaining documents in FineMath 3+, we find other common subject matters in the dataset:  \texttt{53 - Physics}, \texttt{33 - Economics}, and \texttt{621 - Applied Physics}. 

\subsection{Code}
\label{sec:downstream-code-results}

Recent code LMs mix execution‑ready code with code tutorials, documentation, and API docs recalled from web data \citep{hui2024qwen25codertechnicalreport, deepseekai2024deepseekcoderv2breakingbarrierclosedsource, huang2025opencoderopencookbooktoptier}. We curate two web code datasets using the taxonomy and compare them to the only openly available, large-scale web code dataset: OpenCoder FineWeb Code Corpus (OpenCoder FW) \citep{huang2025opencoderopencookbooktoptier}. OpenCoder FW is a 49B token dataset curated with a fastText classifier iteratively trained to maximize math and code recall.\footnote{The classifier is trained with a strategy similar to DeepSeek Math \citep{shao2024deepseekmathpushinglimitsmathematical}. The authors use 500,000 LLM-scored code / math documents as an initial training set \citep{huang2025opencoderopencookbooktoptier}.} We anneal web code datasets (Table~\ref{tab:web_code_datasets}) from \textbf{Code-Base} to get signal on code knowledge and code generation evals. During annealing, we train on a 1:1 mixture of the dataset being evaluated and Python from the Stack-Edu \citep{allal2025smollm2smolgoesbig}. To measure code generation performance, we run 0-shot HumanEval\(^+\) and 3-shot MBPP+\(^+\).\footnote{HumanEval\(^+\) and MBPP+\(^+\) are variants of HumanEval and MBPP with more challenging test cases and corrected ground-truth solutions \citep{chen2021codex, austin2021programsynthesislargelanguage, liu2023codegeneratedchatgptreally}. When running 0-shot MBPP\(^+\), the model struggles to properly output the termination condition (\texttt{"[DONE]"}) in the default \texttt{lm-eval-harness} config given it isn't lack of explanation in the prompt. Therefore, we run with 3-shot to provide in-context examples of the termination condition.} To gauge code-related knowledge, we run 5-shot MMLU-CS, which consists of all MMLU subtasks related to Computer Science.\footnote{MMLU-CS: MMLU College Compute Science, MMLU High School Computer Science, MMLU Computer Security}

\begin{table}[htbp]
\centering
\small
\setlength{\tabcolsep}{6pt}            
\begin{tabular}{l c c p{6cm}}      
\toprule
\textbf{Dataset} & \textbf{Size (B tok)} & \textbf{Code classifier} & \textbf{Domain–specific curation notes} \\
\midrule
OpenCoder FW & 49 & \cmark & iteratively-trained math and web code fastText \\
\ETAX \ Top Code & 145 & \xmark & \ETAX \ filter \\
\ETAX \ Code w/ DCLM & 564 & \xmark & \ETAX \ filter; DCLM fastText\\
\bottomrule
\end{tabular}
\vspace{0.5\baselineskip}
\caption{Overview of existing SOTA open-source web code datasets and taxonomy-based web code datasets along with brief summary of effort to curate. "Code classifier" indicates a domain-specific classifier was used to detect code.}
\label{tab:web_code_datasets}
\end{table}

\subsubsection{Taxonomy-Based Web Code Datasets}

We prepare \ETAX \ Top Code (145B tokens) to target high-quality code documentation that is technically correct and exhibits intermediate to advanced reasoning (Algorithm~\ref{lst:taxonomy-code-filter}). In addition, we construct \ETAX \ Code w/  DCLM (564B tokens), which combines taxonomy filters targeting code documentation and the DCLM classifier, at the same threshold as DCLM-baseline, to filter for instruction-dense documents (Algorithm~\ref{lst:taxonomy-code-dclm-filter}). We add \texttt{51 - Mathematics} to \ETAX \ Code w/ DCLM because OpenCoder FW targets both web code and mathematics. 

\subsubsection{Downstream Code Results}

\begin{table}[ht]
\centering
\setlength{\tabcolsep}{5pt}
\begin{tabular}{lccc}
\toprule
\textbf{Web-Code Dataset} & \textbf{HumanEval$^{+}$} & \textbf{MBPP$^{+}$} & \textbf{MMLU–CS} \\
\midrule
DCLM-baseline          & 28.0\%{\footnotesize$\pm$3.5} & 45.5\%{\footnotesize$\pm$2.6} & 32.0\%{\footnotesize$\pm$2.7} \\
OpenCoder FW           & 26.2\%{\footnotesize$\pm$3.4} & 45.8\%{\footnotesize$\pm$2.6} & 27.7\%{\footnotesize$\pm$2.6} \\
\cmidrule{1-4}
\ETAX \ Top Code          & 27.4\%{\footnotesize$\pm$3.5} & \textbf{46.6\%}{\footnotesize$\pm$2.6} & 29.0\%{\footnotesize$\pm$2.6} \\
\ETAX \ Code w/ DCLM   & \textbf{28.7\%}{\footnotesize$\pm$3.5} & 45.0\%{\footnotesize$\pm$2.6} & \textbf{47.0\%}{\footnotesize$\pm$2.9} \\
\bottomrule
\end{tabular}
\vspace{0.5\baselineskip}
\caption{Pass@1 accuracy (mean {$\pm$} standard error) on 0-shot HumanEval\(^+\), 3-shot MBPP\(^+\), and accuracy on the MMLU computer-science subset.}
\label{tab:code-eval-results}
\end{table}

Across the code-generation evaluations, all datasets perform within standard error of each other (Table~\ref{tab:code-eval-results}). We see an absolute +15.0pp (46.8\%) improvement in the MMLU-CS score from DCLM-baseline to \ETAX \ Code w/ DCLM. Both DCLM-baseline and taxonomy-based code datasets outperform OpenCoder FW on MMLU-CS. However, there is a clear impact on general code knowledge when using a taxonomy-curated web code dataset.

\subsection{Medical}

TheBlueScrubs‑v1 (24B tokens) is the only public, curated Common Crawl medical corpus. The authors train a logistic regression classifier to detect medical documents and use it to filter SlimPajama, a 627B web dataset \citep{felipe2025thebluescrubsv1comprehensivecuratedmedical}. We compare TheBlueScrubs-v1 against DCLM-baseline and taxonomy-based medical datasets by annealing \textbf{General-Base}. We evaluate medical performance using 3-shot PubMedQA, 5-shot CareQA-en, 5-shot MedMCQA, 5-shot MedQA-USMLE-4-options, and 5-shot MMLU-Med \citep{jin-etal-2019-pubmedqa, ariasduart2025automaticevaluationhealthcarellms, jin2020diseasedoespatienthave, medmcqa}.\footnote{MMLU-Med: MMLU Anatomy, MMLU Clinical Knowledge, MMLU College Biology, MMLU College Medicine, MMLU Medical Genetics, MMLU Professional Medicine. These categories were grouped for medical evaluation by \cite{singhal2022largelanguagemodelsencode}}

\begin{table}[htbp]
\centering
\small
\setlength{\tabcolsep}{6pt}            
\begin{tabular}{l c c p{5.5cm}}      
\toprule
\textbf{Dataset} & \textbf{Size (B tok)} & \textbf{Medical classifier} & \textbf{Domain–specific curation notes} \\
\midrule
TheBlueScrubs-v1 & 24 & \cmark & logistic regression to detect medical  \\
\ETAX \ Med & 433 & \xmark & \ETAX \ filter \\
\ETAX \ Med w/ DCLM & 205 & \xmark &  \ETAX \ filter; DCLM fastText\\
\bottomrule
\end{tabular}
\vspace{0.5\baselineskip}
\caption{Overview of existing open-source medical datasets and taxonomy-based medical datasets along with brief summary of effort to curate. "Medical classifier" indicates a domain-specific classifier was used to detect medical.}
\label{tab:medical_datasets}
\end{table}

\subsubsection{Taxonomy-Based Medical Dataset}

We prepare the \ETAX \ Med (433B tokens) to target scientific medical documents that exhibit reasoning and are technically correct (Algorithm~\ref{lst:taxonomy-medical-filter}). We then apply the DCLM classifier, at the same threshold as DCLM-baseline, to the \ETAX \ Med to construct \ETAX \ Med w/ DCLM (205B tokens, Algorithm~\ref{lst:taxonomy-med-w-dclmfilter}). 

\subsubsection{Downstream Medical Results}

\begin{table}[ht]
\centering
\small
\setlength{\tabcolsep}{5pt}
\begin{tabular}{lccccc}
\toprule
\textbf{Model} & \textbf{CareQA-en} & \textbf{MedMCQA} & \textbf{MedQA-USMLE} & \textbf{PubMedQA} & \textbf{MMLU–Med} \\
\midrule
DCLM-baseline      & 26.9\%{\footnotesize$\pm$0.6} & 31.6\%{\footnotesize$\pm$0.7} & 25.9\%{\footnotesize$\pm$1.2} & \textbf{70.6\%}{\footnotesize$\pm$2.0} & 31.0\%{\footnotesize$\pm$1.5} \\
TheBlueScrubs-v1   & 25.1\%{\footnotesize$\pm$0.6} & 32.2\%{\footnotesize$\pm$0.7} & 25.3\%{\footnotesize$\pm$1.2} & 69.2\%{\footnotesize$\pm$2.1} & 25.7\%{\footnotesize$\pm$1.4} \\
\cmidrule{1-6}
Taxonomy Medical       & 27.7\%{\footnotesize$\pm$0.6} & 32.5\%{\footnotesize$\pm$0.7} & 28.1\%{\footnotesize$\pm$1.3} & 67.0\%{\footnotesize$\pm$2.1} & 29.5\%{\footnotesize$\pm$1.5} \\
Taxonomy Medical w/ DCLM & \textbf{31.5\%}{\footnotesize$\pm$0.6} & \textbf{32.7\%}{\footnotesize$\pm$0.7} & \textbf{30.1\%}{\footnotesize$\pm$1.3} & 68.6\%{\footnotesize$\pm$2.1} & \textbf{39.2\%}{\footnotesize$\pm$1.6} \\
\bottomrule
\end{tabular}
\vspace{0.5\baselineskip}
\caption{Accuracy (mean {\footnotesize$\pm$} standard error) on four medical QA
benchmarks and the MMLU medical subset.}
\label{tab:medical-qa-evals}
\end{table}

Across medical evaluations, \ETAX \ Med w/ DCLM either achieves the best performance or performs within standard error of the best dataset (Table~\ref{tab:medical-qa-evals}). Both taxonomy-based medical datasets are able to perform above random chance (\(\approx\)25\%) on MedQA-USMLE, where DCLM-baseline and TheBlueScrubs-v1 are unable to do so. TheBlueScrubs-v1 is also unable to perform above chance on CareQA-en. Across all evals, \ETAX \ Med w/ DCLM beats DCLM-baseline by +3.2pp (8.6\%) and TheBlueScrubs-v1 by +4.9pp (13.8\%). 

\subsection{STEM}

In addition to evaluating \ETAX \ in domains with existing, multi-billion-token baseline datasets curated from Common Crawl, we also select a domain where we wish there was a large-scale dataset available. Given the importance of STEM domains for reasoning and benchmarking performance of LLMs, we curate a large, STEM-specific dataset. We select two high-performing, general datasets as baselines: DCLM-baseline and FineWeb-Edu \citep{penedo2024finewebdatasetsdecantingweb}. We benchmark the performance of each dataset (Table~\ref{tab:stem_and_general_datasets}) by annealing \textbf{General-Base}.

\begin{table}[htbp]
\centering
\setlength{\tabcolsep}{6pt}            
\begin{tabular}{l c  p{7cm}}      
\toprule
\textbf{Dataset} & \textbf{Size (T tok.)} & \textbf{Domain–specific curation notes} \\
\midrule
DCLM-baseline & 3.65 &  instruction density classifier (DCLM fastText) \\
FineWeb-Edu & 1.27 & educational quality classifier  \\
\ETAX \ STEM & 1.74 & \ETAX \ filter \\
\ETAX \ STEM w/ DCLM  & 0.91 & \ETAX \ filter, DCLM fastText\\
\bottomrule
\end{tabular}
\vspace{0.5\baselineskip}
\caption{Overview of high-perfoming general web datasets and taxonomy-based stem dataset.}
\label{tab:stem_and_general_datasets}
\end{table}

\subsubsection{Taxonomy-Based STEM Dataset}

We prepare the \ETAX \ STEM (1742B tokens) targeting science, engineering, medical, and computer science documents. We select high quality document types per sub-topic and filter for documents that exhibit reasoning (Algorithm~\ref{lst:taxonomy-stem-filter}). We then apply the DCLM classifier, at the same threshold as DCLM-baseline, to the \ETAX \ STEM dataset to construct \ETAX \ w/ DCLM (912B tokens, Algorithm~\ref{lst:taxonomy-stem-w-dclmfilter}). 

\subsubsection{Downstream STEM Results}

\begin{table}[h]
\centering
\begin{tabular}{lc}
\toprule
\textbf{Model} & \textbf{MMLU–STEM} \\
\midrule
DCLM-baseline         & 27.7\%$\scriptstyle\pm0.8$ \\
FineWeb-Edu           & 26.7\%$\scriptstyle\pm0.8$ \\
\cmidrule{1-2}
Taxonomy STEM         & 29.1\%$\scriptstyle\pm0.8$ \\
Taxonomy STEM w/ DCLM & \textbf{34.5\%}$\scriptstyle\pm0.8$ \\
\bottomrule
\end{tabular}
\vspace{0.5\baselineskip}
\caption{Accuracy (\% $\pm$ standard error) on the MMLU–STEM subset.}
\label{tab:mmlu-stem-results}
\end{table}

\ETAX \ STEM is able to outperform DCLM-baseline and FineWeb-Edu beyond standard error on MMLU-STEM. \ETAX \ w/ DCLM outperforms DCLM-baseline by +6.8pp (24.5\%) and FineWeb-Edu by +7.8pp (29.2\%).

\section{Developing Taxonomy}\label{sec:developing_taxonomy}


\subsection{Measuring Orthogonality, Correctness, \& Expressivity}
\label{sec:taxonomy_metrics}
In this section, we introduce methods to measure: (1) orthogonality, (2) correctness, and (3) expressivity.

\subsubsection{Orthogonality: Category Independence (NMI)}\label{sec:nmi}
For two categories \(C_i,C_j\), let the random variables \(X,Y\) denote their empirical primary label codes. Redundancy is measured by the normalized mutual information
\[
\mathrm{NMI}(X,Y)=\frac{2\,I(X;Y)}{H(X)+H(Y)},\qquad
I(X;Y)=\sum_{x,y}p_{xy}\log\frac{p_{xy}}{p_xp_y}.
\]
where \(H\) is Shannon entropy. \(\mathrm{NMI}=0\) indicates statistical independence, \(\mathrm{NMI}=1\) perfect duplication. This metric was proposed by \cite{wettig2025organizewebconstructingdomains} to evaluate taxonomies.

\subsubsection{Correctness: Annotator Agreement (annotator \(\kappa\))}\label{sec:kappa}
To gauge label clarity we compare a candidate model \(M\) with two gold annotators (\texttt{GPT-4o} \citep{openai2024gpt4ocard} and \texttt{Claude Sonnet-3.5} \citep{anthropic2024claude35}) via a variant of Cohen’s \(\kappa\) \citep{Cohen1960ACO}.

\paragraph{Annotation format.}
Each annotator outputs an ordered set
\[
S\in\{\varnothing,\{\ell\},\{\ell_1,\ell_2\}\}.
\]
Under the taxonomy (Definition~\ref{def:taxonomy}) a primary label is mandatory; the empty set~\(\varnothing\) arises only when a model’s raw output is malformed and cannot be parsed.  Such issues are negligible for high-capacity LLMs but more frequent for small ones.  Two annotations agree iff \(S_a\cap S_b\neq\varnothing\) or \(S_a\cup S_b = \varnothing\). 

\paragraph{Observed and expected agreement.}
Let \(P_o\) be the empirical agreement rate over held-out documents. The expected agreement \(P_e\) is computed by assuming label sets are constructed independently from each annotator’s empirical fertility distribution (probability of emitting 0/1/2 labels, where the 0-case captures parse failures) and their label-choice distribution (Appendix~\ref{estimatePe}). Then $\kappa$ is computed as usual:
\[
\kappa=\frac{P_o-P_e}{1-P_e},\qquad -1\le\kappa\le 1.
\]
We report the mean of the two gold-vs-\(M\) scores per category (annotator \(\kappa\)); a high \(\kappa\) implies unambiguous label definitions.



\subsubsection{Expressivity: Domain–Recall}\label{sec:domain-recall}

Some rare domains (math, code) drive downstream performance. To measure how well a classifier surfaces such niche material we introduce \textbf{domain–recall} score inspired by recent works that iteratively train a classifier to maximize recall of a specific domain \citep{shao2024deepseekmathpushinglimitsmathematical}:   

\begin{enumerate}[nosep,leftmargin=2em]
\item Select a small set \(\mathcal{U}=\{u_1,\dots,u_m\}\) of human-vetted base URLs that are human-judged \(\ge 90\%\) in-domain. All documents whose URL begins with any \(u\in\mathcal{U}\) form the domain positives \(D^{+}\subset D\).
\item Apply a classifier (taxonomy filter, fastText classifier, etc.) that returns a subset of documents \(\widehat{D}\subseteq D\).
\end{enumerate}

The recall of the classifier on this topic is

\[
\mathrm{Recall}=\frac{|\,\widehat{D}\cap D^{+}\,|}{|D^{+}|}.
\]

\noindent
A higher recall value means the strategy retrieves more of the trusted in-domain pages. We report recall alongside the \textbf{data kept} fraction \(|\widehat{D}|/|D|\) to indicate the overall data volume returned by the classifier.

\subsubsection{Held-out Evaluation Sets}\label{sec:taxonomy-eval-sets}

To evaluate the categories in \ETAX\ (Section~\ref{sec:taxonomy-overview}), we report the normalized mutual information and annotator \(\kappa\) on the following evaluation sets. Both evaluations are sampled from the deduplicated and heuristically-filtered Common Crawl used to prepare \EWEB\ (Appendix~\ref{sec:cc-processing-pipeline}). Neither set was seen during fine-tuning. 

\begin{enumerate}
    \item \textbf{Random Set}: 2,017 randomly sampled documents.
    \item \textbf{STEM Set}: 871 STEM documents \footnote{See Appendix~\ref{sec:stem-eval-set-breakdown} for details about the STEM evaluation set.}
\end{enumerate}

In addition, we report domain-recall for web code and math domains. We label a set of 104.6M documents with \ETAX \ using \texttt{Qwen2.5-32b-Instruct} \citep{qwen2025qwen25technicalreport}. Details about the annotation can be found in Appendix~\ref{sec:qwen32b-annotation-appendix}. Domain-recall is then calculated on documents from two sets of human-vetted "gold" URL sets (Appendix~\ref{sec:domain-recall-gold-urls}): 
\begin{enumerate}
    \item \textbf{Web Code}: 30 base-URLs; \(|D^{+}| =\) 330,934 documents
    \item \textbf{Math}: 42 base-URLs; \(|D^{+}| =\) 16,199 documents
\end{enumerate} 


\subsection{Teacher Model Selection}\label{sec:parent_model_selection}

When selecting a teacher model, we evaluate \texttt{Qwen2.5-32b-Instruct}, \texttt{Qwen2.5-72b-Instruct}, and \texttt{DeepSeek-V3} using NMI (orthogonality), annotator \(\kappa\) (correctness), and domain-recall (expressivity). We seek to maximize performance and inference efficiency of the teacher model.


\subsubsection{Annotator \(\kappa\) Results}\label{sec:kappa-results}

We report annotator \(\kappa\) between powerful open source LLMs (\texttt{DeepSeek-V3} \citep{deepseekai2025deepseekv3technicalreport}, \texttt{Qwen2.5-72b-Instruct}, and \texttt{Qwen2.5-32b-Instruct}) and our two gold annotators (\texttt{GPT-4o}, \texttt{Claude Sonnet 3.5}) for all 12 categories of \ETAX \ on the random and STEM evaluation sets in Table~\ref{tab:kappa-consolidated}. A discussion justifying the use of LLMs as gold annotators instead of humans can be found in Appendix \ref{sec:human-vs-llm}.

\begin{table}[htbp]
\centering
\begin{tabular}{lcccccc}
\toprule
\multirow{2}{*}{\textbf{Category}} &
\multicolumn{2}{c}{\textbf{DeepSeek-V3}} &
\multicolumn{2}{c}{\textbf{Qwen 2.5-72B-Inst.}} &
\multicolumn{2}{c}{\textbf{Qwen 2.5-32B-Inst.}} \\
 & Random & STEM & Random & STEM & Random & STEM \\ \midrule
Bloom Knowledge Domain       & 0.69{\scriptsize$\pm$\,0.02} & 0.64{\scriptsize$\pm$\,0.03} & 0.46{\scriptsize$\pm$\,0.02} & 0.39{\scriptsize$\pm$\,0.03} & 0.62{\scriptsize$\pm$\,0.02} & 0.68{\scriptsize$\pm$\,0.02} \\
Bloom Cognitive Process      & 0.76{\scriptsize$\pm$\,0.01} & 0.76{\scriptsize$\pm$\,0.02} & 0.67{\scriptsize$\pm$\,0.02} & 0.70{\scriptsize$\pm$\,0.02} & 0.73{\scriptsize$\pm$\,0.01} & 0.79{\scriptsize$\pm$\,0.02} \\
Document Type V1             & 0.90{\scriptsize$\pm$\,0.01} & 0.91{\scriptsize$\pm$\,0.01} & 0.88{\scriptsize$\pm$\,0.01} & 0.91{\scriptsize$\pm$\,0.01} & 0.86{\scriptsize$\pm$\,0.01} & 0.89{\scriptsize$\pm$\,0.01} \\
Free Decimal Corr. (level 1) & 0.92{\scriptsize$\pm$\,0.01} & 0.95{\scriptsize$\pm$\,0.01} & 0.90{\scriptsize$\pm$\,0.01} & 0.92{\scriptsize$\pm$\,0.01} & 0.88{\scriptsize$\pm$\,0.01} & 0.92{\scriptsize$\pm$\,0.01} \\
Free Decimal Corr. (level 2) & 0.86{\scriptsize$\pm$\,0.01} & 0.87{\scriptsize$\pm$\,0.01} & 0.83{\scriptsize$\pm$\,0.01} & 0.81{\scriptsize$\pm$\,0.01} & 0.81{\scriptsize$\pm$\,0.01} & 0.81{\scriptsize$\pm$\,0.01} \\
Free Decimal Corr. (level 3) & 0.71{\scriptsize$\pm$\,0.01} & 0.70{\scriptsize$\pm$\,0.01} & 0.67{\scriptsize$\pm$\,0.01} & 0.63{\scriptsize$\pm$\,0.01} & 0.64{\scriptsize$\pm$\,0.01} & 0.60{\scriptsize$\pm$\,0.01} \\
Extraction Artifacts         & 0.81{\scriptsize$\pm$\,0.02} & 0.86{\scriptsize$\pm$\,0.03} & 0.57{\scriptsize$\pm$\,0.02} & 0.53{\scriptsize$\pm$\,0.03} & 0.74{\scriptsize$\pm$\,0.01} & 0.65{\scriptsize$\pm$\,0.03} \\
Missing Content              & 0.83{\scriptsize$\pm$\,0.01} & 0.85{\scriptsize$\pm$\,0.02} & 0.63{\scriptsize$\pm$\,0.01} & 0.65{\scriptsize$\pm$\,0.02} & 0.66{\scriptsize$\pm$\,0.02} & 0.65{\scriptsize$\pm$\,0.02} \\
Document Type V2             & 0.89{\scriptsize$\pm$\,0.01} & 0.89{\scriptsize$\pm$\,0.01} & 0.80{\scriptsize$\pm$\,0.01} & 0.80{\scriptsize$\pm$\,0.01} & 0.85{\scriptsize$\pm$\,0.01} & 0.83{\scriptsize$\pm$\,0.01} \\
Education Level              & 0.89{\scriptsize$\pm$\,0.01} & 0.86{\scriptsize$\pm$\,0.02} & 0.82{\scriptsize$\pm$\,0.01} & 0.81{\scriptsize$\pm$\,0.02} & 0.88{\scriptsize$\pm$\,0.01} & 0.85{\scriptsize$\pm$\,0.02} \\
Reasoning Depth              & 0.75{\scriptsize$\pm$\,0.01} & 0.72{\scriptsize$\pm$\,0.02} & 0.70{\scriptsize$\pm$\,0.01} & 0.67{\scriptsize$\pm$\,0.02} & 0.67{\scriptsize$\pm$\,0.02} & 0.67{\scriptsize$\pm$\,0.02} \\
Technical Correctness        & 0.60{\scriptsize$\pm$\,0.02} & 0.61{\scriptsize$\pm$\,0.02} & 0.52{\scriptsize$\pm$\,0.01} & 0.60{\scriptsize$\pm$\,0.02} & 0.52{\scriptsize$\pm$\,0.01} & 0.51{\scriptsize$\pm$\,0.02} \\ \midrule
\textbf{Overall mean}          & 0.80 & 0.80 & 0.70 & 0.70 & 0.74 & 0.74 \\ \bottomrule
\end{tabular}
\vspace{0.5\baselineskip}
\caption{Annotator \(\kappa\) (\(\pm\) standard error) between each candidate model and two gold annotators (\texttt{GPT-4o}, \texttt{Claude Sonnet-3.5}) on the random (\(n{=}2{,}017\)) and STEM (\(n{=}871\)) evaluation sets.}
\label{tab:kappa-consolidated}
\end{table}

\paragraph{Observations.}
\texttt{DeepSeek-V3} achieves the highest average pairwise Cohen's \(\kappa\) for both random and STEM of 0.80. However, the 671B-parameter Mixture-of-Experts is expensive to serve for large-scale inference.\footnote{At the time of annotation, the \texttt{SGLang} image for DeepSeek-V3 on AMD MI300x was much slower. See Table~\ref{tab:performance-ablation} and Appendix~\ref{sec:dsv3-old-image} \citep{zheng2024sglangefficientexecutionstructured}.} \texttt{Qwen2.5-32b-Instruct} provides faster inference and is within 0.06 (7.5\%) of \texttt{DeepSeek-V3}'s performance, achieving 0.74 for the random set and STEM set, outperforming its larger sibling \texttt{Qwen2.5-72b-Instruct}. We also measure the performance of \texttt{Qwen2.5-14B-Instruct} and find that random and STEM annotator \(\kappa\) drop to 0.53 and 0.52 (see Appendix~\ref{sec:pairwise-cohen-w-qwen14b} for full table). We select \texttt{Qwen2.5-32b-Instruct} to label the training set of \EMODEL \ given its balance of fast inference speed and high annotator \(\kappa\) (see Table~\ref{tab:performance-ablation} for an analysis of inference performance).



\subsubsection{Inter-Category NMI Results}\label{sec:nmi-results}
\begin{figure}[htbp]
  \centering
  \begin{subfigure}[b]{0.45\textwidth}
    \centering
    \includegraphics[width=\linewidth]{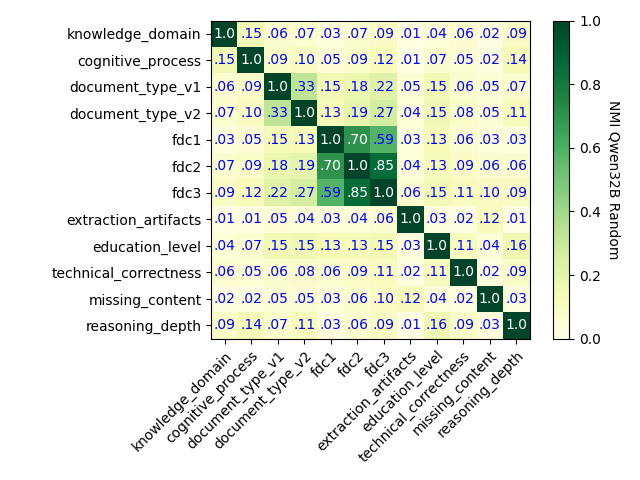}
    \caption{random NMI}
    \label{fig:qwen32b-nmi-random}
  \end{subfigure}
  \hfill
  \begin{subfigure}[b]{0.45\textwidth}
    \centering
    \includegraphics[width=\linewidth]{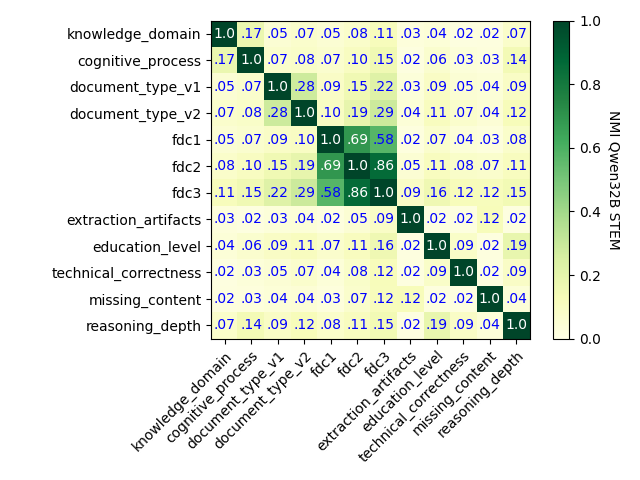}
    \caption{STEM NMI}
    \label{fig:qwen32b-nmi-stem}
  \end{subfigure}
  \caption{Comparison of NMI heatmaps on random and STEM eval sets for \texttt{Qwen2.5-32B-Instruct}}
  
  \label{fig:qwen32b-nmi-comparison}
\end{figure}

In Figure~\ref{fig:qwen32b-nmi-comparison}, we report the inter-category NMI values as a heatmap. The NMI heatmaps for \texttt{DeepSeek-V3} and \texttt{Qwen2.5-72b-Instruct} are in the Appendix~\ref{sec:os-nmi-heatmaps} and average inter-category NMIs can be found in Table~\ref{tab:nmi-avg-all-models}.

\paragraph{Observations.} The average inter-category NMI for \texttt{Qwen2.5-32b-Instruct} is 0.079 and 0.083 for random and STEM.\footnote{When calculating the average inter-category NMI we omit Document Type V2, given its similar purpose to Document Type V1, and Free Decimal Correspondence Level 1 and Level 2 to avoid double‑counting FDC hierarchical splits.} This value indicates very weak dependence between different categories of \ETAX. In addition, this value is in line with the reported inter-category NMI of 0.10 in the 2-category taxonomy defined by \cite{wettig2025organizewebconstructingdomains}. The average NMI for \texttt{Qwen2.5-32b-Instruct}, \texttt{Qwen2.5-72b-Instruct}, and \texttt{DeepSeek-V3} are all under 0.10 for both random and STEM. \texttt{Qwen2.5-32b-Instruct} has the lowest average NMI out of the three tested open-source models (Table~\ref{tab:nmi-avg-all-models}).

\subsubsection{Domain–Recall Results}\label{sec:domain-recall-results}

Table~\ref{tab:recall-qwen32b-combined} shows how effectively different filtering strategies retrieve documents from two human-vetted "gold" URL sets: web code (30 base-URLs; 330,934 documents) and math (42 base-URLs; 16,199 documents) drawn from a 104M-document Common Crawl sample. To provide a baseline we train two fastText classifiers using high quality math and code documents as target sets, with random samples of Common Crawl as negative sets \citep{bojanowski2017enrichingwordvectorssubword}. The fastText classifiers should be considered simple baselines given we do not apply techniques such as iterative recall bootstrapping or BPE tokenization to improve recall performance \citep{shao2024deepseekmathpushinglimitsmathematical, huang2025opencoderopencookbooktoptier}. Details on the training of the fastText classifiers can be found in Appendix~\ref{sec:fasttext-appendix}.

\begin{table}[htbp]
\centering
\begin{tabular}{llccc}
\toprule
\textbf{Domain} & \textbf{Filter} & \textbf{Classifier} & \textbf{Recall (\%)} & \textbf{Data kept (\%)} \\
\midrule
\multirow{3}{*}{Web Code} & FDC \(\in\) \texttt{004,005}\(^\ast\) \& Doc Type V1\(^\dagger\) & \ETAX \ filter & 94.9 & 4.80 \\
 & FDC \(\in\) \texttt{004,005}\(^\ast\) & \ETAX \ filter & 96.5 & 7.50 \\
 & score $>0.01$ & fastText web code & 25.2 & 6.40 \\
\midrule
\multirow{3}{*}{Math} & FDC \(\in\) \texttt{51 Mathematics} & \ETAX \ filter & 98.0 & 0.50 \\
 & score $>0.5$ & fastText math & 74.6 & 3.80 \\
 & score $>0.1$ & fastText math & 98.5 & 47.9 \\
\bottomrule
\end{tabular}
\vspace{0.5\baselineskip}
\caption{Domain–recall for web code and math of \texttt{Qwen2.5-32b-Instruct}-annotated \ETAX \ in comparison to basic fastText classifiers. \(^\ast\)FDC codes \texttt{004} and \texttt{005} denote Computer Science and Software Engineering. \(^\dagger\)Document Type V1 filters for "Code/Software", "Reference/Encyclopedic/Educational", and "Social/Forum".}
\label{tab:recall-qwen32b-combined}
\end{table}

\paragraph{Code.}
The Free Decimal Correspondence (FDC) alone recalls 96.5\% of vetted code pages while discarding 92.5\% of the web. Adding Document Type V1 constraint trims 35.9\% of volume of data kept with only 1.6pp loss in recall of code pages. The baseline fastText model performs significantly worse recalling only 25.2\% of target web code documents.


\paragraph{Math.}
Tagging with the single FDC level 2 code \texttt{51 - Mathematics} surfaces 98.0\% of vetted math pages while keeping just 0.5\% of the corpus—an efficient high-recall, low-volume filter. Lowering thresholds of the fastText can match recall (98.5\%) but at the cost of retaining nearly half the crawl resulting in a much lower density of math documents in the returned set of documents.

The hierarchical Free Decimal Correspondence, combined with the Document Type V1 category tags, yields very high recall for specialized domains at modest data volumes, outperforming simple, topic-specific fastText classifiers with no need for domain-specific training.

\subsubsection{Teacher Model Selection Summary}

\begin{enumerate}[leftmargin=*]
\item \textbf{Model choice.}  Given its annotator \(\kappa\)–speed trade‑off, we adopt \texttt{Qwen‑2.5‑32B‑Instruct} as the teacher for \texttt{EAI‑Taxonomy‑0.5B}.
\item \textbf{Low redundancy.}  Inter‑category NMI of <0.10 on both random and STEM evaluation sets indicates that \ETAX \ captures largely orthogonal signals.
\item \textbf{High label clarity.}  \texttt{Qwen2.5-32b-Instruct} reaches an annotator \(\kappa\) of 0.74 vs.\ two powerful gold LLM annotators.
\item \textbf{Efficient domain targeting.}  Simple \ETAX \ filters recall >96\% of vetted math and code pages while retaining <5\% of Common Crawl, far denser than fastText baselines.
\end{enumerate}
\section{Running at Scale}
\label{sec:running-at-scale}

Using a sample of the 104.6M Common Crawl documents labeled with \texttt{Qwen2.5-32b-Instruct} (Appendix~\ref{sec:qwen32b-annotation-appendix}), we fine-tune \texttt{Qwen2.5-0.5b-Instruct} to perform the taxonomy classification task. The fine-tuned classifier, \EMODEL, achieves \textbf{50 times faster} inference speed relative to prompting \texttt{Qwen2.5-32b-Instuct} while maintaining performance across NMI, annotator \(\kappa\), and domain-recall for math and web code.

\subsection{Performance Considerations}
\label{sec:performance-considerations}

To annotate at a billion-document scale, we optimize three levers to maximize throughput: (1) reducing model size, (2) shorter generations, and (3) context distillation. Table~\ref{tab:performance-ablation} summarizes speedups.

\begin{table}[htbp]
\centering
\begin{tabular}{lcccc}
\toprule
\textbf{Model} & \textbf{Prompt Tokens} & \textbf{Generation Tokens} & \textbf{RPS/GPU} & \textbf{Speed $\Delta$} \\
\midrule
DeepSeek-V3 (old image) & Original & Original & 0.14 & 0.10 \\
\cmidrule(lr){1-5}
DeepSeek-V3 & Original & Original & 0.54 & 0.39 \\
\cmidrule(lr){1-5}
Qwen2.5-72B & Original & Original & 0.62 & 0.44 \\
\cmidrule(lr){1-5}
\multirow{2}{*}{Qwen2.5-32B} & Original & Original & 1.40 & 1.00 \\
 & None & Condensed & 6.30 & 4.50 \\
\cmidrule(lr){1-5}
\multirow{4}{*}{Qwen2.5-0.5B} & Original & Original & 5.07 & 3.62 \\
 & None & Original & 5.46 & 3.90 \\
 & Original & Condensed & 50.91 & 36.36 \\
 & \textit{None} & \textit{Condensed} & \textit{70.28} & \textit{50.20} \\
  & None & Embedding & 189.23 & 135.16 \\
\bottomrule
\end{tabular}
\vspace{0.5\baselineskip}
\caption{Inference performance comparison. Requests per second per GPU are measured with prefix caching enabled. Speed $\Delta$ is calculated relative to \texttt{Qwen2.5-32B} with the original prompt which is provided in Appendix~\ref{qwen32b-prompt-1}. The italicized row indicates the configuration used by \EMODEL. When we were annotating the sample of 104M document, the \texttt{SGLang} image for \texttt{DeepSeek-V3} was much slower as shown in the table. Embedding signifies using a classification-head instead of token generation.}
\label{tab:performance-ablation}
\end{table}

\paragraph{Measuring performance.} To compare performance of different model sizes and prefill/generation strategies, we report requests per second per GPU (RPS/GPU). Given the large batch workload, we care exclusively about maximizing throughput (total number of documents processed per second). To calculate the effects of design decisions on performance, we calculate the number of tokens in the shared prefix, average document, and average generation.\footnote{These calculation were done solely using the prompt from Appendix~\ref{qwen32b-prompt-1} given we annotated the 104.6M Common Crawl documents in two passes to add additional categories before fine-tuning (Appendix~\ref{sec:qwen32b-annotation-appendix}). Therefore, these are lower bounds on the true gains given  the \textit{Original} prompt and \textit{Original} generation only encapsulate 8 categories, whereas the \textit{Condensed} generation contains all 12 categories.} We use \texttt{vLLM} for inference of the \texttt{Qwen2.5} models and \texttt{SGLang} for \texttt{DeepSeek-V3} \citep{zheng2024sglangefficientexecutionstructured, kwon2023efficientmemorymanagementlarge}.

\paragraph{Condensing model size.} We select the smallest model in the \texttt{Qwen2.5} model family. We decide to use a pre-trained small LM to take advantage of strong priors developed during training.\footnote{Recently, even smaller pre-trained LMs have been released such as \texttt{SmolLM2-135M-Instruct}, which could potentially further improve inference performance \citep{allal2025smollm2smolgoesbig}}.  When maintaining the original prompt and generation format with prefix caching enabled, we only see a \SI{3.6}{\times} increase in RPS/GPU.

\paragraph{Condensing generation tokens.} We are able to extract and condense the output generated by \texttt{Qwen2.5-32b-Instruct} programatically before fine-tuning. By reducing the average generation tokens from 791 to 51, the RPS/GPU of \texttt{Qwen2.5-0.5b} increase by \SI{10}{\times}.

\paragraph{Context distillation.} We remove the prompt during finetuning, which increases the RPS/GPU of \texttt{Qwen2.5-0.5b} by \SI{1.4}{\times} \citep{snell2022learningdistillingcontext}. This speed increase is in spite of the fact that we have prefix caching enabled during inference.

\paragraph{Why not use a classification head?} As a first iteration, we opted to condense the number of generated tokens instead of relying on a classification-head. However, there are many potential benefits to using a classification head such as persisting a document-embedding that can be re-used and faster inference (\SI{2.7}{\times} increase in RPS/GPU over the configuration used for \EMODEL). We hope to look into this in the future.

\subsection{Distillation}

To train the \texttt{Qwen2.5-0.5b-Instruct} model, we fine-tune the model on 82B tokens of documents synthetically labeled by \texttt{Qwen2.5-32b-Instruct}. We perform context distillation and condense the number of generated tokens. See Appendix~\ref{sec:qwen500-ft-gen-template} for the output format of \EMODEL. The loss is computed only on the \texttt{Qwen2.5-32b-Instruct}'s completion tokens; the input document, chat template, and system prompt are masked out during loss calculation. The hyper-parameters used for fine-tuning the model can be found in Table~\ref{tab:final-qwen500m-hparams}, Appendix~\ref{sec:qwen500-ft-final-hparams}. An ablation of prompt/generation formatting, learning rate, and token budget can be found in Appendix~\ref{sec:qwen500-ft-all-ablations}. The ablations find that there is little-to-no degradation from context distillation and generation condensation. We also find that we achieve similar annotator \(\kappa\) across categories with as little as 12B training tokens. \EMODEL \ is available on HuggingFace: \href{https://huggingface.co/EssentialAI/eai-distill-0.5b}{EssentialAI/eai-distill-0.5b}.

\subsection{\EMODEL \ Performance}
We evaluate the student model, \EMODEL, against its teacher, \texttt{Qwen2.5-32b-Instruct}, along three metrics: NMI (orthogonality), annotator \(\kappa\) (correctness), and domain recall (expressivity).


\subsubsection{Annotation Consistency (\texorpdfstring{\(\kappa\)}{kappa})}

\paragraph{Observations.}  \EMODEL's annotator \(\kappa\) decreases by 0.03 (4.1\%) and 0.01 (1.4\%) relative to \texttt{Qwen2.5-32b-Instruct} (Table~\ref{tab:kappa-finetuned-consolidated}). We observe consistent performance across the two models in both Bloom categories, Document Type V1/V2, and the Free Decimal Correspondence. \EMODEL \ exhibits stronger performance than its teacher model on Reasoning Depth and Technical Correctness and weaker performance on Extraction Artifacts, Missing Content, and Education Level. We provide in depth analysis of these shifts in Appendix~\ref{sec: kappaswings}.

\begin{table}[htbp]
\centering
\begin{tabular}{lcccc}
\toprule
\multirow{2}{*}{\textbf{Category}} &
\multicolumn{2}{c}{\textbf{Qwen2.5-32B-Inst.}} &
\multicolumn{2}{c}{\EMODELBOLD} \\
 & Random & STEM & Random & STEM \\ \midrule
Bloom Knowledge Domain       & 0.62{\scriptsize$\pm$\,0.02} & 0.68{\scriptsize$\pm$\,0.02} & 0.63{\scriptsize$\pm$\,0.02} & 0.65{\scriptsize$\pm$\,0.03} \\
Bloom Cognitive Process        & 0.73{\scriptsize$\pm$\,0.01} & 0.79{\scriptsize$\pm$\,0.02} & 0.70{\scriptsize$\pm$\,0.02} & 0.76{\scriptsize$\pm$\,0.02} \\
Document Type V1             & 0.86{\scriptsize$\pm$\,0.01} & 0.89{\scriptsize$\pm$\,0.01} & 0.83{\scriptsize$\pm$\,0.01} & 0.86{\scriptsize$\pm$\,0.01} \\
Free Decimal Corr. (level~1) & 0.88{\scriptsize$\pm$\,0.01} & 0.92{\scriptsize$\pm$\,0.01} & 0.88{\scriptsize$\pm$\,0.01} & 0.93{\scriptsize$\pm$\,0.01} \\
Free Decimal Corr. (level~2) & 0.81{\scriptsize$\pm$\,0.01} & 0.81{\scriptsize$\pm$\,0.01} & 0.81{\scriptsize$\pm$\,0.01} & 0.80{\scriptsize$\pm$\,0.01} \\
Free Decimal Corr. (level~3) & 0.64{\scriptsize$\pm$\,0.01} & 0.60{\scriptsize$\pm$\,0.01} & 0.63{\scriptsize$\pm$\,0.01} & 0.61{\scriptsize$\pm$\,0.01} \\
Extraction Artifacts         & 0.74{\scriptsize$\pm$\,0.01} & 0.65{\scriptsize$\pm$\,0.03} & 0.27{\scriptsize$\pm$\,0.02} & 0.37{\scriptsize$\pm$\,0.03} \\
Missing Content              & 0.66{\scriptsize$\pm$\,0.02} & 0.65{\scriptsize$\pm$\,0.02} & 0.48{\scriptsize$\pm$\,0.01} & 0.57{\scriptsize$\pm$\,0.02} \\
Document Type V2             & 0.85{\scriptsize$\pm$\,0.01} & 0.83{\scriptsize$\pm$\,0.01} & 0.88{\scriptsize$\pm$\,0.01} & 0.86{\scriptsize$\pm$\,0.01} \\
Education Level              & 0.88{\scriptsize$\pm$\,0.01} & 0.85{\scriptsize$\pm$\,0.02} & 0.79{\scriptsize$\pm$\,0.02} & 0.86{\scriptsize$\pm$\,0.02} \\
Reasoning Depth              & 0.67{\scriptsize$\pm$\,0.02} & 0.67{\scriptsize$\pm$\,0.02} & 0.87{\scriptsize$\pm$\,0.01} & 0.76{\scriptsize$\pm$\,0.02} \\
Technical Correctness        & 0.52{\scriptsize$\pm$\,0.01} & 0.51{\scriptsize$\pm$\,0.02} & 0.72{\scriptsize$\pm$\,0.01} & 0.75{\scriptsize$\pm$\,0.02} \\ \midrule
\textbf{Overall mean}          & 0.74 & 0.74 & 0.71 & 0.73 \\
\bottomrule
\end{tabular}
\vspace{0.5\baselineskip}
\caption{Annotator \(\kappa\) (\(\pm\) standard error) against gold annotators on the random (\(n=2{,}017\)) and STEM (\(n=871\)) evaluation sets, after fine-tuning. We did not report \texttt{Qwen2.5-0.5b-Instruct} because it achieves an annotator \(\kappa\) of \(0.00\) for both random and STEM evaluation sets.}
\label{tab:kappa-finetuned-consolidated}
\end{table}

\subsubsection{Inter-Category NMI Results}
\label{sec:finetune-nmi-results}

\begin{table}[htbp]
\centering
\begin{tabular}{lcc}
\toprule
\textbf{Model} & \multicolumn{2}{c}{\textbf{Average NMI}} \\
\cmidrule(lr){2-3}
& \textbf{Random} & \textbf{STEM} \\
\midrule
\texttt{Claude Sonnet-3.5} & 0.083 & 0.085 \\
\texttt{GPT-4o} & 0.063 & 0.084 \\
\texttt{DeepSeek-V3} & 0.093 & 0.099 \\
\texttt{Qwen2.5-72b-Instruct} & 0.092 & 0.103 \\
\texttt{Qwen2.5-32b-Instruct} & 0.079 & 0.083 \\
\EMODEL & 0.092 & 0.093 \\
\bottomrule
\end{tabular}
\vspace{0.5\baselineskip}
\caption{Average inter-category NMI results for random and STEM evaluation sets. We omit Document Type V1, given its similar purpose to Document Type V2, and Free Decimal Correspondence Level 1 and Level 2 to avoid double‑counting FDC hierarchical splits.}
\label{tab:nmi-avg-all-models}
\end{table}

\begin{figure}[htbp]
  \centering
  \begin{subfigure}[b]{0.45\textwidth}
    \centering
    \includegraphics[width=\linewidth]{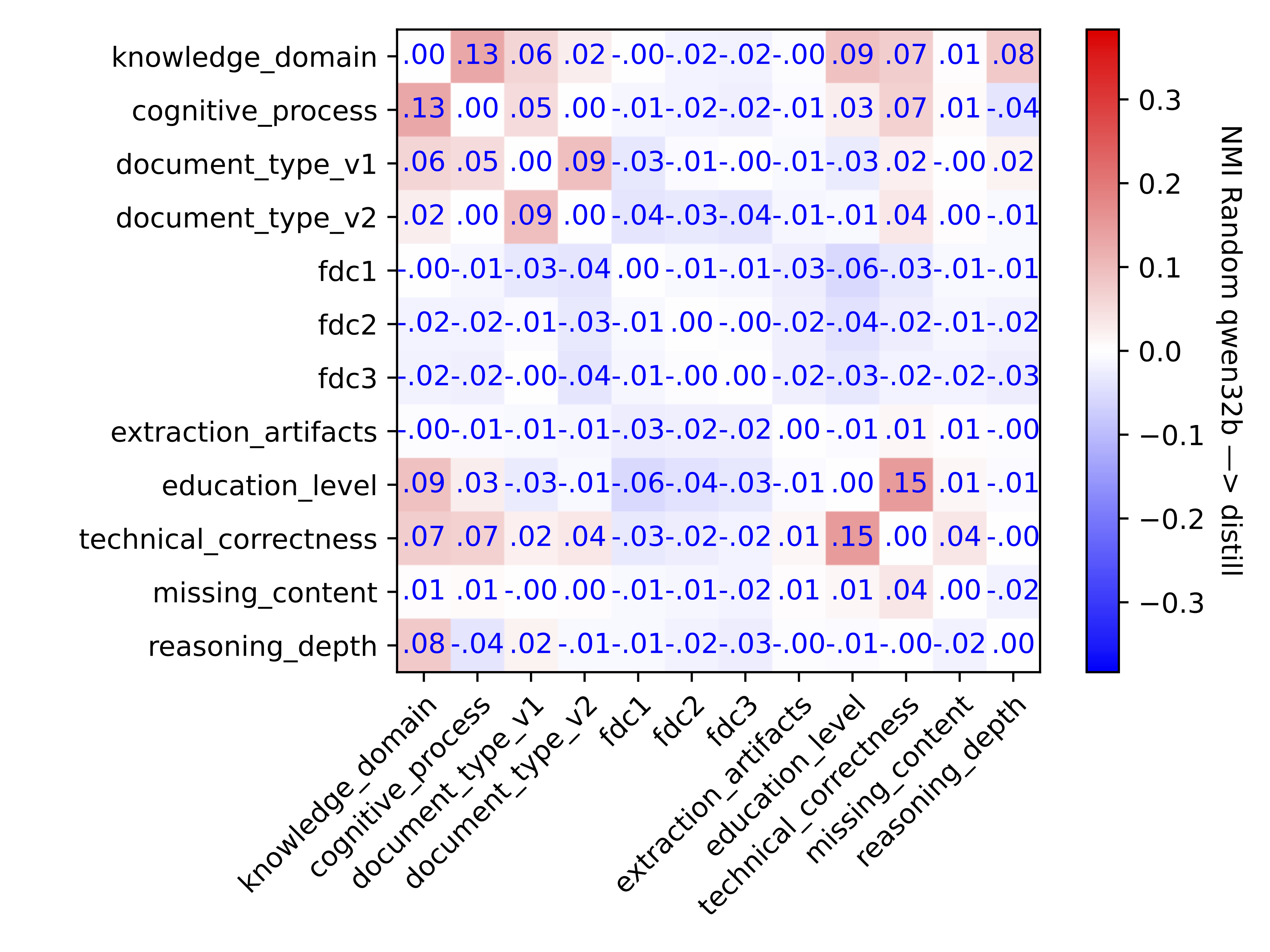}
    \caption{Random}
    \label{fig:32b-to-ft-nmi-random}
  \end{subfigure}
  \hfill
  \begin{subfigure}[b]{0.45\textwidth}
    \centering
    \includegraphics[width=\linewidth]{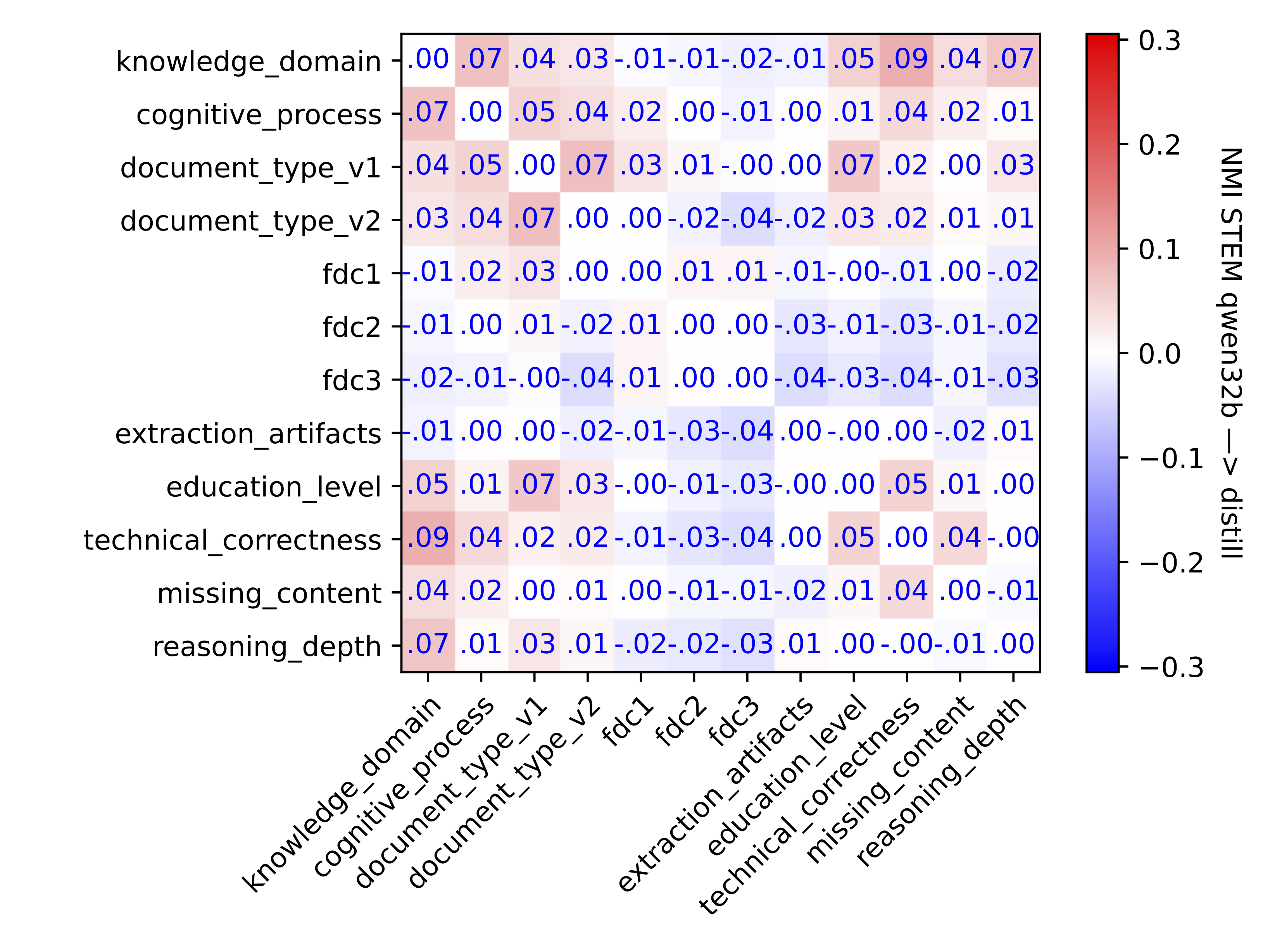}
    \caption{STEM}
    \label{fig:32b-to-ft-nmi-stem}
  \end{subfigure}
  \caption{Change in inter-category NMI per category from \texttt{Qwen2.5-32b-Instruct} to \EMODEL. A higher value indicates an increase in inter-category co-occurrence.}
  
  \label{fig:ft-nmi-comparison}
\end{figure}

The average inter-category NMI of \EMODEL \ is 0.092 and 0.093 for random and STEM, increasing by 0.013 (16.5\%) and 0.01 (12.0\%) relative to \texttt{Qwen2.5-32b-Instruct}. However, \EMODEL \ remains comparable to other powerful open-source LLMs and stays below 0.10, indicating low redundancy. See Table~\ref{tab:nmi-avg-all-models} for the average inter-category NMI of all models tested and Figure~\ref{fig:ft-nmi-comparison} for heatmaps with the relative change in inter-category NMI from \texttt{Qwen2.5-32b-Instruct} to \EMODEL.

\subsubsection{Domain–Recall on Unseen 100M Document Slice}
\label{sec:finetune-recall-results}

When calculating domain-recall for \EMODEL, we sample a fresh set 102.6M unseen documents from Common Crawl because a subset of the data used to calculate domain-recall in Section~\ref{sec:domain-recall-results} was used to train \EMODEL. The size of \(D^+\) for math and web code for the new set of 102.6M documents are are:
\begin{itemize}
    \item \textbf{Web Code}: \(|D^{+}| = 266{,}051 \text{ documents}\)
    \item \textbf{Math}: \(|D^{+}| = 12{,}835 \text{ documents}\)
\end{itemize} 

\begin{table}[ht]
\centering
\begin{tabular}{llccc}
\toprule
\textbf{Domain} & \textbf{Filter} & \textbf{Model} & \textbf{Recall (\%)} & \textbf{Data kept (\%)} \\
\midrule
\multirow{4}{*}{Web Code} & FDC \(\in\) \texttt{004,005} \& Doc Type V1 & \EMODEL & 95.8 & 4.91 \\
 & FDC \(\in\) \texttt{004,005} \& Doc Type V1 & \texttt{Qwen2.5-32b-Instruct} & 94.9 & 4.79 \\
 & FDC \(\in\) \texttt{004,005} & \EMODEL & 97.1 & 7.61 \\
 & FDC \(\in\) \texttt{004,005} & \texttt{Qwen2.5-32b-Instruct} & 96.5 & 7.47 \\
\midrule
\multirow{2}{*}{Math} & FDC \(\in\) \texttt{51 Mathematics} & \EMODEL & 97.6 & 0.52 \\
 & FDC \(\in\) \texttt{51 Mathematics} & \texttt{Qwen2.5-32b-Instruct} & 98.0 & 0.50 \\
\bottomrule
\end{tabular}
\vspace{0.5\baselineskip}
\caption{Domain–recall for web code and math domains. FDC codes \texttt{004} and \texttt{005} correspond to Computer Science and Software Engineering. The Document Type V1 filters for "Code/Software", "Reference/Encyclopedic/Educational", and "Social/Forum". \texttt{Qwen2.5-32b-Instruct} measured on the original 104.6M-document corpus; numbers shown for reference.}
\label{tab:finetuned-recall-combined}
\end{table}

\medskip
\noindent
\textbf{Observations.}  \EMODEL \ model matches \texttt{Qwen2.5-32b-Instruct} within \(\pm\)1pp recall while preserving the same thin data footprint (<1\% of crawl for math, \(\approx\)5\% for code with tight filters) as seen in Table~\ref{tab:finetuned-recall-combined}.\footnote{Because \texttt{Qwen2.5-32b-Instruct} was evaluated on a different sample of Common Crawl, small drift is expected; nevertheless, parity holds. We did not re-evaluate \texttt{Qwen2.5-32b-Instruct} on the new sample given the cost to relabel 102.6M documents.}

\subsubsection{\EMODEL\ Performance Summary}
\label{sec:ft-taxonomy-eval-summary}

\begin{enumerate}[leftmargin=*]
\item \textbf{50× faster annotation.}  Shrinking to a 0.5B-parameter model, removing the prompt, and condensing generations boost throughput from 1.4 to 70 requests per second per GPU.
\item \textbf{Minimal quality loss.}  Average annotator \(\kappa\) across random and STEM sets of \EMODEL \ falls by <3\% (0.74 → 0.72) and inter‑category NMI stays <0.10, preserving label quality and low redundancy.
\item \textbf{Domain coverage intact.}  On an unseen 102M‑document sample of Common Crawl, math and code recall remain within 1pp of \texttt{Qwen2.5-32b-Instruct} while keeping the same compact filtered sets (<5\% of data).
\end{enumerate}


\section{Conclusion}

\EWEB \ delivers 24 trillion tokens of web data with document-level annotations spanning subject, page type, content complexity, and quality. Applying simple filters on these labels produce competitive corpora for mathematics, code, medicine, and STEM in minutes, not months — competing with or surpassing specialist datasets while avoiding domain‑specific training.

Building such a corpus required three technical ingredients: (1) A principled metric suite of NMI for redundancy, annotator \(\kappa\) for label clarity, and domain‑recall for domain-specific filtering to guide taxonomy design. (2) A fast annotator, \EMODEL, that retains the teacher’s quality (<3\% drop in annotator \(\kappa\), <1pp drop in recall of math and code, average inter-category NMI below 0.10) yet runs \SI{50}{\times} faster, making billion‑page labeling practical. (3) A transparent release with all data and \EMODEL \ available on HuggingFace, enabling reproducibility and iteration.

\subsection{Broader impact}
Structured web data transforms corpus curation from an complex, expensive processing pipeline into a search problem that anyone can solve. We hope \EWEB \ becomes a community commons: a foundation others can refine, audit, or curate in new ways, accelerating open research on LLM training data, arguably the most valuable, yet least shared, asset contributing to modern LLM capabilities.

\subsection{Possible Next Steps}
Taxonomies serve as an excellent bridge to build an interpretable coordinate system on data, despite their biases. In addition to using our \EWEB\ for organic and synthetic data curation, we see new opportunities to automatically curate data to train large neural networks at scale. Whether taxonomies will remain a core element of state-of-the-art data pipelines or merely serve as a stepping-stone en route to fully unsupervised methods is still an open question.

\bibliography{main}

\begin{thebibliography}{67}
\providecommand{\natexlab}[1]{#1}
\providecommand{\url}[1]{\texttt{#1}}
\expandafter\ifx\csname urlstyle\endcsname\relax
  \providecommand{\doi}[1]{doi: #1}\else
  \providecommand{\doi}{doi: \begingroup \urlstyle{rm}\Url}\fi

\bibitem[Allal et~al.(2025)Allal, Lozhkov, Bakouch, Blázquez, Penedo, Tunstall, Marafioti, Kydlíček, Lajarín, Srivastav, Lochner, Fahlgren, Nguyen, Fourrier, Burtenshaw, Larcher, Zhao, Zakka, Morlon, Raffel, von Werra, and Wolf]{allal2025smollm2smolgoesbig}
Loubna~Ben Allal, Anton Lozhkov, Elie Bakouch, Gabriel~Martín Blázquez, Guilherme Penedo, Lewis Tunstall, Andrés Marafioti, Hynek Kydlíček, Agustín~Piqueres Lajarín, Vaibhav Srivastav, Joshua Lochner, Caleb Fahlgren, Xuan-Son Nguyen, Clémentine Fourrier, Ben Burtenshaw, Hugo Larcher, Haojun Zhao, Cyril Zakka, Mathieu Morlon, Colin Raffel, Leandro von Werra, and Thomas Wolf.
\newblock Smollm2: When smol goes big -- data-centric training of a small language model, 2025.
\newblock URL \url{https://arxiv.org/abs/2502.02737}.

\bibitem[Anderson and Krathwohl(2001)]{bloom2001}
L.W. Anderson and D.R. Krathwohl.
\newblock \emph{A Taxonomy for Learning, Teaching, and Assessing: A Revision of Bloom's Taxonomy of Educational Objectives}.
\newblock Longman, 2001.

\bibitem[{Anthropic}(2024)]{anthropic2024claude35}
{Anthropic}.
\newblock {Claude 3.5 Sonnet Model Card Addendum}.
\newblock Model card, Anthropic, June 2024.
\newblock URL \url{https://www-cdn.anthropic.com/fed9cc193a14b84131812372d8d5857f8f304c52/Model_Card_Claude_3_Addendum.pdf}.
\newblock Accessed 23 May 2025.

\bibitem[Arias-Duart et~al.(2025)Arias-Duart, Martin-Torres, Hinjos, Bernabeu-Perez, Ganzabal, Mallo, Gururajan, Lopez-Cuena, Alvarez-Napagao, and Garcia-Gasulla]{ariasduart2025automaticevaluationhealthcarellms}
Anna Arias-Duart, Pablo~Agustin Martin-Torres, Daniel Hinjos, Pablo Bernabeu-Perez, Lucia~Urcelay Ganzabal, Marta~Gonzalez Mallo, Ashwin~Kumar Gururajan, Enrique Lopez-Cuena, Sergio Alvarez-Napagao, and Dario Garcia-Gasulla.
\newblock Automatic evaluation of healthcare llms beyond question-answering, 2025.
\newblock URL \url{https://arxiv.org/abs/2502.06666}.

\bibitem[Austin et~al.(2021)Austin, Odena, Nye, Bosma, Michalewski, Dohan, Jiang, Cai, Terry, Le, and Sutton]{austin2021programsynthesislargelanguage}
Jacob Austin, Augustus Odena, Maxwell Nye, Maarten Bosma, Henryk Michalewski, David Dohan, Ellen Jiang, Carrie Cai, Michael Terry, Quoc Le, and Charles Sutton.
\newblock Program synthesis with large language models, 2021.
\newblock URL \url{https://arxiv.org/abs/2108.07732}.

\bibitem[Bahri et~al.(2024)Bahri, Dyer, Kaplan, Lee, and Sharma]{Bahri_2024}
Yasaman Bahri, Ethan Dyer, Jared Kaplan, Jaehoon Lee, and Utkarsh Sharma.
\newblock Explaining neural scaling laws.
\newblock \emph{Proceedings of the National Academy of Sciences}, 121\penalty0 (27), June 2024.
\newblock ISSN 1091-6490.
\newblock \doi{10.1073/pnas.2311878121}.
\newblock URL \url{http://dx.doi.org/10.1073/pnas.2311878121}.

\bibitem[Bai et~al.(2023)Bai, Bai, Chu, Cui, Dang, Deng, Fan, Ge, Han, Huang, Hui, Ji, Li, Lin, Lin, Liu, Liu, Lu, Lu, Ma, Men, Ren, Ren, Tan, Tan, Tu, Wang, Wang, Wang, Wu, Xu, Xu, Yang, Yang, Yang, Yang, Yao, Yu, Yuan, Yuan, Zhang, Zhang, Zhang, Zhang, Zhou, Zhou, Zhou, and Zhu]{bai2023qwentechnicalreport}
Jinze Bai, Shuai Bai, Yunfei Chu, Zeyu Cui, Kai Dang, Xiaodong Deng, Yang Fan, Wenbin Ge, Yu~Han, Fei Huang, Binyuan Hui, Luo Ji, Mei Li, Junyang Lin, Runji Lin, Dayiheng Liu, Gao Liu, Chengqiang Lu, Keming Lu, Jianxin Ma, Rui Men, Xingzhang Ren, Xuancheng Ren, Chuanqi Tan, Sinan Tan, Jianhong Tu, Peng Wang, Shijie Wang, Wei Wang, Shengguang Wu, Benfeng Xu, Jin Xu, An~Yang, Hao Yang, Jian Yang, Shusheng Yang, Yang Yao, Bowen Yu, Hongyi Yuan, Zheng Yuan, Jianwei Zhang, Xingxuan Zhang, Yichang Zhang, Zhenru Zhang, Chang Zhou, Jingren Zhou, Xiaohuan Zhou, and Tianhang Zhu.
\newblock Qwen technical report, 2023.
\newblock URL \url{https://arxiv.org/abs/2309.16609}.

\bibitem[Beck et~al.(2024)Beck, Pöppel, Spanring, Auer, Prudnikova, Kopp, Klambauer, Brandstetter, and Hochreiter]{beck2024xlstmextendedlongshortterm}
Maximilian Beck, Korbinian Pöppel, Markus Spanring, Andreas Auer, Oleksandra Prudnikova, Michael Kopp, Günter Klambauer, Johannes Brandstetter, and Sepp Hochreiter.
\newblock xlstm: Extended long short-term memory, 2024.
\newblock URL \url{https://arxiv.org/abs/2405.04517}.

\bibitem[Bevendorff et~al.(2018)Bevendorff, Stein, Hagen, and Potthast]{bevendorff:2018}
Janek Bevendorff, Benno Stein, Matthias Hagen, and Martin Potthast.
\newblock {Elastic ChatNoir: Search Engine for the ClueWeb and the Common Crawl}.
\newblock In Leif Azzopardi, Allan Hanbury, Gabriella Pasi, and Benjamin Piwowarski, editors, \emph{Advances in Information Retrieval. 40th European Conference on IR Research (ECIR 2018)}, Lecture Notes in Computer Science, Berlin Heidelberg New York, March 2018. Springer.

\bibitem[Bojanowski et~al.(2017)Bojanowski, Grave, Joulin, and Mikolov]{bojanowski2017enrichingwordvectorssubword}
Piotr Bojanowski, Edouard Grave, Armand Joulin, and Tomas Mikolov.
\newblock Enriching word vectors with subword information, 2017.
\newblock URL \url{https://arxiv.org/abs/1607.04606}.

\bibitem[Chen et~al.(2021)Chen, Tworek, Jun, Yuan, de~Oliveira~Pinto, Kaplan, Edwards, Burda, Joseph, Brockman, Ray, Puri, Krueger, Petrov, Khlaaf, Sastry, Mishkin, Chan, Gray, Ryder, Pavlov, Power, Kaiser, Bavarian, Winter, Tillet, Such, Cummings, Plappert, Chantzis, Barnes, Herbert-Voss, Guss, Nichol, Paino, Tezak, Tang, Babuschkin, Balaji, Jain, Saunders, Hesse, Carr, Leike, Achiam, Misra, Morikawa, Radford, Knight, Brundage, Murati, Mayer, Welinder, McGrew, Amodei, McCandlish, Sutskever, and Zaremba]{chen2021codex}
Mark Chen, Jerry Tworek, Heewoo Jun, Qiming Yuan, Henrique~Ponde de~Oliveira~Pinto, Jared Kaplan, Harri Edwards, Yuri Burda, Nicholas Joseph, Greg Brockman, Alex Ray, Raul Puri, Gretchen Krueger, Michael Petrov, Heidy Khlaaf, Girish Sastry, Pamela Mishkin, Brooke Chan, Scott Gray, Nick Ryder, Mikhail Pavlov, Alethea Power, Lukasz Kaiser, Mohammad Bavarian, Clemens Winter, Philippe Tillet, Felipe~Petroski Such, Dave Cummings, Matthias Plappert, Fotios Chantzis, Elizabeth Barnes, Ariel Herbert-Voss, William~Hebgen Guss, Alex Nichol, Alex Paino, Nikolas Tezak, Jie Tang, Igor Babuschkin, Suchir Balaji, Shantanu Jain, William Saunders, Christopher Hesse, Andrew~N. Carr, Jan Leike, Josh Achiam, Vedant Misra, Evan Morikawa, Alec Radford, Matthew Knight, Miles Brundage, Mira Murati, Katie Mayer, Peter Welinder, Bob McGrew, Dario Amodei, Sam McCandlish, Ilya Sutskever, and Wojciech Zaremba.
\newblock Evaluating large language models trained on code, 2021.

\bibitem[Cohen(1960)]{Cohen1960ACO}
Jacob Cohen.
\newblock A coefficient of agreement for nominal scales.
\newblock \emph{Educational and Psychological Measurement}, 20:\penalty0 37 -- 46, 1960.
\newblock URL \url{https://api.semanticscholar.org/CorpusID:15926286}.

\bibitem[Collet(2025)]{xxhash}
Yann Collet.
\newblock xxhash, 2025.
\newblock URL \url{https://xxhash.com}.

\bibitem[{Common Crawl Foundation}(2025)]{commoncrawl}
{Common Crawl Foundation}.
\newblock {Common Crawl Corpus}.
\newblock \url{https://commoncrawl.org}, 2025.
\newblock all shards uo to CC-MAIN-2024-38.

\bibitem[DeepSeek-AI(2024{\natexlab{a}})]{deepseekai2024deepseekcoderv2breakingbarrierclosedsource}
DeepSeek-AI.
\newblock Deepseek-coder-v2: Breaking the barrier of closed-source models in code intelligence, 2024{\natexlab{a}}.
\newblock URL \url{https://arxiv.org/abs/2406.11931}.

\bibitem[DeepSeek-AI(2024{\natexlab{b}})]{deepseekai2024deepseekllmscalingopensource}
DeepSeek-AI.
\newblock Deepseek llm: Scaling open-source language models with longtermism, 2024{\natexlab{b}}.
\newblock URL \url{https://arxiv.org/abs/2401.02954}.

\bibitem[DeepSeek-AI(2024{\natexlab{c}})]{deepseekai2024deepseekv2strongeconomicalefficient}
DeepSeek-AI.
\newblock Deepseek-v2: A strong, economical, and efficient mixture-of-experts language model, 2024{\natexlab{c}}.
\newblock URL \url{https://arxiv.org/abs/2405.04434}.

\bibitem[DeepSeek-AI(2025)]{deepseekai2025deepseekv3technicalreport}
DeepSeek-AI.
\newblock Deepseek-v3 technical report, 2025.
\newblock URL \url{https://arxiv.org/abs/2412.19437}.

\bibitem[Feinstein and Cicchetti(1990)]{feinstein1990high}
Alvan~R Feinstein and Domenic~V Cicchetti.
\newblock High agreement but low kappa: I. the problems of two paradoxes.
\newblock \emph{Journal of clinical epidemiology}, 43\penalty0 (6):\penalty0 543--549, 1990.

\bibitem[Felipe et~al.(2025)Felipe, Garcia, Naqa, Shotande, Tripathi, Rudrapatna, Rasool, Bitterman, and Valdes]{felipe2025thebluescrubsv1comprehensivecuratedmedical}
Luis Felipe, Carlos Garcia, Issam~El Naqa, Monique Shotande, Aakash Tripathi, Vivek Rudrapatna, Ghulam Rasool, Danielle Bitterman, and Gilmer Valdes.
\newblock Thebluescrubs-v1, a comprehensive curated medical dataset derived from the internet, 2025.
\newblock URL \url{https://arxiv.org/abs/2504.02874}.

\bibitem[Gao et~al.(2024)Gao, Tow, Abbasi, Biderman, Black, DiPofi, Foster, Golding, Hsu, Le~Noac'h, Li, McDonell, Muennighoff, Ociepa, Phang, Reynolds, Schoelkopf, Skowron, Sutawika, Tang, Thite, Wang, Wang, and Zou]{eval-harness}
Leo Gao, Jonathan Tow, Baber Abbasi, Stella Biderman, Sid Black, Anthony DiPofi, Charles Foster, Laurence Golding, Jeffrey Hsu, Alain Le~Noac'h, Haonan Li, Kyle McDonell, Niklas Muennighoff, Chris Ociepa, Jason Phang, Laria Reynolds, Hailey Schoelkopf, Aviya Skowron, Lintang Sutawika, Eric Tang, Anish Thite, Ben Wang, Kevin Wang, and Andy Zou.
\newblock The language model evaluation harness, 07 2024.
\newblock URL \url{https://zenodo.org/records/12608602}.

\bibitem[{Gemma 2 Team}(2024)]{gemmateam2024gemma2improvingopen}
{Gemma 2 Team}.
\newblock Gemma 2: Improving open language models at a practical size, 2024.
\newblock URL \url{https://arxiv.org/abs/2408.00118}.

\bibitem[{Gemma 3 Team}(2025)]{gemmateam2025gemma3technicalreport}
{Gemma 3 Team}.
\newblock Gemma 3 technical report, 2025.
\newblock URL \url{https://arxiv.org/abs/2503.19786}.

\bibitem[Hendrycks et~al.(2021)Hendrycks, Burns, Basart, Zou, Mazeika, Song, and Steinhardt]{hendrycks2021measuringmassivemultitasklanguage}
Dan Hendrycks, Collin Burns, Steven Basart, Andy Zou, Mantas Mazeika, Dawn Song, and Jacob Steinhardt.
\newblock Measuring massive multitask language understanding, 2021.
\newblock URL \url{https://arxiv.org/abs/2009.03300}.

\bibitem[Hoffmann et~al.(2022)Hoffmann, Borgeaud, Mensch, Buchatskaya, Cai, Rutherford, de~Las~Casas, Hendricks, Welbl, Clark, Hennigan, Noland, Millican, van~den Driessche, Damoc, Guy, Osindero, Simonyan, Elsen, Vinyals, Rae, and Sifre]{hoffmann2022chinchilla}
Jordan Hoffmann, Sebastian Borgeaud, Arthur Mensch, Elena Buchatskaya, Trevor Cai, Eliza Rutherford, Diego de~Las~Casas, Lisa~Anne Hendricks, Johannes Welbl, Aidan Clark, Thomas Hennigan, Eric Noland, Katherine Millican, George van~den Driessche, Bogdan Damoc, Aurelia Guy, Simon Osindero, Kar\'{e}n Simonyan, Erich Elsen, Oriol Vinyals, Jack Rae, and Laurent Sifre.
\newblock An empirical analysis of compute-optimal large language model training.
\newblock In S.~Koyejo, S.~Mohamed, A.~Agarwal, D.~Belgrave, K.~Cho, and A.~Oh, editors, \emph{Advances in Neural Information Processing Systems}, volume~35, pages 30016--30030. Curran Associates, Inc., 2022.
\newblock URL \url{https://proceedings.neurips.cc/paper_files/paper/2022/file/c1e2faff6f588870935f114ebe04a3e5-Paper-Conference.pdf}.

\bibitem[Huang et~al.(2025)Huang, Cheng, Liu, Hao, Song, Xu, Yang, Liu, Zhang, Chai, Yuan, Zhang, Fu, Liu, Zhang, Wang, Qi, Xu, and Chu]{huang2025opencoderopencookbooktoptier}
Siming Huang, Tianhao Cheng, J.~K. Liu, Jiaran Hao, Liuyihan Song, Yang Xu, J.~Yang, Jiaheng Liu, Chenchen Zhang, Linzheng Chai, Ruifeng Yuan, Zhaoxiang Zhang, Jie Fu, Qian Liu, Ge~Zhang, Zili Wang, Yuan Qi, Yinghui Xu, and Wei Chu.
\newblock Opencoder: The open cookbook for top-tier code large language models, 2025.
\newblock URL \url{https://arxiv.org/abs/2411.04905}.

\bibitem[Hui et~al.(2024)Hui, Yang, Cui, Yang, Liu, Zhang, Liu, Zhang, Yu, Lu, Dang, Fan, Zhang, Yang, Men, Huang, Zheng, Miao, Quan, Feng, Ren, Ren, Zhou, and Lin]{hui2024qwen25codertechnicalreport}
Binyuan Hui, Jian Yang, Zeyu Cui, Jiaxi Yang, Dayiheng Liu, Lei Zhang, Tianyu Liu, Jiajun Zhang, Bowen Yu, Keming Lu, Kai Dang, Yang Fan, Yichang Zhang, An~Yang, Rui Men, Fei Huang, Bo~Zheng, Yibo Miao, Shanghaoran Quan, Yunlong Feng, Xingzhang Ren, Xuancheng Ren, Jingren Zhou, and Junyang Lin.
\newblock Qwen2.5-coder technical report, 2024.
\newblock URL \url{https://arxiv.org/abs/2409.12186}.

\bibitem[Jin et~al.(2020)Jin, Pan, Oufattole, Weng, Fang, and Szolovits]{jin2020diseasedoespatienthave}
Di~Jin, Eileen Pan, Nassim Oufattole, Wei-Hung Weng, Hanyi Fang, and Peter Szolovits.
\newblock What disease does this patient have? a large-scale open domain question answering dataset from medical exams, 2020.
\newblock URL \url{https://arxiv.org/abs/2009.13081}.

\bibitem[Jin et~al.(2019)Jin, Dhingra, Liu, Cohen, and Lu]{jin-etal-2019-pubmedqa}
Qiao Jin, Bhuwan Dhingra, Zhengping Liu, William Cohen, and Xinghua Lu.
\newblock {P}ub{M}ed{QA}: A dataset for biomedical research question answering.
\newblock In Kentaro Inui, Jing Jiang, Vincent Ng, and Xiaojun Wan, editors, \emph{Proceedings of the 2019 Conference on Empirical Methods in Natural Language Processing and the 9th International Joint Conference on Natural Language Processing (EMNLP-IJCNLP)}, pages 2567--2577, Hong Kong, China, November 2019. Association for Computational Linguistics.
\newblock \doi{10.18653/v1/D19-1259}.
\newblock URL \url{https://aclanthology.org/D19-1259/}.

\bibitem[Kaplan et~al.(2020)Kaplan, McCandlish, Henighan, Brown, Chess, Child, Gray, Radford, Wu, and Amodei]{kaplan2020scalinglawsneurallanguage}
Jared Kaplan, Sam McCandlish, Tom Henighan, Tom~B. Brown, Benjamin Chess, Rewon Child, Scott Gray, Alec Radford, Jeffrey Wu, and Dario Amodei.
\newblock Scaling laws for neural language models, 2020.
\newblock URL \url{https://arxiv.org/abs/2001.08361}.

\bibitem[Kolesnyk and Khairova(2022)]{kolesnyk2022justification}
AS~Kolesnyk and NF~Khairova.
\newblock Justification for the use of cohen’s kappa statistic in experimental studies of nlp and text mining.
\newblock \emph{Cybernetics and Systems Analysis}, 58\penalty0 (2):\penalty0 280--288, 2022.

\bibitem[Kwon et~al.(2023)Kwon, Li, Zhuang, Sheng, Zheng, Yu, Gonzalez, Zhang, and Stoica]{kwon2023efficientmemorymanagementlarge}
Woosuk Kwon, Zhuohan Li, Siyuan Zhuang, Ying Sheng, Lianmin Zheng, Cody~Hao Yu, Joseph~E. Gonzalez, Hao Zhang, and Ion Stoica.
\newblock Efficient memory management for large language model serving with pagedattention, 2023.
\newblock URL \url{https://arxiv.org/abs/2309.06180}.

\bibitem[Li et~al.(2025)Li, Fang, Smyrnis, Ivgi, Jordan, Gadre, Bansal, Guha, Keh, Arora, Garg, Xin, Muennighoff, Heckel, Mercat, Chen, Gururangan, Wortsman, Albalak, Bitton, Nezhurina, Abbas, Hsieh, Ghosh, Gardner, Kilian, Zhang, Shao, Pratt, Sanyal, Ilharco, Daras, Marathe, Gokaslan, Zhang, Chandu, Nguyen, Vasiljevic, Kakade, Song, Sanghavi, Faghri, Oh, Zettlemoyer, Lo, El-Nouby, Pouransari, Toshev, Wang, Groeneveld, Soldaini, Koh, Jitsev, Kollar, Dimakis, Carmon, Dave, Schmidt, and Shankar]{li2025datacomplmsearchgenerationtraining}
Jeffrey Li, Alex Fang, Georgios Smyrnis, Maor Ivgi, Matt Jordan, Samir Gadre, Hritik Bansal, Etash Guha, Sedrick Keh, Kushal Arora, Saurabh Garg, Rui Xin, Niklas Muennighoff, Reinhard Heckel, Jean Mercat, Mayee Chen, Suchin Gururangan, Mitchell Wortsman, Alon Albalak, Yonatan Bitton, Marianna Nezhurina, Amro Abbas, Cheng-Yu Hsieh, Dhruba Ghosh, Josh Gardner, Maciej Kilian, Hanlin Zhang, Rulin Shao, Sarah Pratt, Sunny Sanyal, Gabriel Ilharco, Giannis Daras, Kalyani Marathe, Aaron Gokaslan, Jieyu Zhang, Khyathi Chandu, Thao Nguyen, Igor Vasiljevic, Sham Kakade, Shuran Song, Sujay Sanghavi, Fartash Faghri, Sewoong Oh, Luke Zettlemoyer, Kyle Lo, Alaaeldin El-Nouby, Hadi Pouransari, Alexander Toshev, Stephanie Wang, Dirk Groeneveld, Luca Soldaini, Pang~Wei Koh, Jenia Jitsev, Thomas Kollar, Alexandros~G. Dimakis, Yair Carmon, Achal Dave, Ludwig Schmidt, and Vaishaal Shankar.
\newblock Datacomp-lm: In search of the next generation of training sets for language models, 2025.
\newblock URL \url{https://arxiv.org/abs/2406.11794}.

\bibitem[Liu et~al.(2023)Liu, Xia, Wang, and Zhang]{liu2023codegeneratedchatgptreally}
Jiawei Liu, Chunqiu~Steven Xia, Yuyao Wang, and Lingming Zhang.
\newblock Is your code generated by chatgpt really correct? rigorous evaluation of large language models for code generation, 2023.
\newblock URL \url{https://arxiv.org/abs/2305.01210}.

\bibitem[Liu et~al.(2025)Liu, Su, Yao, Jiang, Lai, Du, Qin, Xu, Lu, Yan, Chen, Zheng, Liu, Liu, Yin, He, Zhu, Wang, Wang, Dong, Zhang, Kang, Zhang, Xu, Zhang, Wu, Zhou, and Yang]{liu2025muonscalablellmtraining}
Jingyuan Liu, Jianlin Su, Xingcheng Yao, Zhejun Jiang, Guokun Lai, Yulun Du, Yidao Qin, Weixin Xu, Enzhe Lu, Junjie Yan, Yanru Chen, Huabin Zheng, Yibo Liu, Shaowei Liu, Bohong Yin, Weiran He, Han Zhu, Yuzhi Wang, Jianzhou Wang, Mengnan Dong, Zheng Zhang, Yongsheng Kang, Hao Zhang, Xinran Xu, Yutao Zhang, Yuxin Wu, Xinyu Zhou, and Zhilin Yang.
\newblock Muon is scalable for llm training, 2025.
\newblock URL \url{https://arxiv.org/abs/2502.16982}.

\bibitem[{Llama Team}(2024)]{grattafiori2024llama3herdmodels}
{Llama Team}.
\newblock The llama 3 herd of models, 2024.
\newblock URL \url{https://arxiv.org/abs/2407.21783}.

\bibitem[Lozhkov et~al.(2024)Lozhkov, Li, Allal, Cassano, Lamy-Poirier, Tazi, Tang, Pykhtar, Liu, Wei, Liu, Tian, Kocetkov, Zucker, Belkada, Wang, Liu, Abulkhanov, Paul, Li, Li, Risdal, Li, Zhu, Zhuo, Zheltonozhskii, Dade, Yu, Krauß, Jain, Su, He, Dey, Abati, Chai, Muennighoff, Tang, Oblokulov, Akiki, Marone, Mou, Mishra, Gu, Hui, Dao, Zebaze, Dehaene, Patry, Xu, McAuley, Hu, Scholak, Paquet, Robinson, Anderson, Chapados, Patwary, Tajbakhsh, Jernite, Ferrandis, Zhang, Hughes, Wolf, Guha, von Werra, and de~Vries]{lozhkov2024starcoder2stackv2}
Anton Lozhkov, Raymond Li, Loubna~Ben Allal, Federico Cassano, Joel Lamy-Poirier, Nouamane Tazi, Ao~Tang, Dmytro Pykhtar, Jiawei Liu, Yuxiang Wei, Tianyang Liu, Max Tian, Denis Kocetkov, Arthur Zucker, Younes Belkada, Zijian Wang, Qian Liu, Dmitry Abulkhanov, Indraneil Paul, Zhuang Li, Wen-Ding Li, Megan Risdal, Jia Li, Jian Zhu, Terry~Yue Zhuo, Evgenii Zheltonozhskii, Nii Osae~Osae Dade, Wenhao Yu, Lucas Krauß, Naman Jain, Yixuan Su, Xuanli He, Manan Dey, Edoardo Abati, Yekun Chai, Niklas Muennighoff, Xiangru Tang, Muhtasham Oblokulov, Christopher Akiki, Marc Marone, Chenghao Mou, Mayank Mishra, Alex Gu, Binyuan Hui, Tri Dao, Armel Zebaze, Olivier Dehaene, Nicolas Patry, Canwen Xu, Julian McAuley, Han Hu, Torsten Scholak, Sebastien Paquet, Jennifer Robinson, Carolyn~Jane Anderson, Nicolas Chapados, Mostofa Patwary, Nima Tajbakhsh, Yacine Jernite, Carlos~Muñoz Ferrandis, Lingming Zhang, Sean Hughes, Thomas Wolf, Arjun Guha, Leandro von Werra, and Harm de~Vries.
\newblock Starcoder 2 and the stack v2: The next generation, 2024.
\newblock URL \url{https://arxiv.org/abs/2402.19173}.

\bibitem[{Meta AI}(2025)]{meta2025llama4}
{Meta AI}.
\newblock The llama 4 herd: The beginning of a new era of natively multimodal ai innovation, April 2025.
\newblock URL \url{https://ai.meta.com/blog/llama-4-multimodal-intelligence/}.
\newblock Accessed: 2025-06-15.

\bibitem[Ockerbloom(2010)]{ockerbloom2010fdc}
John~Mark Ockerbloom.
\newblock Free decimal correspondence.
\newblock \url{https://everybodyslibraries.com/free-decimal-correspondence/}, August 2010.
\newblock Released 19 August 2010; dedicated to the public domain (CC0).

\bibitem[OpenAI(2024)]{openai2024gpt4ocard}
OpenAI.
\newblock Gpt-4o system card, 2024.
\newblock URL \url{https://arxiv.org/abs/2410.21276}.

\bibitem[Pal et~al.(2022)Pal, Umapathi, and Sankarasubbu]{medmcqa}
Ankit Pal, Logesh~Kumar Umapathi, and Malaikannan Sankarasubbu.
\newblock Medmcqa: A large-scale multi-subject multi-choice dataset for medical domain question answering.
\newblock In Gerardo Flores, George~H Chen, Tom Pollard, Joyce~C Ho, and Tristan Naumann, editors, \emph{Proceedings of the Conference on Health, Inference, and Learning}, volume 174 of \emph{Proceedings of Machine Learning Research}, pages 248--260. PMLR, 07--08 Apr 2022.
\newblock URL \url{https://proceedings.mlr.press/v174/pal22a.html}.

\bibitem[Paster et~al.(2023)Paster, Santos, Azerbayev, and Ba]{paster2023openwebmathopendatasethighquality}
Keiran Paster, Marco~Dos Santos, Zhangir Azerbayev, and Jimmy Ba.
\newblock Openwebmath: An open dataset of high-quality mathematical web text, 2023.
\newblock URL \url{https://arxiv.org/abs/2310.06786}.

\bibitem[Penedo et~al.(2023)Penedo, Malartic, Hesslow, Cojocaru, Cappelli, Alobeidli, Pannier, Almazrouei, and Launay]{penedo2023refinedwebdatasetfalconllm}
Guilherme Penedo, Quentin Malartic, Daniel Hesslow, Ruxandra Cojocaru, Alessandro Cappelli, Hamza Alobeidli, Baptiste Pannier, Ebtesam Almazrouei, and Julien Launay.
\newblock The refinedweb dataset for falcon llm: Outperforming curated corpora with web data, and web data only, 2023.
\newblock URL \url{https://arxiv.org/abs/2306.01116}.

\bibitem[Penedo et~al.(2024)Penedo, Kydlíček, allal, Lozhkov, Mitchell, Raffel, Werra, and Wolf]{penedo2024finewebdatasetsdecantingweb}
Guilherme Penedo, Hynek Kydlíček, Loubna~Ben allal, Anton Lozhkov, Margaret Mitchell, Colin Raffel, Leandro~Von Werra, and Thomas Wolf.
\newblock The fineweb datasets: Decanting the web for the finest text data at scale, 2024.
\newblock URL \url{https://arxiv.org/abs/2406.17557}.

\bibitem[Qwen(2025)]{qwen2025qwen25technicalreport}
Qwen.
\newblock Qwen2.5 technical report, 2025.
\newblock URL \url{https://arxiv.org/abs/2412.15115}.

\bibitem[Rae et~al.(2022)Rae, Borgeaud, Cai, Millican, Hoffmann, Song, Aslanides, Henderson, Ring, Young, Rutherford, Hennigan, Menick, Cassirer, Powell, van~den Driessche, Hendricks, Rauh, Huang, Glaese, Welbl, Dathathri, Huang, Uesato, Mellor, Higgins, Creswell, McAleese, Wu, Elsen, Jayakumar, Buchatskaya, Budden, Sutherland, Simonyan, Paganini, Sifre, Martens, Li, Kuncoro, Nematzadeh, Gribovskaya, Donato, Lazaridou, Mensch, Lespiau, Tsimpoukelli, Grigorev, Fritz, Sottiaux, Pajarskas, Pohlen, Gong, Toyama, de~Masson~d'Autume, Li, Terzi, Mikulik, Babuschkin, Clark, de~Las~Casas, Guy, Jones, Bradbury, Johnson, Hechtman, Weidinger, Gabriel, Isaac, Lockhart, Osindero, Rimell, Dyer, Vinyals, Ayoub, Stanway, Bennett, Hassabis, Kavukcuoglu, and Irving]{rae2022scalinglanguagemodelsmethods}
Jack~W. Rae, Sebastian Borgeaud, Trevor Cai, Katie Millican, Jordan Hoffmann, Francis Song, John Aslanides, Sarah Henderson, Roman Ring, Susannah Young, Eliza Rutherford, Tom Hennigan, Jacob Menick, Albin Cassirer, Richard Powell, George van~den Driessche, Lisa~Anne Hendricks, Maribeth Rauh, Po-Sen Huang, Amelia Glaese, Johannes Welbl, Sumanth Dathathri, Saffron Huang, Jonathan Uesato, John Mellor, Irina Higgins, Antonia Creswell, Nat McAleese, Amy Wu, Erich Elsen, Siddhant Jayakumar, Elena Buchatskaya, David Budden, Esme Sutherland, Karen Simonyan, Michela Paganini, Laurent Sifre, Lena Martens, Xiang~Lorraine Li, Adhiguna Kuncoro, Aida Nematzadeh, Elena Gribovskaya, Domenic Donato, Angeliki Lazaridou, Arthur Mensch, Jean-Baptiste Lespiau, Maria Tsimpoukelli, Nikolai Grigorev, Doug Fritz, Thibault Sottiaux, Mantas Pajarskas, Toby Pohlen, Zhitao Gong, Daniel Toyama, Cyprien de~Masson~d'Autume, Yujia Li, Tayfun Terzi, Vladimir Mikulik, Igor Babuschkin, Aidan Clark, Diego de~Las~Casas, Aurelia Guy, Chris Jones,
  James Bradbury, Matthew Johnson, Blake Hechtman, Laura Weidinger, Iason Gabriel, William Isaac, Ed~Lockhart, Simon Osindero, Laura Rimell, Chris Dyer, Oriol Vinyals, Kareem Ayoub, Jeff Stanway, Lorrayne Bennett, Demis Hassabis, Koray Kavukcuoglu, and Geoffrey Irving.
\newblock Scaling language models: Methods, analysis \& insights from training gopher, 2022.
\newblock URL \url{https://arxiv.org/abs/2112.11446}.

\bibitem[Raffel et~al.(2023)Raffel, Shazeer, Roberts, Lee, Narang, Matena, Zhou, Li, and Liu]{raffel2023exploringlimitstransferlearning}
Colin Raffel, Noam Shazeer, Adam Roberts, Katherine Lee, Sharan Narang, Michael Matena, Yanqi Zhou, Wei Li, and Peter~J. Liu.
\newblock Exploring the limits of transfer learning with a unified text-to-text transformer, 2023.
\newblock URL \url{https://arxiv.org/abs/1910.10683}.

\bibitem[Shao et~al.(2024)Shao, Wang, Zhu, Xu, Song, Bi, Zhang, Zhang, Li, Wu, and Guo]{shao2024deepseekmathpushinglimitsmathematical}
Zhihong Shao, Peiyi Wang, Qihao Zhu, Runxin Xu, Junxiao Song, Xiao Bi, Haowei Zhang, Mingchuan Zhang, Y.~K. Li, Y.~Wu, and Daya Guo.
\newblock Deepseekmath: Pushing the limits of mathematical reasoning in open language models, 2024.
\newblock URL \url{https://arxiv.org/abs/2402.03300}.

\bibitem[Shen et~al.(2024{\natexlab{a}})Shen, Li, Leng, Qin, Sun, and Zhong]{shen2024scalinglawslinearcomplexity}
Xuyang Shen, Dong Li, Ruitao Leng, Zhen Qin, Weigao Sun, and Yiran Zhong.
\newblock Scaling laws for linear complexity language models, 2024{\natexlab{a}}.
\newblock URL \url{https://arxiv.org/abs/2406.16690}.

\bibitem[Shen et~al.(2024{\natexlab{b}})Shen, Tao, Ma, Neiswanger, Liu, Wang, Tan, Hestness, Vassilieva, Soboleva, and Xing]{shen2024slimpajamadcunderstandingdatacombinations}
Zhiqiang Shen, Tianhua Tao, Liqun Ma, Willie Neiswanger, Zhengzhong Liu, Hongyi Wang, Bowen Tan, Joel Hestness, Natalia Vassilieva, Daria Soboleva, and Eric Xing.
\newblock Slimpajama-dc: Understanding data combinations for llm training, 2024{\natexlab{b}}.
\newblock URL \url{https://arxiv.org/abs/2309.10818}.

\bibitem[Singhal et~al.(2022)Singhal, Azizi, Tu, Mahdavi, Wei, Chung, Scales, Tanwani, Cole-Lewis, Pfohl, Payne, Seneviratne, Gamble, Kelly, Scharli, Chowdhery, Mansfield, y~Arcas, Webster, Corrado, Matias, Chou, Gottweis, Tomasev, Liu, Rajkomar, Barral, Semturs, Karthikesalingam, and Natarajan]{singhal2022largelanguagemodelsencode}
Karan Singhal, Shekoofeh Azizi, Tao Tu, S.~Sara Mahdavi, Jason Wei, Hyung~Won Chung, Nathan Scales, Ajay Tanwani, Heather Cole-Lewis, Stephen Pfohl, Perry Payne, Martin Seneviratne, Paul Gamble, Chris Kelly, Nathaneal Scharli, Aakanksha Chowdhery, Philip Mansfield, Blaise~Aguera y~Arcas, Dale Webster, Greg~S. Corrado, Yossi Matias, Katherine Chou, Juraj Gottweis, Nenad Tomasev, Yun Liu, Alvin Rajkomar, Joelle Barral, Christopher Semturs, Alan Karthikesalingam, and Vivek Natarajan.
\newblock Large language models encode clinical knowledge, 2022.
\newblock URL \url{https://arxiv.org/abs/2212.13138}.

\bibitem[Snell et~al.(2022)Snell, Klein, and Zhong]{snell2022learningdistillingcontext}
Charlie Snell, Dan Klein, and Ruiqi Zhong.
\newblock Learning by distilling context, 2022.
\newblock URL \url{https://arxiv.org/abs/2209.15189}.

\bibitem[{Software Heritage Foundation}(2025)]{software_heritage}
{Software Heritage Foundation}.
\newblock {Software Heritage Archive}.
\newblock \url{https://archive.softwareheritage.org}, 2025.
\newblock Snapshot accessed 10 June 2025.

\bibitem[Soldaini et~al.(2024)Soldaini, Kinney, Bhagia, Schwenk, Atkinson, Authur, Bogin, Chandu, Dumas, Elazar, Hofmann, Jha, Kumar, Lucy, Lyu, Lambert, Magnusson, Morrison, Muennighoff, Naik, Nam, Peters, Ravichander, Richardson, Shen, Strubell, Subramani, Tafjord, Walsh, Zettlemoyer, Smith, Hajishirzi, Beltagy, Groeneveld, Dodge, and Lo]{soldaini2024dolmaopencorpustrillion}
Luca Soldaini, Rodney Kinney, Akshita Bhagia, Dustin Schwenk, David Atkinson, Russell Authur, Ben Bogin, Khyathi Chandu, Jennifer Dumas, Yanai Elazar, Valentin Hofmann, Ananya~Harsh Jha, Sachin Kumar, Li~Lucy, Xinxi Lyu, Nathan Lambert, Ian Magnusson, Jacob Morrison, Niklas Muennighoff, Aakanksha Naik, Crystal Nam, Matthew~E. Peters, Abhilasha Ravichander, Kyle Richardson, Zejiang Shen, Emma Strubell, Nishant Subramani, Oyvind Tafjord, Pete Walsh, Luke Zettlemoyer, Noah~A. Smith, Hannaneh Hajishirzi, Iz~Beltagy, Dirk Groeneveld, Jesse Dodge, and Kyle Lo.
\newblock Dolma: an open corpus of three trillion tokens for language model pretraining research, 2024.
\newblock URL \url{https://arxiv.org/abs/2402.00159}.

\bibitem[Sorscher et~al.(2023)Sorscher, Geirhos, Shekhar, Ganguli, and Morcos]{sorscher2023neuralscalinglawsbeating}
Ben Sorscher, Robert Geirhos, Shashank Shekhar, Surya Ganguli, and Ari~S. Morcos.
\newblock Beyond neural scaling laws: beating power law scaling via data pruning, 2023.
\newblock URL \url{https://arxiv.org/abs/2206.14486}.

\bibitem[Su et~al.(2025)Su, Kong, Lin, Jennings, Norick, Kliegl, Patwary, Shoeybi, and Catanzaro]{su2025nemotroncctransformingcommoncrawl}
Dan Su, Kezhi Kong, Ying Lin, Joseph Jennings, Brandon Norick, Markus Kliegl, Mostofa Patwary, Mohammad Shoeybi, and Bryan Catanzaro.
\newblock Nemotron-cc: Transforming common crawl into a refined long-horizon pretraining dataset, 2025.
\newblock URL \url{https://arxiv.org/abs/2412.02595}.

\bibitem[Teknium(2023)]{OpenHermes2.5}
Teknium.
\newblock Openhermes 2.5: An open dataset of synthetic data for generalist llm assistants, 2023.
\newblock URL \url{https://huggingface.co/datasets/teknium/OpenHermes-2.5}.

\bibitem[Touvron et~al.(2023{\natexlab{a}})Touvron, Lavril, Izacard, Martinet, Lachaux, Lacroix, Rozière, Goyal, Hambro, Azhar, Rodriguez, Joulin, Grave, and Lample]{touvron2023llamaopenefficientfoundation}
Hugo Touvron, Thibaut Lavril, Gautier Izacard, Xavier Martinet, Marie-Anne Lachaux, Timothée Lacroix, Baptiste Rozière, Naman Goyal, Eric Hambro, Faisal Azhar, Aurelien Rodriguez, Armand Joulin, Edouard Grave, and Guillaume Lample.
\newblock Llama: Open and efficient foundation language models, 2023{\natexlab{a}}.
\newblock URL \url{https://arxiv.org/abs/2302.13971}.

\bibitem[Touvron et~al.(2023{\natexlab{b}})Touvron, Martin, Stone, Albert, Almahairi, Babaei, Bashlykov, Batra, Bhargava, Bhosale, Bikel, Blecher, Ferrer, Chen, Cucurull, Esiobu, Fernandes, Fu, Fu, Fuller, Gao, Goswami, Goyal, Hartshorn, Hosseini, Hou, Inan, Kardas, Kerkez, Khabsa, Kloumann, Korenev, Koura, Lachaux, Lavril, Lee, Liskovich, Lu, Mao, Martinet, Mihaylov, Mishra, Molybog, Nie, Poulton, Reizenstein, Rungta, Saladi, Schelten, Silva, Smith, Subramanian, Tan, Tang, Taylor, Williams, Kuan, Xu, Yan, Zarov, Zhang, Fan, Kambadur, Narang, Rodriguez, Stojnic, Edunov, and Scialom]{touvron2023llama2openfoundation}
Hugo Touvron, Louis Martin, Kevin Stone, Peter Albert, Amjad Almahairi, Yasmine Babaei, Nikolay Bashlykov, Soumya Batra, Prajjwal Bhargava, Shruti Bhosale, Dan Bikel, Lukas Blecher, Cristian~Canton Ferrer, Moya Chen, Guillem Cucurull, David Esiobu, Jude Fernandes, Jeremy Fu, Wenyin Fu, Brian Fuller, Cynthia Gao, Vedanuj Goswami, Naman Goyal, Anthony Hartshorn, Saghar Hosseini, Rui Hou, Hakan Inan, Marcin Kardas, Viktor Kerkez, Madian Khabsa, Isabel Kloumann, Artem Korenev, Punit~Singh Koura, Marie-Anne Lachaux, Thibaut Lavril, Jenya Lee, Diana Liskovich, Yinghai Lu, Yuning Mao, Xavier Martinet, Todor Mihaylov, Pushkar Mishra, Igor Molybog, Yixin Nie, Andrew Poulton, Jeremy Reizenstein, Rashi Rungta, Kalyan Saladi, Alan Schelten, Ruan Silva, Eric~Michael Smith, Ranjan Subramanian, Xiaoqing~Ellen Tan, Binh Tang, Ross Taylor, Adina Williams, Jian~Xiang Kuan, Puxin Xu, Zheng Yan, Iliyan Zarov, Yuchen Zhang, Angela Fan, Melanie Kambadur, Sharan Narang, Aurelien Rodriguez, Robert Stojnic, Sergey Edunov, and Thomas
  Scialom.
\newblock Llama 2: Open foundation and fine-tuned chat models, 2023{\natexlab{b}}.
\newblock URL \url{https://arxiv.org/abs/2307.09288}.

\bibitem[Weber et~al.(2024{\natexlab{a}})Weber, Fu, Anthony, Oren, Adams, Alexandrov, Lyu, Nguyen, Yao, Adams, Athiwaratkun, Chalamala, Chen, Ryabinin, Dao, Liang, Ré, Rish, and Zhang]{weber2024redpajamaopendatasettraining}
Maurice Weber, Daniel Fu, Quentin Anthony, Yonatan Oren, Shane Adams, Anton Alexandrov, Xiaozhong Lyu, Huu Nguyen, Xiaozhe Yao, Virginia Adams, Ben Athiwaratkun, Rahul Chalamala, Kezhen Chen, Max Ryabinin, Tri Dao, Percy Liang, Christopher Ré, Irina Rish, and Ce~Zhang.
\newblock Redpajama: an open dataset for training large language models, 2024{\natexlab{a}}.
\newblock URL \url{https://arxiv.org/abs/2411.12372}.

\bibitem[Weber et~al.(2024{\natexlab{b}})Weber, Fu, Anthony, Oren, Adams, Alexandrov, Lyu, Nguyen, Yao, Adams, Athiwaratkun, Chalamala, Chen, Ryabinin, Dao, Liang, Ré, Rish, and Zhang]{weber2024redpajama}
Maurice Weber, Daniel~Y. Fu, Quentin Anthony, Yonatan Oren, Shane Adams, Anton Alexandrov, Xiaozhong Lyu, Huu Nguyen, Xiaozhe Yao, Virginia Adams, Ben Athiwaratkun, Rahul Chalamala, Kezhen Chen, Max Ryabinin, Tri Dao, Percy Liang, Christopher Ré, Irina Rish, and Ce~Zhang.
\newblock Redpajama: an open dataset for training large language models.
\newblock \emph{NeurIPS Datasets and Benchmarks Track}, 2024{\natexlab{b}}.

\bibitem[Wettig et~al.(2025)Wettig, Lo, Min, Hajishirzi, Chen, and Soldaini]{wettig2025organizewebconstructingdomains}
Alexander Wettig, Kyle Lo, Sewon Min, Hannaneh Hajishirzi, Danqi Chen, and Luca Soldaini.
\newblock Organize the web: Constructing domains enhances pre-training data curation, 2025.
\newblock URL \url{https://arxiv.org/abs/2502.10341}.

\bibitem[Yang et~al.(2024)Yang, Yang, Hui, Zheng, Yu, Zhou, Li, Li, Liu, Huang, Dong, Wei, Lin, Tang, Wang, Yang, Tu, Zhang, Ma, Yang, Xu, Zhou, Bai, He, Lin, Dang, Lu, Chen, Yang, Li, Xue, Ni, Zhang, Wang, Peng, Men, Gao, Lin, Wang, Bai, Tan, Zhu, Li, Liu, Ge, Deng, Zhou, Ren, Zhang, Wei, Ren, Liu, Fan, Yao, Zhang, Wan, Chu, Liu, Cui, Zhang, Guo, and Fan]{yang2024qwen2technicalreport}
An~Yang, Baosong Yang, Binyuan Hui, Bo~Zheng, Bowen Yu, Chang Zhou, Chengpeng Li, Chengyuan Li, Dayiheng Liu, Fei Huang, Guanting Dong, Haoran Wei, Huan Lin, Jialong Tang, Jialin Wang, Jian Yang, Jianhong Tu, Jianwei Zhang, Jianxin Ma, Jianxin Yang, Jin Xu, Jingren Zhou, Jinze Bai, Jinzheng He, Junyang Lin, Kai Dang, Keming Lu, Keqin Chen, Kexin Yang, Mei Li, Mingfeng Xue, Na~Ni, Pei Zhang, Peng Wang, Ru~Peng, Rui Men, Ruize Gao, Runji Lin, Shijie Wang, Shuai Bai, Sinan Tan, Tianhang Zhu, Tianhao Li, Tianyu Liu, Wenbin Ge, Xiaodong Deng, Xiaohuan Zhou, Xingzhang Ren, Xinyu Zhang, Xipin Wei, Xuancheng Ren, Xuejing Liu, Yang Fan, Yang Yao, Yichang Zhang, Yu~Wan, Yunfei Chu, Yuqiong Liu, Zeyu Cui, Zhenru Zhang, Zhifang Guo, and Zhihao Fan.
\newblock Qwen2 technical report, 2024.
\newblock URL \url{https://arxiv.org/abs/2407.10671}.

\bibitem[Yang et~al.(2025)Yang, Li, Yang, Zhang, Hui, Zheng, Yu, Gao, Huang, Lv, Zheng, Liu, Zhou, Huang, Hu, Ge, Wei, Lin, Tang, Yang, Tu, Zhang, Yang, Yang, Zhou, Zhou, Lin, Dang, Bao, Yang, Yu, Deng, Li, Xue, Li, Zhang, Wang, Zhu, Men, Gao, Liu, Luo, Li, Tang, Yin, Ren, Wang, Zhang, Ren, Fan, Su, Zhang, Zhang, Wan, Liu, Wang, Cui, Zhang, Zhou, and Qiu]{yang2025qwen3technicalreport}
An~Yang, Anfeng Li, Baosong Yang, Beichen Zhang, Binyuan Hui, Bo~Zheng, Bowen Yu, Chang Gao, Chengen Huang, Chenxu Lv, Chujie Zheng, Dayiheng Liu, Fan Zhou, Fei Huang, Feng Hu, Hao Ge, Haoran Wei, Huan Lin, Jialong Tang, Jian Yang, Jianhong Tu, Jianwei Zhang, Jianxin Yang, Jiaxi Yang, Jing Zhou, Jingren Zhou, Junyang Lin, Kai Dang, Keqin Bao, Kexin Yang, Le~Yu, Lianghao Deng, Mei Li, Mingfeng Xue, Mingze Li, Pei Zhang, Peng Wang, Qin Zhu, Rui Men, Ruize Gao, Shixuan Liu, Shuang Luo, Tianhao Li, Tianyi Tang, Wenbiao Yin, Xingzhang Ren, Xinyu Wang, Xinyu Zhang, Xuancheng Ren, Yang Fan, Yang Su, Yichang Zhang, Yinger Zhang, Yu~Wan, Yuqiong Liu, Zekun Wang, Zeyu Cui, Zhenru Zhang, Zhipeng Zhou, and Zihan Qiu.
\newblock Qwen3 technical report, 2025.
\newblock URL \url{https://arxiv.org/abs/2505.09388}.

\bibitem[Yuan et~al.(2025)Yuan, Yu, Jiang, Padthe, Li, Wang, Kulikov, Cho, Tian, Weston, and Li]{yuan2025naturalreasoningreasoningwild28m}
Weizhe Yuan, Jane Yu, Song Jiang, Karthik Padthe, Yang Li, Dong Wang, Ilia Kulikov, Kyunghyun Cho, Yuandong Tian, Jason~E Weston, and Xian Li.
\newblock Naturalreasoning: Reasoning in the wild with 2.8m challenging questions, 2025.
\newblock URL \url{https://arxiv.org/abs/2502.13124}.

\bibitem[Zheng et~al.(2024)Zheng, Yin, Xie, Sun, Huang, Yu, Cao, Kozyrakis, Stoica, Gonzalez, Barrett, and Sheng]{zheng2024sglangefficientexecutionstructured}
Lianmin Zheng, Liangsheng Yin, Zhiqiang Xie, Chuyue Sun, Jeff Huang, Cody~Hao Yu, Shiyi Cao, Christos Kozyrakis, Ion Stoica, Joseph~E. Gonzalez, Clark Barrett, and Ying Sheng.
\newblock Sglang: Efficient execution of structured language model programs, 2024.
\newblock URL \url{https://arxiv.org/abs/2312.07104}.

\bibitem[Zhou et~al.(2025)Zhou, Wang, Ranjan, Cheng, Tang, He, Liu, and Xing]{zhou2025megamathpushinglimitsopen}
Fan Zhou, Zengzhi Wang, Nikhil Ranjan, Zhoujun Cheng, Liping Tang, Guowei He, Zhengzhong Liu, and Eric~P. Xing.
\newblock Megamath: Pushing the limits of open math corpora, 2025.
\newblock URL \url{https://arxiv.org/abs/2504.02807}.

\end{thebibliography}
\appendix


\section{Appendix}

\addtocontents{toc}{\protect\setcounter{tocdepth}{-1}}

\subsection{Contributions and Acknowledgments}

\subsubsection*{Core Contributors}

Andrew Hojel\(^*\) \\
Michael Pust\(^*\) \\
Tim Romanski \\
Yash Vanjani \\
Ritvik Kapila \\ 
Mohit Parmar \\
Ashish Vaswani

\begingroup
\renewcommand\thefootnote{\fnsymbol{footnote}}
\footnotetext[1]{Equal contribution.}
\endgroup

\subsubsection*{Contributors}

Adarsh Chaluvaraju \\
Alok Tripathy \\
Anil Thomas \\
Ashish Tanwer \\
Darsh J Shah \\
Ishaan Shah \\
Karl Stratos \\
Khoi Nguyen \\
Kurt Smith \\
Michael Callahan \\
Peter Rushton \\
Philip Monk \\
Platon Mazarakis \\
Saad Jamal \\
Saurabh Srivastava \\
Somanshu Singla \\

\FloatBarrier

\subsection{\EWEB: Dataset Overview}
\label{sec:cc-processing-pipeline}

We begin with DCLM Pool (89 resiliparse-extracted Common Crawl WARC snapshots from \texttt{CC-MAIN-2013-20} to \texttt{CC-MAIN-2022-49} \citep{commoncrawl, li2025datacomplmsearchgenerationtraining}. We then extract 12 additional snapshots from \texttt{CC-MAIN-2023-06} to \texttt{CC-MAIN-2024-38} using resiliparse \citep{bevendorff:2018}.

Post-extraction, we process the data using the following steps:
\begin{enumerate}
    \item \textbf{Generate document ids} using \texttt{xxhash.xxh3\_64\_intdigest(document)} \citep{xxhash}
    \item \textbf{Globally deduplicate} all 101 snapshots of Common Crawl using the hash-based document identifier.
    \item \textbf{Minhash LSH de-duplication} at a snapshot-level. We run Minhash LSH with a target Jaccard threshold of \texttt{0.7}, using 14 bands and 9 rows per band. Running at a snapshot-level was inspired by \cite{penedo2024finewebdatasetsdecantingweb} and \cite{deepseekai2024deepseekllmscalingopensource}.
    \item \textbf{Annotate} every document with statistical and model-based quality signals using a variant of the \texttt{RedPajama-Data-V2} processing pipeline \citep{weber2024redpajama}. The model-based signals include the DCLM-baseline fastText classifier \citep{li2025datacomplmsearchgenerationtraining}.
    \item \textbf{Filter} out low quality documents and only keep English, while maximally allowing math and code, using manually tuned quality signal filters (see Algorithm~\ref{lst:quality-score-filters}). The filters and starting values for tuning were inspired by \cite{penedo2024finewebdatasetsdecantingweb}, \cite{weber2024redpajama}, and \cite{rae2022scalinglanguagemodelsmethods}.\footnote{In the future, we hope to remove this step or automatically tune the quality signal thresholds to minimize the human-in-the-loop.}
    \item \textbf{Label} every document with \ETAX \ using \EMODEL \ (Section~\ref{sec:running-at-scale}).
\end{enumerate}

See Table~\ref{tab:cc-pipeline-table} for the stage-by-stage removal rates.

\begin{lstlisting}[basicstyle=\footnotesize\ttfamily, float,language=Python, caption={Quality Signal Filter Rules}, label={lst:quality-score-filters}]
# Rule 1: Initial Quality Filters (applied to ALL documents)
RULE_1_CONDITIONS = [
    word_count < 50,
    frac_chars_top_2gram > 0.20,
    frac_chars_top_3gram > 0.18,
    frac_chars_dupe_10grams > 0.50,
    frac_chars_dupe_9grams > 0.52,
    frac_chars_dupe_8grams > 0.54,
    frac_chars_dupe_7grams > 0.56,
    frac_chars_dupe_6grams > 0.58,
    frac_chars_dupe_5grams > 0.60
]

# Rule 2: Bypass Conditions (if ANY is true, skip Rule 3)
RULE_2_CONDITIONS = [
    ml_math_score > 0.3,
    ml_web_code_score > 0.3
]

# Rule 3: Additional Filters (only if Rule 2 bypass fails)
RULE_3_CONDITIONS = [
    frac_unique_words > 0.95,
    frac_no_alph_words > 0.6,
    ldnoobw_words > 10,
    ml_english_score < 0.6
]

# Quality Signals Filter Pipeline
for each document d in corpus D:
    # Step 1: Apply initial quality filters
    if ANY condition in RULE_1_CONDITIONS is true:
        REJECT document d
    
    # Step 2: Check for bypass conditions
    if ANY condition in RULE_2_CONDITIONS is true:
        ACCEPT document d  # Skip Rule 3
    
    # Step 3: Apply additional filters
    if ANY condition in RULE_3_CONDITIONS is true:
        REJECT document d
    else:
        ACCEPT document d
\end{lstlisting}

\begin{table}[ht!]
\centering
\begin{tabular}{lccc}
\toprule
\textbf{Processing step} & \textbf{\% removed} & \textbf{Cumulative \% removed} & \textbf{Docs (B)} \\
\midrule
101 raw CC snapshots          & --      & --   & 248.4 \\
Snapshot exact‐dedup   & 32.1\%  & 32.1\% & 168.7 \\
Global exact‐dedup     & 45.6\%  & 63.1\% & 91.7 \\
Snapshot LSH dedup     & 23.1\%  & 71.6\% & 70.5 \\
Language filter        & 50.8\%  & 86.0\% & 34.8 \\
Document quality filters  & 32.1\%  & 90.5\% & 23.6 \\
\bottomrule
\end{tabular}
\vspace{0.5\baselineskip}
\caption{Common-Crawl processing pipeline. "\% removed’’ is the drop at that
stage; "Cumulative \% removed’’ is with respect to the original 248.4 B documents. Some steps are run at the same time or in parallel, so we estimate the step-level removal rate on a subset of documents.}
\label{tab:cc-pipeline-table}
\end{table}

See Table~\ref{tab:eai-datasets-table} for the datasets used and released in this work.

\begin{table}[ht]
\centering
\begin{tabular}{ll}
\toprule
\textbf{Dataset Type} & \textbf{Hugging Face Repository} \\
\midrule
\EWEB & \href{https://huggingface.co/datasets/EssentialAI/essential-web-v1.0}{EssentialAI/essential-web-v1.0} \\
\EWEB\ 1T FDC L2 & \href{https://huggingface.co/datasets/EssentialAI/essential-web-fdc-level-2-partitioned}{EssentialAI/essential-web-1t-sample-fdc-partitioned} \\
\ETAX\ w/ FM & \href{https://huggingface.co/datasets/EssentialAI/eai-taxonomy-math-w-fm}{EssentialAI/eai-taxonomy-math-w-fm} \\
\ETAX\ w/ DCLM & \href{https://huggingface.co/datasets/EssentialAI/eai-taxonomy-code-w-dclm}{EssentialAI/eai-taxonomy-code-w-dclm} \\
\ETAX\ Med w/ DCLM & \href{https://huggingface.co/datasets/EssentialAI/eai-taxonomy-med-w-dclm}{EssentialAI/eai-taxonomy-med-w-dclm} \\
\ETAX\ STEM w/ DCLM & \href{https://huggingface.co/datasets/EssentialAI/eai-taxonomy-stem-w-dclm}{EssentialAI/eai-taxonomy-stem-w-dclm} \\
\bottomrule
\end{tabular}
\vspace{0.5\baselineskip}
\caption{Datasets we release in this work.}
\label{tab:eai-datasets-table}
\end{table}

\FloatBarrier

\subsection{Descriptions of Taxonomy Categories}\label{sec:in-depth-taxonomy}
In Section~\ref{sec:taxonomy-overview} we briefly introduce the categories in our taxonomy. What follows is an in-depth description of those categories and their labels.

\subsubsection{FDC}
The Free Decimal Correspondence \citep{ockerbloom2010fdc} is a set of decimal numbers, each corresponding to a group of subjects or disciplines. It is intended to be compatible with the Dewey Decimal System \protect\footnote{"Dewey," "Dewey Decimal," "Dewey Decimal Classification", and "DDC" are trademarks of OCLC.}, popularly used to catalog libraries. We use the FDC to provide three nested categories -- \textbf{Level 1,2,3} -- each successive level being a refinement of its parent. Enumerating all codes and their corresponding labels would be impractical, but Table \ref{tab:fdc-labels} lists the codes and labels for \textbf{Level 1}.

\begin{table}[ht]
\small
\begin{tabularx}{\textwidth}{l X}
\toprule
\textbf{Code} & \textbf{Label} \\
\midrule
\textbf{Level 1} & \\
\midrule
0 & General works, books and libraries, information sciences \\
1 & Philosophy and psychology \\
2 & Religion \\
3 & Social Sciences \\
4 & Philology and Laguage and languages \\
5 & Science and Natural history \\
6 & Industrial arts and Technology and Engineering \\
7 & Arts and recreation \\
8 & Literature \\
9 & History and Geography \\
\midrule
\textbf{Level 2} & \\
\midrule
00 - 99 & Sub-divisions of Level 1 categories \\
\midrule
\textbf{Level 3} & \\
\midrule
000 - 999 & Sub-divisions of Level 2 categories \\
\bottomrule
\end{tabularx}
\label{tab:overview-tax-quality-bands}
\vspace{0.5\baselineskip}
\caption{FDC category codes and labels. Level 2 and Level 3 labels omitted for brevity. A full set of label/code correspondences can be found at \href{https://github.com/JohnMarkOckerbloom/fdc/blob/master/fdc.txt}{https://github.com/JohnMarkOckerbloom/fdc/blob/master/fdc.txt}. A web viewer can be found at \href{https://www.librarything.com/mds}{https://www.librarything.com/mds}.}
\label{tab:fdc-labels}
\end{table}

\subsubsection{Bloom}
Bloom's Taxonomy of Educational Objectives has had many updates since its introduction in 1948. We use two categories and their labels from Anderson and Krathwohl's 2001 revision of the taxonomy \cite{bloom2001}. Table \ref{tab:bloom-labels} describes the categories and labels used, but note that we did not provide our teacher LLM model with the label descriptions, instead relying on its internal knowledge of Bloom's taxonomy.

\begin{table}[ht]
\small
\centering
\begin{tabularx}{\textwidth}{l X}
\toprule
\textbf{Label} & \textbf{Description} \\ \midrule
\textbf{Knowledge Domain} & Author demonstrates use of... \\ \midrule
Factual & basic elements to learn or solve problems in the discipline \\
Conceptual & interrelationships between basic elements within a larger context \\
Procedural & methods in the discipline \\
Metacognitive & awareness of how learning works in relation to one’s self \\ \midrule
\textbf{Cognitive Processing Level} &  Author demonstrates ability to... \\ \midrule
Remember & retrieve relevant knowledge from memory \\
Understand & determine the meaning of instructional messages \\
Apply & use a procedure in a given situation \\
Analyze & break materials into components and determine how they work together \\
Evaluate & make judgments based on criteria and standards \\
Create & create a new or original work \\ \bottomrule
\end{tabularx}
\vspace{0.5\baselineskip}
\caption{Bloom categories and label descriptions.}
\label{tab:bloom-labels}
\end{table}

\subsubsection{Document Type}
Our taxonomy includes two collections of labels that categorize by common web document types. The two collections have a good degree of overlap. Version 1 was created in-house. Version 2 is a replication of the document type category in \texttt{WebOrganizer} \citep{wettig2025organizewebconstructingdomains}. The labels for both versions are found in Table \ref{tab:doctype-labels}. In addition to the labels described, each category also has a label option \textit{Unclassified}, for documents that resist classification.

\begin{table}[ht]
\footnotesize
\begin{tabularx}{\textwidth}{>{\raggedright\arraybackslash}X >{\raggedright\arraybackslash}X | >{\raggedright\arraybackslash}X >{\raggedright\arraybackslash}X}
\toprule
\textbf{Label} & \textbf{Examples} & \textbf{Label} & \textbf{Examples} \\
\midrule
\multicolumn{4}{l}{\textbf{V1}} \\
\midrule
News / Editorial & CNN articles, opinion columns & 
Academic / Research & ArXiv papers, articles \\
Reference / Encyclopedic / Educational & FAQs, Wikipedia &
Code / Software & Github repos, code examples \\
Social / Forum & conversation threads, Q\&A boards &
Promotional / Advertisement & product pages, calls to action \\
Search / Directory / Bibliography & link pages, search results &
Adult / Pornographic & Justice Potter Stewart quotes \\
Personal / Misc & blogs, user profiles &
Machine-Generated & "lorem ipsum", garbled text \\
Legal / Regulatory & contracts, terms of service &
Government / Political & legislation proposals, press releases \\
Literary / Creative & poems, short stories &
Reviews / Criticism &  film critiques, product reviews \\
E-Commerce / Marketplace & eBay listings, Amazon product pages &
Images / Videos / Audio & YouTube video, Imgur page \\
\midrule
\multicolumn{4}{l}{\textbf{V2}} \\
\midrule
About (Org.) & org. self-description &
About (Personal) & personal profile or intro \\
Academic Writing & research paper, abstract & 
Audio Transcript & interview or court transcript, captions \\
Comment Section & Reddit, comment sections &
Content Listing & site maps, product catalogs \\
Creative Writing & song lyrics, Novel exerpts &
Documentation & API docs, README files \\
FAQ & question / answer lists &
Knowledge Article & Wikipedia, Britannica \\
Legal Notices & privacy policy, license agreement &
Listicle & Buzzfeed-style articles \\
News (Org.) & government blog posts &
News Article & newspaper digital article, CNN article \\
Nonfiction Writing & editorials, obituaries, memoirs \\
Personal Blog & individual's daily journal &
Product Page & product promotion, course description \\
Q\&A Forum & Quora, Stack Exchange &
Spam / Ads & spam content, SEO keyword stuffing \\
Structured Data & datasheets, glossaries, JSON files &
Customer Support & troubleshooting guides \\
Truncated & pay-walled site, Image galleries &
Tutorial & cooking recipies, WikiHow page \\
UserReview & Yelp or TripAdvisor review & & \\
\bottomrule
\end{tabularx}
\vspace{0.5\baselineskip}
\caption{Document Type categories and labels / examples}
\label{tab:doctype-labels}
\end{table}

\subsubsection{Content Quality}
The categories in this group are designed to assess the sophistication of material discussed in a document. They are inspired by two other efforts to categorize web data by information quality: \texttt{NaturalReasoning} \citep{yuan2025naturalreasoningreasoningwild28m}, and \texttt{FineWeb} \citep{penedo2024finewebdatasetsdecantingweb}. Table \ref{tab:iq-labels} describes the categories and labels we developed based on their work. In addition to the labels described, each category also has a label option \textit{Indeterminate}, if a document does not have enough context to make a judgement for that category. 

\begin{table}[ht]
\small
\begin{tabularx}{\textwidth}{l X}
\toprule
\textbf{Label} & \textbf{Description} \\
\midrule
\textbf{Reasoning Depth} & Document contains... \\
\midrule
No Reasoning & facts, but no evidence of reasoning about facts \\
Basic Reasoning & basic analysis, with minimal explanation and summarization \\
Intermediate Reasoning & some logical steps that connect ideas, and structured thinking, but is missing deep examples of either \\
Advanced Reasoning & multi-step reasoning, and thorough analysis with well-developed explanations \\
Exceptional Reasoning & novel abstractions, theoretical frameworks, long chain-of-thought, original insights, or proofs \\ 
\midrule
\textbf{Technical Correctness} & Document demonstrates...\\ \midrule
Technically Flawed & significant errors, inaccuracies, undermining validity of contents \\
Partially Correct & some technical correctness, but contains flaws, omissions, errors in calculation or terminology. generally understandable \\
Mostly Correct & technical correctness, with minor flaws, omissions, incomplete explanations, or other small issues \\
Highly Correct & high technical correctness, precise definitions, accurate and clear explanations, and strong command of technical material with at most minimal errors \\
Exceptionally Correct & exceptional technical correctness, formal proofs, flawless technical content, precise calculations, and mastery of material \\
\midrule
\textbf{Education Level} & Content... \\
\midrule 
General Audience & is accessible to anyone with basic literacy. Uses simple terms most can understand \\
High School Level & requires high school level education to fully comprehend. Contains specialized terminology explained for non-experts. Assumes basic background knowledge \\
Undergraduate Level & requires college-level education. Uses specialized terminology/concepts, and assumes significant subject-area background knowledge \\
Graduate/Expert Level & requires graduate-level education or domain expertise. Assumes deep background knowledge and specialized training to comprehend \\ 
\bottomrule
\end{tabularx}
\vspace{0.5\baselineskip}
\caption{Content Quality categories and label descriptions.}
\label{tab:iq-labels}
\end{table}

\subsubsection{Extraction}
We provide two sets of labels to flag issues that make a document difficult to read, likely from errors converting structured text formats such as HTML to plain text: \textbf{Extraction Artifacts} and \textbf{Missing Content}. Their labels are found in Table \ref{tab:extraction-labels}. Documents could potentially be flagged with multiple errors, but only the most notable errors are allowed to be flagged for primary / secondary labels. Either category may label a document \textit{Indeterminate} if the document does not have enough content to pass judgment on it.

\begin{table}[ht]
\small
\begin{tabularx}{\textwidth}{l X}
\toprule
\textbf{Label} & \textbf{Description} \\
\midrule
\textbf{Extraction Artifacts} & Document has... \\
\midrule
No Artifacts & no leftover HTML or irrelevant elements. Text is clean \\
Leftover HTML & HTML/code artifacts remaining after extraction \\
Text Extraction Errors & broken math expressions, encoding errors, improperly parsed tables \\
Irrelevant Content & headers, footers, nav menus, side-bars, or non-core sections extracted by mistake \\
\midrule
\textbf{Missing Content} & Document exhibits...\\
\midrule
No Missing Content & no signs of missing content. text seems complete/coherent \\
Truncated Snippets & obvious "...", incomplete paragraphs, or cut-off text \\
Click Here References & "Download here", "Click here", or references to content not present in text \\
Incoherent Flow & signs of unreadable or illogical flow due to missing key context \\
Missing Images or Figures & placeholders or references to images/figures not included in text \\
Missing Referenced Data & signs that data/tables/datasets are not included. e.g. "See Table 3", but it's absent \\
\bottomrule
\end{tabularx}
\vspace{0.5\baselineskip}
\caption{Extraction categories and label descriptions.}
\label{tab:extraction-labels}
\end{table}

\FloatBarrier

\subsection{Downstream Eval Details}
\label{sec:detailed_evals}
\FloatBarrier
\begin{table}[ht]
\centering
\setlength{\tabcolsep}{6pt}
\begin{tabular}{lc p{3cm} p{6.5cm}}
\toprule
\textbf{Eval} & \textbf{\# shots} & \textbf{Task Name} & \textbf{Config} \\
\midrule
MMLU           & 5 & \texttt{mmlu}              & \texttt{mmlu/default/\_mmlu.yaml} \\
GSM8K          & 8 & \texttt{gsm8k}             & \texttt{gsm8k/gsm8k.yaml} \\
MATH           & 4 & \texttt{hendrycks\_math}   & \texttt{hendrycks\_math/hendrycks\_math.yaml} \\
MBPP$^{+}$     & 3 & \texttt{mbpp\_plus}        & \texttt{mbpp/mbpp\_plus.yaml} \\
HumanEval$^{+}$& 0 & \texttt{humaneval\_plus}   & \texttt{humaneval/humaneval\_plus.yaml} \\
CareQA\_en     & 5 & \texttt{careqa\_en}        & \texttt{careqa/careqa\_en.yaml} \\
MedMCQA        & 5 & \texttt{medmcqa}           & \texttt{medmcqa/medmcqa.yaml} \\
MedQA–USMLE    & 5 & \texttt{medqa\_4options}   & \texttt{medqa/medqa.yaml} \\
PubMedQA       & 3 & \texttt{pubmedqa}          & \texttt{pubmedqa/pubmedqa.yaml} \\
\bottomrule
\end{tabular}
\vspace{0.5\baselineskip}
\caption{Number of shots and configuration files used for each evaluation. Files refer to commit \texttt{d09e03d} of the \texttt{EleutherAI/lm-evaluation-harness} repository; prepend \texttt{lm\_eval/tasks/} to each path for the full location.}
\label{tab:eval-configs}
\end{table}

\FloatBarrier

\subsection{Evaluated Open-Source Datasets}

\begin{table}[ht]
\centering
\begin{tabular}{ll}
\toprule
\textbf{Dataset Type} & \textbf{Hugging Face Repository} \\
\midrule
Math & \href{https://huggingface.co/datasets/HuggingFaceTB/finemath}{HuggingFaceTB/finemath} \\
Math & \href{https://huggingface.co/datasets/open-web-math/open-web-math}{open-web-math/open-web-math} \\
Math & \href{https://huggingface.co/datasets/LLM360/MegaMath}{LLM360/MegaMath} \\
General & \href{https://huggingface.co/datasets/mlfoundations/dclm-baseline-1.0}{mlfoundations/dclm-baseline-1.0} \\
General & \href{https://huggingface.co/datasets/HuggingFaceFW/fineweb-edu}{HuggingFaceFW/fineweb-edu} \\
Code & \href{https://huggingface.co/datasets/OpenCoder-LLM/opc-fineweb-code-corpus}{OpenCoder-LLM/opc-fineweb-code-corpus} \\
Code & \href{https://huggingface.co/datasets/bigcode/the-stack-v2-dedup}{bigcode/the-stack-v2-dedup} \\
Code & \href{https://huggingface.co/datasets/HuggingFaceTB/stack-edu}{HuggingFaceTB/stack-edu} \\
Medical & \href{https://huggingface.co/datasets/TheBlueScrubs/TheBlueScrubs-v1}{TheBlueScrubs/TheBlueScrubs-v1} \\
\bottomrule
\end{tabular}
\vspace{0.5\baselineskip}
\caption{Overview of all the open-source datasets used in Section~\ref{sec:downstream-results}.}
\label{tab:public-datasets-table}
\end{table}

\subsection{Model Config and Hyperparameters}
\label{sec:model_config_and_hparam}

\subsubsection{Architecture}
We use a modern decoder-only transformer model based on Gemma 3, which replaces the soft-capping method in Gemma 2 with QK-norm for attention stability \citep{gemmateam2025gemma3technicalreport, gemmateam2024gemma2improvingopen}. We use a local attention sliding window size of 4096.

\subsubsection{2.3B Model Config} 
For the 2.3B parameter model, we use Llama3's 128k tokenizer, an embedding dimension of 2304, an MLP dimension of 9216, and 26 layers \citep{grattafiori2024llama3herdmodels}. Given that the embedding and un-embedding matrices are tied, the total embedding parameters is 294.9M parameters. 

\subsubsection{2.3B Training Configurations}

\begin{table}[ht!]
\centering
\setlength{\tabcolsep}{5pt}
\begin{minipage}[t]{0.48\linewidth}
\centering
\begin{tabular}{lc}
\toprule
\textbf{Hyperparameter} & \textbf{Value} \\
\midrule
Optimizer                  & AdamW \\
$\beta_{1},\;\beta_{2}$    & 0.9, 0.95 \\
Weight decay               & 0.1 \\
Global batch size          & 2M tokens \\
Peak LR                    & $3.5\times10^{-5}$ \\
LR warm-up                 & 2B tokens \\
Cosine decay               & $3.5\times10^{-5}\!\rightarrow\!3.5\times10^{-6}$ \\
Total training tokens      & 320B \\
Sequence length            & 8,192 \\
\bottomrule
\end{tabular}
\subcaption{Base-model training config.}
\label{tab:base-hparams}
\end{minipage}\hfill
\begin{minipage}[t]{0.48\linewidth}
\centering
\begin{tabular}{lc}
\toprule
\textbf{Hyperparameter} & \textbf{Value} \\
\midrule
Peak LR                     & $3.5\times10^{-6}$ \\
Linear decay                & $3.5\times10^{-6}\!\rightarrow 0$ \\
Total annealing tokens      & 80B \\
\bottomrule
\end{tabular}
\subcaption{Annealing training config. Load the parameters and optimizer state from base model checkpoint at 320B tokens.}
\label{tab:anneal-hparams}
\end{minipage}
\caption{Training and annealing configs. Values not listed in Table~\ref{tab:anneal-hparams} are inherited from Table~\ref{tab:base-hparams}.}
\label{tab:base-and-anneal-hparam}
\end{table}

\subsection{Additional Math Results}
\label{sec:additional-math-results}
In Table~\ref{tab:expanded-downstream-math}, we report the math performance on an extended set of datasets. It include MegaMath Web Pro, which rewrites top math documents using Llama3.3-70b-Instruct \citep{grattafiori2024llama3herdmodels}. Its strong performance shows the promise of LLM-assisted data cleaning \citep{zhou2025megamathpushinglimitsopen}.

\begin{table}[h]
\centering
\begin{tabular}{lccc}
\toprule
\textbf{Dataset} & \textbf{GSM8K} & \textbf{MATH} & \textbf{MMLU–Math} \\
\midrule
MegaMath Web Pro         & \textbf{27.3\%}$\scriptstyle\pm1.2$ & \textbf{12.2\%}$\scriptstyle\pm0.4$ & \textbf{41.4\%}$\scriptstyle\pm0.9$ \\
FineMath 3+              & 26.4\%$\scriptstyle\pm1.4$ & 11.7\%$\scriptstyle\pm0.4$ & 32.3\%$\scriptstyle\pm1.5$ \\
OpenWebMath              & 14.6\%$\scriptstyle\pm1.1$ & 9.3\%$\scriptstyle\pm0.4$ & 29.9\%$\scriptstyle\pm1.5$ \\
MegaMath Web (Top 10\%)  & 9.8\%$\scriptstyle\pm0.9$  & 7.9\%$\scriptstyle\pm0.3$ & 29.9\%$\scriptstyle\pm1.5$ \\
DCLM-baseline            & 4.8\%$\scriptstyle\pm0.7$  & 4.4\%$\scriptstyle\pm0.3$  & 27.0\%$\scriptstyle\pm1.4$ \\
\cmidrule{1-4}
\ETAX\ Top Math          & 21.3\%$\scriptstyle\pm1.3$ & 11.0\%$\scriptstyle\pm0.4$ & 30.5\%$\scriptstyle\pm1.5$ \\
\ETAX\ Math w/ FM        & 22.4\%$\scriptstyle\pm1.3$ & 11.5\%$\scriptstyle\pm0.4$ & 30.9\%$\scriptstyle\pm1.5$ \\
\bottomrule
\end{tabular}
\vspace{0.5\baselineskip}
\caption{Model performance (mean $\pm$ standard error) on GSM8K, Hendrycks Math, and MMLU-Math on expanded set of math datasets that include LLM-assisted rewriting of documents.}
\label{tab:expanded-downstream-math}
\end{table}

\FloatBarrier

\subsection{Taxonomy-Based Dataset Filters}
\label{sec:taxonomy_filters_appendix}

\begin{lstlisting}[basicstyle=\small\ttfamily,language=Python, caption={Semantic filter used to curate \textbf{Taxonomy Top Math} dataset}, label={lst:taxonomy-top-math-filter}]
DOC_TYPE_V1 = [
    "Reference/Encyclopedic/Educational", "Code/Software", "Social/Forum", "Personal/Misc"
]

DOC_TYPE_V2 = [
    "Comment Section", "Documentation", "FAQ", "Knowledge Article", "Nonfiction Writing", 
    "Personal Blog", "Q&A Forum", "Structured Data", "Tutorial"
]

REASONING_DEPTH = [
    "Basic Reasoning", "Intermediate Reasoning", "Advanced Reasoning", "Exceptional Reasoning"
]

TECH_CORRECTNESS = ["Highly Correct", "Exceptionally Correct"]

# Free Decimal Correspondence: 51 = Mathematics
FDC_KEEP = ["51"]

# D: Essential Common Crawl (23.6B documents)
# R: Taxonomy Top Math (19.8M documents)
R = {
    d for d in D
    if prefix(d.fdc.primary) in FDC_KEEP
    and d.doc_type_v1.primary in DOC_TYPE_V1
    and d.doc_type_v2.primary in DOC_TYPE_V2
    and d.reasoning_depth.primary in REASONING_DEPTH
    and d.technical_correctness.primary in TECH_CORRECTNESS
}
\end{lstlisting}

\begin{lstlisting}[float=h!,basicstyle=\small\ttfamily,language=Python, caption={Semantic filter used to curate \textbf{Taxonomy Math w/ FM} dataset}, label={lst:taxonomy-math-w-fm-filter}]

# Free Decimal Correspondence: 51 = Mathematics
FDC_KEEP = ["51"]

# D: Essential Common Crawl (23.6B documents)
# R: Taxonomy Math w/ FM (21.6M documents)
R = {
    d for d in D
    if (prefix(d.fdc.primary) in FDC_KEEP or prefix(d.fdc.secondary) in FDC_KEEP)
    and d.finemath_score >= 3.25
}
\end{lstlisting}

\begin{lstlisting}[float=h!,basicstyle=\small\ttfamily, label={lst:taxonomy-code-filter}, caption={Filter used to curate \textbf{Taxonomy Code} dataset}]
DOC_TYPE_V1 = [
    "Reference/Encyclopedic/Educational", "Social/Forum"
]

DOC_TYPE_V2 = [
    "Comment Section", "Documentation", "Knowledge Article", "Tutorial", "Personal Blog", 
    "Q&A Forum"
]

REASONING_DEPTH = [
    "Intermediate Reasoning", "Advanced Reasoning", "Exceptional Reasoning"
]

TECH_CORRECTNESS = ["Highly Correct"]

# Free Decimal Correspondence: 005.1 = Programming, 005.3 = Systems Programming
FDC_KEEP = ["005.1", "005.3"]   

# D: Essential Common Crawl (23.6B documents)
# R: Taxonomy Code (42.7M documents)
R = {
    d for d in D
    if prefix(d.fdc.primary) in FDC_KEEP
    and d.doc_type_v1.primary in DOC_TYPE_V1
    and d.doc_type_v2.primary in DOC_TYPE_V2
    and d.reasoning_depth.primary in REASONING_DEPTH
    and d.technical_correctness.primary in TECH_CORRECTNESS
}
\end{lstlisting}

\begin{lstlisting}[float=h!,basicstyle=\small\ttfamily, label={lst:taxonomy-code-dclm-filter}, caption={Filter used to curate \textbf{Taxonomy Code w/ DCLM} dataset}]
DOC_TYPE_V2 = [
    "Personal Blog", "Knowledge Article", "Comment Section", "Documentation", "Tutorial", "Q&A Forum"
]

REASONING_DEPTH = [
    "Basic Reasoning", "Intermediate Reasoning", "Advanced Reasoning", "Exceptional Reasoning"
]

# threshold used to filter DCLM Pool -> DCLM-baseline
DCLM_baseline_thresh = 0.01811

# Free Decimal Correspondence: 004 = Computer Science, 005 = Software Development, 
#                              51 = Mathematics
FDC_KEEP = ["004", "005", "51"]   

# D: Essential Common Crawl (23.6B documents)
# R: Taxonomy Code w/ DCLM (274M documents)
R = {
    d for d in D
    if prefix(d.fdc.primary) in FDC_KEEP
    and d.doc_type_v2.primary in DOC_TYPE_V2
    and d.reasoning_depth.primary in REASONING_DEPTH
    and d.quality_signals.rps_doc_ml_eli5_score > DCLM_baseline_thresh
}
\end{lstlisting}


\begin{lstlisting}[float,basicstyle=\scriptsize\ttfamily,language=Python,
  caption={Filter filter used to curate \textbf{Taxonomy Medical} },
  label={lst:taxonomy-medical-filter}]
DOC_TYPE_V1 = [
    "Academic/Research", "Reference/Encyclopedic/Educational"
]

DOC_TYPE_V2 = [
    "Academic Writing", "Documentation", "Knowledge Article", "Q&A Forum"
]

REASONING_DEPTH = [
    "Basic Reasoning", "Intermediate Reasoning", "Advanced Reasoning", "Exceptional Reasoning"
]

TECH_CORRECTNESS = ["Highly Correct", "Exceptionally Correct"]

# Hard-science Free Decimal Correspondence Prefixes
SCIENCE_CODES = ["50", "51", "54", "57", "58", "59", "61"]

# Free Decimal Correspondence: 61 = Medicine
FDC_KEEP = ["61"]

# Document-type blacklists
DOC_TYPE_V1_BLACKLIST = [
    "News/Editorial", "Code/Software", "Social/Forum", "Promotional/Advertisement", "Adult/Pornographic", 
    "Personal/Misc", "Machine-Generated", "E-Commerce/Marketplace", "Images/Videos/Audio"
]

DOC_TYPE_V2_BLACKLIST = [
    "About (Org.)", "About (Personal)", "Audio Transcript", "Comment Section", "Content Listing", "Creative Writing", 
    "Legal Notices", "Listicle", "News (Org.)", "News Article", "Personal Blog", "Product Page", "Spam / Ads", 
    "Structured Data", "Truncated", "Tutorial", "User Review"
]

# D: Essential Common Crawl (23.6B documents)
# S: Taxonomy Medical (235M documents)
S = {
    d for d in D
    # FDC pairing: Medicine (61) w/ another science code
    if (
        (prefix(d.dds.primary) in FDC_KEEP and prefix(d.dds.secondary) in SCIENCE_CODES) or
        (prefix(d.dds.secondary) in FDC_KEEP and prefix(d.dds.primary) in SCIENCE_CODES)
    )
    # Document-type whitelist & blacklist checks 
    and (
        d.doc_type_v1.primary in DOC_TYPE_V1 or
        d.doc_type_v2.primary in DOC_TYPE_V2
    )
    and d.doc_type_v1.primary not in DOC_TYPE_V1_BLACKLIST
    and d.doc_type_v2.primary not in DOC_TYPE_V2_BLACKLIST
    # Quality 
    and d.reasoning_depth.primary in REASONING_DEPTH
    and d.technical_correctness.primary in TECH_CORRECTNESS
}
\end{lstlisting}

\begin{lstlisting}[float,label={lst:taxonomy-med-w-dclmfilter}, caption={Filter to curate the \textbf{Taxonomy Medical w/ DCLM}. Refer to Algorithm~\ref{lst:taxonomy-medical-filter} for \textbf{Taxonomy Medical}.}]
# threshold used to filter DCLM Pool -> DCLM-baseline
DCLM_baseline_thresh = 0.01811

# D: Taxonomy Medical (235M documents)
# R: Taxonomy Medical w/ DCLM (81.2M documents)
R = {
    d for d in D
    if d.quality_signals.rps_doc_ml_eli5_score > DCLM_baseline_thresh
}
\end{lstlisting}

\begin{lstlisting}[float,basicstyle=\scriptsize\ttfamily, label={lst:taxonomy-stem-filter}, caption={Filter used to curate the \textbf{Taxonomy STEM} corpus}]
# Document-types
CODE_DOC_TYPE_V1 = ["Academic/Research", "Reference/Encyclopedic/Educational", "Code/Software", "Social/Forum"]
CODE_DOC_TYPE_V2 = ["Academic Writing", "Comment Section", "Documentation", "Knowledge Article", "Personal Blog", 
                    "Q&A Forum", "Tutorial"]

MED_DOC_TYPE_V1  = ["Academic/Research", "Reference/Encyclopedic/Educational", "Code/Software", "Legal/Regulatory"]
MED_DOC_TYPE_V2  = ["Academic Writing", "Documentation", "FAQ", "Knowledge Article", "News Article", "Tutorial"]

ENG_DOC_TYPE_V1  = ["Academic/Research", "Reference/Encyclopedic/Educational", "Personal/Misc", "Legal/Regulatory"]
ENG_DOC_TYPE_V2  = ["Academic Writing", "Audio Transcript", "Documentation", "FAQ", "Knowledge Article", "News Article", 
                    "Tutorial"]

DEF_DOC_TYPE_V1  = ["Academic/Research", "Reference/Encyclopedic/Educational"]
DEF_DOC_TYPE_V2  = ["Academic Writing", "Knowledge Article", "News Article"]

# Reasoning depth 
REASONING_DEPTH_BAD = ["Abstain", "Indeterminate"]

# Free Decimal Correspondence prefixes 
SCIENCE_FDC = ["50","51","52","53","54","55","56","57","58","59"]
TECH_FDC    = ["60","61","62","66","00"]   # 00 = Computer science
VALID_FDC   = SCIENCE_FDC + TECH_FDC
PROG_FDC    = ["005.1","005.4"]            # programming & systems prog.

# D: Essential Common Crawl (23.6 B docs)
# R: Taxonomy-STEM (1B docs)
R = {
    d for d in D
    if prefix(d.fdc.primary) in VALID_FDC
       and prefix(d.fdc.secondary) in VALID_FDC
       and d.reasoning_depth.primary not in REASONING_DEPTH_BAD
       and d.reasoning_depth.primary is not None
       and (
           # Code
           ((prefix(d.fdc.primary,5) in PROG_FDC or prefix(d.fdc.secondary,5) in PROG_FDC)
            and d.doc_type_v1.primary in CODE_DOC_TYPE_V1
            and d.doc_type_v2.primary in CODE_DOC_TYPE_V2)
        or # Medical
           ((prefix(d.fdc.primary)=="61" or prefix(d.fdc.secondary)=="61")
            and d.doc_type_v1.primary in MED_DOC_TYPE_V1
            and d.doc_type_v2.primary in MED_DOC_TYPE_V2)
        or # Engineering
           ((prefix(d.fdc.primary)=="62" or prefix(d.fdc.secondary)=="62")
            and d.doc_type_v1.primary in ENG_DOC_TYPE_V1
            and d.doc_type_v2.primary in ENG_DOC_TYPE_V2)
        or # Default science/tech
           (d.doc_type_v1.primary in DEF_DOC_TYPE_V1
            and d.doc_type_v2.primary in DEF_DOC_TYPE_V2)
    )
}
\end{lstlisting}

\begin{lstlisting}[float,label={lst:taxonomy-stem-w-dclmfilter}, caption={Filter to curate the \textbf{Taxonomy STEM w/ DCLM}. Refer to Algorithm~\ref{lst:taxonomy-stem-filter} for \textbf{Taxonomy STEM}.}]
# threshold used to filter DCLM Pool -> DCLM-baseline
DCLM_baseline_thresh = 0.01811

# D: Taxonomy STEM (1B documents)
# R: Taxonomy STEM w/ DCLM (342M documents)
R = {
    d for d in D
    if d.quality_signals.rps_doc_ml_eli5_score > DCLM_baseline_thresh
}
\end{lstlisting}

\FloatBarrier
\subsection{Agreement Metric Discussion}

\subsubsection{Motivating Our Variant of Cohen's Kappa} \label{sec: motivatekappa}

Cohen’s \(\kappa\) is a popular choice for measuring inter-annotator agreement across a number of fields. Its robustness against annotators guessing with an imbalanced label distribution has been both criticized \citep{feinstein1990high} and lauded \citep{kolesnyk2022justification}. For our part, we are attempting to measure agreement across multiple categories, and there are several points in our analysis where it is helpful to aggregate scores. While not fool-proof, we believe we are on firmer ground doing so using a kappa metric, than raw accuracy, since a kappa's normalization term can make comparing categories with different numbers of labels easier.

Standard Cohen’s \(\kappa\) (and discrete agreement metrics in general) is most appropriate when the decision boundaries between class labels are clear. In cases where there is potential for fuzzy agreement (for example "assign a quality score from 1 to 10”), other agreement metrics like Kendall’s W are preferred. For our tasks, however, we believe there is fuzzyness in some of our class labels that cannot be solved by sorting the label space. For example, a document can match multiple document types (eg. a document can be both a "Tutorial" and "FAQ"), or its subject-matter can cross multiple FDC labels.

To this end we developed a variant of Cohen’s \(\kappa\) that works with overlapping agreement. We allow our annotators to output up to two labels for each category. Overlaps between annotation label sets $x$ and $y$ can either be scored as 0 or 1
\[
s(x,y) =1_{(x\cap y\neq\varnothing)\vee(x\cup y = \varnothing)}
\]
or weighted by proportion overlap

\[
s(x, y) = \begin{cases}
1 & \text{if } x \cap y = \varnothing \\
\frac{|x\cap y|}{|x\cup y|} & \text{otherwise}
\end{cases}
\]

with appropriate $P_e$ for each.

\subsubsection{On wild swings of \texorpdfstring{$\kappa$}{kappa}.} \label{sec: kappaswings}
The $\kappa$ function is normally described as
\[
\kappa = \frac{P_o - P_e}{1 - P_e} \quad 0 \leq P_o \leq 1 , \quad 0 \leq P_e < 1, \quad -1 \leq \kappa  \leq 1
\]
but its not necessary for $P_o$ and $P_e$ to take on values over their whole range for $\kappa$ to do so. In fact, for any $\epsilon > 0$
\[
\quad 1-\epsilon \leq P_o \leq 1 , \quad 1-\epsilon \leq P_e < 1 \implies -1 \leq \kappa  \leq 1
\]
This implies that as $P_o,P_e \rightarrow 1$, $\kappa$ must become increasingly sensitive to small changes in its inputs.

In Table~\ref{tab:kappa-finetuned-consolidated}, there are big swings of $\kappa$ in a few categories, the largest occurring to the Extraction Artifacts category. If we include $P_e$ and $P_o$ when looking at $\kappa$ values (Table~\ref{tab:kappa-components}), we can see the effect of small changes to $\kappa$'s inputs. While the  Extraction Artifacts category's $\kappa$ falls almost 50pp, the $P_o$'s are within a margin of 15pp, and the $P_e$'s are within 4pp.  In the Education Level category on the random evaluation set, $P_o$ and $\kappa$ disagree on whether \texttt{Qwen2.5-32b-Instruct} or \EMODEL \ performs better.

\begin{table}[htbp]
\centering
\begin{tabular}{lcccccc}
\toprule
\multirow{2}{*}{\textbf{Category}} & \multicolumn{3}{c}{\textbf{Qwen2.5-32B-Inst.}} & \multicolumn{3}{c}{\EMODELBOLD}\\
 & $P_o$ & $P_e$ & $\kappa$ & $P_o$ & $P_e$ & $\kappa$  \\
\midrule
Extraction Artifacts (Random) & 0.93 & 0.75 & 0.74 & 0.79 & 0.71 & 0.27 \\
Reasoning Depth (Random)      & 0.90 & 0.69 & 0.67 & 0.98 & 0.89 & 0.87 \\
Technical Correctness (STEM)  & 0.77 & 0.55 & 0.51 & 0.94 & 0.78 & 0.75 \\
Education Level (Random)    & 0.96 & 0.72 & 0.88 & 0.98 & 0.91 & 0.79 \\
\bottomrule
\end{tabular}
\vspace{0.5\baselineskip}
\caption{$P_o$ and $P_e$ components help to explain whether a rise or drop  in $\kappa$ is real, or an artifact of the metric. In the first row, $\kappa$ exaggerates $P_o$. In the second two, $P_o$ and $\kappa$ are roughly in line with relative improvement. The last row illustrates an ordering disagreement between $P_o$ and $\kappa$}
\label{tab:kappa-components}
\end{table}

There are at least two sources for simultaneously large $P_o$ and $P_e$:
\begin{itemize}
  \item Unbalanced distribution of labels, as discussed in \cite{feinstein1990high}.
  \item An agreement score that is too relaxed. To that end, we created a version of our $\kappa$ that weights agreement via degree of overlap (Appendix~\ref{estimatePe}). This version reduced ordering discrepancies and big swings in $\kappa$, but since the unweighted version of our $\kappa$ did not affect any decisions after averaging across categories, we do not report this version of our $\kappa$ scores.
\end{itemize}

\subsubsection{Motivating the Use of LLMs as Reference Annotators}\label{sec:human-vs-llm}

While assessing the quality of the different versions of our labeler, we've compared agreement with labeling carried out by a set of reference LLMs. A fair question would be whether we are missing something by not evaluating against humans. To that end, we enlisted six people to annotate 1150 documents with four of our taxonomy categories: FDC, Document Type V2, Reasoning Depth, and Education Level. Three annotators (A, B, and C  in Table \ref{table:humans}) annotated 575 documents, and the other three annotators (D, E, and F) labeled the remaining 575 documents. We then compared the human annotators against each other, and against two strong LLMs, using average Cohen's $\kappa$ across the labeled categories. For every human annotator, we found that the annotator had higher average $\kappa$ with the LLMs than they did with other human annotators.

\begin{table}[t]
\small
\begin{tabular}{lcccccc|cc}
\toprule
\textbf{Annotators}&\textbf{A}&\textbf{B}&\textbf{C}&\textbf{D}&\textbf{E}&\textbf{F}&\textbf{Claude Sonnet 3.5}&\textbf{DeepSeek-V3}\\
\midrule
\textbf{A}                 &      & 0.41 & 0.50 &      &      &      & 0.55 & 0.59       \\
\textbf{B}                 & 0.41 &      & 0.54 &      &      &      & 0.68 & 0.71       \\
\textbf{C}                 & 0.50 & 0.54 &      &      &      &      & 0.72 & 0.74       \\
\textbf{D}                 &      &      &      &      & 0.38 & 0.42 & 0.48 & 0.56       \\
\textbf{E}                 &      &      &      & 0.38 &      & 0.51 & 0.65 & 0.64       \\
\textbf{F}                 &      &      &      & 0.42 & 0.51 &      & 0.66 & 0.70       \\
\midrule
\textbf{Claude Sonnet 3.5} & 0.55 & 0.68 & 0.72 & 0.48 & 0.65 & 0.66 &      & 0.81       \\
\textbf{DeepSeek-V3}       & 0.59 & 0.71 & 0.74 & 0.56 & 0.64 & 0.70 & 0.81 &            \\
\bottomrule
\end{tabular}
\vspace{0.5\baselineskip}
\caption{Cohen $\kappa$ between human (A, B, C, D, E, and F) and LLM annotators, averaged across four taxonomy categories. All human annotators have higher agreement with LLM annotators than they do with other human annotators}
\label{table:humans}
\end{table}
\FloatBarrier
\subsubsection{Deriving \texorpdfstring{\(P_e\)}{Pe} Estimator for Annotator \texorpdfstring{$\kappa$}{kappa} \label{estimatePe}}
For annotators \(A_n, n=1,2\) let \(f_{n,k} , k= 0,1, 2\) be the probability \(A_n\) generates \(k\) labels for a document.
Assuming \(k > 0\),  let \(w_{n,x}\)  be the probability of \(A_n\) first labeling the document with label \(x\). Assume the second label is picked with the same \(w_n\), without replacement, so that if \(k = 2\), then \(\frac{w_{n,y}}{1 - w_{n,x}}\) is the probability \(A_n\) picks label \(y\) given that \(x\) was picked first.

Denote \(p_{j,k}\) the probability of an agreement between \(A_1\) and \(A_2\)'s annotations, given \(A_1\)'s annotation is length \(j\) and \(A_2\)'s annotation is length \(k\), as discussed in \ref{sec:kappa}. Then
\[P_e = f_{1,0}f_{2,0} + 
\sum_{j=1}^2\sum_{k=1}^2f_{1,j}f_{2,k}p_{j,k}\]
So we need to calculate \(p_{j,k}\) for \(j = 1,2\), \(k = 1,2\)
\paragraph{\texorpdfstring{$j=1$, $k = 1$}{j = 1, k = 1}}
This is just the same calculation for Cohen's \(\kappa\)

\begin{align*} 
p_{1,1} &= \sum _x \sum_y 1_{x=y} w_{1,x}w_{2,y}s([x],[y])\\
&= s_{1,1}\sum_x w_{1,x}w_{2,x}
\end{align*}
$s([x], [y])$ is the weighted or unweighted matching score, and in both the weighted and unweighted case, $s_{1,1} = 1$ (in subsequent calculations, $s_{i,j}$ is always 1 in the unweighted case)

\paragraph{\texorpdfstring{$j = 1$,  $k=2$}{j = 1, k = 2}\label{sec: p12}}
Begin calculating all possible matches of \([x]\) and \([y, z]\)
\[p_{1,2} = \sum_x \sum_y \sum_z 
1_{y \ne z \wedge (x = y \vee x = z)} w_{1,x}w_{2,y} \frac{w_{2,z}}{1-w_{2,y}}s([x],[y,z])\]
we can break the calculation into two conditions: \(x=y\) and \(x = z\). The conditions are disjoint, since both being true would imply \(y = z\)
\begin{align*}
p_{1,2} &= s_{1,2}\sum_x\sum_z 1_{x\ne z} w_{1,x} w_{2,x}
\frac{w_{2,z}}{1-w_{2,x}} \\
&+ s_{1,2}\sum_x \sum_y 1_{y\ne x}w_{1,x}w_{2,y}
\frac{w_{2,x}}{1-w_{2,y}} \\
p_{1,2} &= s_{1,2}\sum_x w_{1,x}w_{2,x} \sum_z  
1_{x\ne z}\frac{w_{2,z}}{1-w_{2,x}} \\
&+ s_{1,2}\sum_x w_{1,x}w_{2,x}\sum_y 
1_{y\ne x}\frac{w_{2,y}}{1-w_{2,y}} \\
\end{align*}
In the weighted case, $s_{1,2} = 1/2$ . If we denote 
\begin{equation}r_{n,y} = \sum_x 1_{x \ne y}\frac{w_{n,x}}{1 - w_{n,x}}\label{rny}\end{equation}
And note the identity
\begin{equation}
\\\sum_x 1_{y \ne x}  \frac{w_{n,x}}{1 - w_{n,y}} = 1
\label{sum1}
\end{equation}
then
\begin{align}
p_{1,2} = s_{1,2}\sum_x w_{1,x}w_{2,x}(1 + r_{2,x}) \label{eq:p12}
\end{align}

\paragraph{\texorpdfstring{$j=2$, $k=1$}{j = 2, k = 1}}
The same calculation leading to equation \ref{eq:p12} implies
\[p_{2,1} = s_{2,1}\sum_x w_{1,x}w_{2,x}(1 + r_{1,x})\]
where $s_{2,1} = s_{1,2}$
\paragraph{\texorpdfstring{$j=2$, $k=2$}{j = 2, k = 2}}
Begin by calculating all possible matches of \([x,y]\) and \([z,v]\).
\[p_{2,2} = \sum_x \sum_y \sum_z \sum_v 1_{x \neq y \wedge z \neq v \wedge(x=z \vee x=v \vee y=z \vee y=v)}  \frac{w_{1,x}w_{1,y}}{1 - w_{1,x}}  \frac{w_{2,z}w_{2,v}}{1 - w_{2,z}}s([x,y],[z,v])\]
We can calculate the four scenarios \(x=z\), \(x=v\), \(y=z\), and \(y=v\) separately, but unlike case $j=1$,$k=2$, there are intersections we will end up double-counting, which will require subtractions (or weight adjustment in the weighted case). Thinking about the implications of intersections of scenarios in \(x \neq y \wedge z \neq v\) leads to determining that the only non-empty intersections are \(x=z \wedge y=w\) and \(x=w \wedge y=z\). So, \(p_{2,2}\) can be rewritten:
\begin{align*} 
p_{2,2} &= s_{2,2}\sum_x \sum_y \sum_v 1_{x \ne y \wedge x \ne v}\frac{w_{1,x}w_{1,y}}{1-w_{1,x}}\frac{w_{2,x}w_{2,v}}{1-w_{2,x}} &\quad( x=z)\\
&+ s_{2,2}\sum_x \sum_y \sum_z 1_{x \ne y \wedge x \ne z}\frac{w_{1,x}w_{1,y}}{1 - w_{1,x}}  \frac{w_{2,z}w_{2,x}}{1 - w_{2,z}}  &\quad (x=v)\\
&+s_{2,2}\sum_x \sum_y \sum_v 1_{x \ne y \wedge y \ne v}\frac{w_{1,x}w_{1,y}}{1 - w_{1,x}}  \frac{w_{2,y}w_{2,v}}{1 - w_{2,y}}  &\quad(y=z)\\
&+s_{2,2}\sum_x \sum_y \sum_z 1_{x \ne y \wedge z \ne y}\frac{w_{1,x}w_{1,y}}{1 - w_{1,x}}  \frac{w_{2,z}w_{2,y}}{1 - w_{2,z}}  &\quad(y=v)\\
&+a\sum_x \sum_y 1_{x \ne y} \frac{w_{1,x}w_{1,y}}{1-w_{1,x}}\frac{w_{2,x}w_{2,y}}{1-w_{2,x}}&\quad(x=z\wedge y=v)\\
 &+ a\sum_x \sum_y 1_{x \ne y} \frac{w_{1,x}w_{1,y}}{1-w_{1,x}}\frac{w_{2,y}w_{2,x}}{1-w_{2,y}} &\quad(x=v\wedge y=z)
\end{align*}
Where $s_{2,2} = 1/3$ in the weighted case, and $a$ is an over-count adjustment. $a=-1$ in the unweighted case, since we have counted matches of the form $[x,y],[x,y]$ and $[x,y],[y,x]$ twice.   $a = +1/3$ in the weighted case, because the true weighted score for such matches is $1$, not $1/3$, and they were counted twice. After a little rearrangement on the four scenarios, and then substituting in equations \ref{rny} and \ref{sum1}
\begin{align*} 
p_{2,2} &=s_{2,2}\sum_x w_{1,x} w_{2,x} \sum_y 1_{x \ne y} \frac{w_{1,y}}{1-w_{1,x}}\sum_v 1_{x \ne v}\frac{w_{2,v}}{1-w_{2,x}} \\
&+ s_{2,2}\sum_x w_{1,x} w_{2,x} \sum_y 1_{x \ne y}\frac{w_{1,y}}{1 - w_{1,x}}\sum_z 1_{x \ne z}\frac{w_{2,z}}{1 - w_{2,z}}  \\
&+s_{2,2}\sum_y w_{1,y}w_{2,y}\sum_x 1_{x \ne y}\frac{w_{1,x}}{1 - w_{1,x}}\sum_v 1_{y \ne v}  \frac{w_{2,v}}{1 - w_{2,y}}  \\
&+s_{2,2}\sum_y w_{1,y}w_{2,y}\sum_x 1_{x \ne y}\frac{w_{1,x}}{1 - w_{1,x}}\sum_z 1_{y \ne z} \frac{w_{2,z}}{1 - w_{2,z}}  \\
&+a\sum_x \sum_y 1_{x \ne y} \frac{w_{1,x}w_{1,y}}{1-w_{1,x}}\frac{w_{2,x}w_{2,y}}{1-w_{2,x}}  \\
&+a\sum_x \sum_y 1_{x \ne y} \frac{w_{1,x}w_{1,y}}{1-w_{1,x}}\frac{w_{2,y}w_{2,x}}{1-w_{2,y}} \\
p_{2,2} &=s_{2,2}\sum_x w_{1,x} w_{2,x}(1 + r_{2,x} + r_{1,x} + r_{1,x}r_{2,x})  \\
&+a\sum_x \sum_y 1_{x \ne y} \frac{w_{1,x}w_{1,y}}{1-w_{1,x}}\frac{w_{2,x}w_{2,y}}{1-w_{2,x}} \\
&+a \sum_x \sum_y 1_{x \ne y} \frac{w_{1,x}w_{1,y}}{1-w_{1,x}}\frac{w_{2,y}w_{2,x}}{1-w_{2,y}}
\end{align*}
\FloatBarrier


\FloatBarrier
\subsection{Evaluation Sets}
\label{sec:evaluation-sets}

\subsubsection{Domain-recall Gold URL Sets}
\label{sec:domain-recall-gold-urls}

For the human-vetted "gold" URL sets, we manually inspect and collect a set of high quality base URLs for the math and web code domains. Our goal is to ensure that these vetted URLs have a very high density of documents from their respective domain. For example, when visiting \href{https://dlmf.nist.gov/}{https://dlmf.nist.gov/} it is clear that the vast majority of the website should be designated as a mathematics. The math URLs can be found at Algorithm~\ref{lst:vetted-math-urls} and the code can be found at Algorithm~\ref{lst:vetted-code-urls}.

\begin{lstlisting}[float,basicstyle=\scriptsize\ttfamily, language=Python, caption={Vetted Math Base URLS}, label={lst:vetted-math-urls}]
vetted_math_urls =  ["https://dlmf.nist.gov/", "https://tutorial.math.lamar.edu/", "https://math24.net", "https://whitman.edu/mathematics/calculus_online", "https://sfu.ca/math-coursenotes", "https://www2.clarku.edu/faculty/djoyce", "https://clarku.edu/faculty/djoyce/trig", "https://math.libretexts.org", "https://sophisticatedprimate.com", "https://mathisfun.com", "https://chilimath.com", "https://encyclopediaofmath.org/wiki", "https://betterexplained.com/articles", "https://settheory.net/", "https://math24.net", "https://www.opentextbookstore.com/buscalc/buscalc/", "https://oeis.org/", "https://yutsumura.com", "https://math.mit.edu/~djk/calculus_beginners", "https://bookdown.org/slcodia/Stat_136", "https://bookdown.org/huckley/Physical_Processes_In_Ecosystems/", "https://stacks.math.columbia.edu/tag", "https://golem.ph.utexas.edu/", "https://johncarlosbaez.wordpress.com/", "https://qchu.wordpress.com/", "https://cornellmath.wordpress.com/", "https://ncatlab.org", "https://proofwiki.org/wiki/", "https://mathoverflow.net/questions", "https://math.stackexchange.com/questions", "https://projecteuler.net", "https://aperiodical.com/", "https://unapologetic.wordpress.com/", "https://www.jeremykun.com/", "https://terrytao.wordpress.com/", "https://11011110.github.io/blog/", "https://planetmath.org/", "https://alexsisto.wordpress.com/", "https://cstheory.stackexchange.com/questions", "https://web.ma.utexas.edu/mediawiki", "http://theoremoftheweek.wordpress.com/", "http://tqft.net/mlp"]
\end{lstlisting}

\begin{lstlisting}[float,basicstyle=\scriptsize\ttfamily, language=Python, caption={Vetted Web Code Base URLS}, label={lst:vetted-code-urls}]
vetted = ["github.com", "www.geeksforgeeks.org", "stackoverflow.com/questions", "www.experts-exchange.com/questions", "mail.python.org/pipermail", "www.linuxquestions.org/questions", "www.tomshardware.com/forum", "www.coderanch.com/t", "superuser.com/questions", "community.spiceworks.com/topic", "sourceforge.net/p", "sourceforge.net/directory", "docs.microsoft.com/en-us", "serverfault.com/questions", "askubuntu.com/questions", "www.bleepingcomputer.com/forums", "msdn.microsoft.com/en-us", "coderanch.com/t", "learn.microsoft.com/en-us", "unix.stackexchange.com/questions", "forums.unrealengine.com/t", "community.filemaker.com/thread", "www.geekstogo.com/forum", "forum.arduino.cc/t", "www.daniweb.com/programming", "sharepoint.stackexchange.com/questions", "www.construct.net/en/forum", "apple.stackexchange.com/questions", "forums.developer.nvidia.com/t", "techcommunity.microsoft.com/t5"]
\end{lstlisting}

\subsubsection{Sampling STEM Evaluation Set}
\label{sec:stem-eval-set-breakdown}

The STEM evaluation set was created by up-sampling STEM categories as labeled by \texttt{Qwen2.5-32b-Instruct}. A breakdown of the Free Decimal Classification counts of the STEM evaluation set can be found in Table~\ref{tab:qwen32b-fdc1-counts}.

\begin{table}[ht]
\centering
\begin{tabular}{crr}
\toprule
\textbf{Class} & \textbf{FDC top-level category} & \textbf{Documents} \\
\midrule
0 & Information sciences &  54 \\
1 & Philosophy \& psychology                               & 156 \\
3 & Social sciences                                        & 130 \\
4 & Languages \& linguistics                               &  46 \\
5 & Science \& natural history                             & 237 \\
6 & Technology \& applied sciences                         & 232 \\
7 & Arts \& recreation                                     &   9 \\
8 & Literature                                             &   4 \\
9 & History \& geography                                   &   3 \\
\midrule
\textbf{Total} &                                            & \textbf{871} \\
\bottomrule
\end{tabular}
\vspace{0.5\baselineskip}
\caption{Document counts by  Free Decimal Classification level 1 of the STEM evaluation set as labeled by \texttt{Qwen2.5-32b-Instruct}.}
\label{tab:qwen32b-fdc1-counts}
\end{table}

\subsubsection{fastText Classifier Details}
\label{sec:fasttext-appendix}
The fastText classifier for math and web code were both trained on 500k positive examples and 500k negative examples. The negative samples were documents randomly sampled from Common Crawl \citep{shao2024deepseekmathpushinglimitsmathematical}. The hyper-parameters used to train the classifiers can be found in Table~\ref{tab:fasttext-hyperparam}. The positive sets used to train the classifiers we use in the domain-recall experiments are:
\begin{itemize}
    \item \textbf{Math}: sample from OpenWebMath \citep{paster2023openwebmathopendatasethighquality}
    \item \textbf{Web Code}: documentation from package manager platforms (such as \texttt{npm}, \texttt{PyPI}, etc.)
\end{itemize}

\begin{table}[ht]
\centering
\begin{tabular}{ll}
\toprule
\textbf{Hyperparameter} & \textbf{Value} \\
\midrule
Tokenizer & whitespace \\
Word n-grams & 4 \\
Embedding dimension & 128 \\
Hash bucket size & 1,000,000 \\
Min count & 10 \\
Epochs & 2 \\
\bottomrule
\end{tabular}
\vspace{0.5\baselineskip}
\caption{Hyperparameters use to train math and web code fastText classifiers.}
\label{tab:fasttext-hyperparam}
\end{table}

The fastText classifiers did not utilize BPE tokenization or iterative training, which have both been shown to improve performance on domains such as math and code \citep{shao2024deepseekmathpushinglimitsmathematical, huang2025opencoderopencookbooktoptier}.
\FloatBarrier
\subsection{Detailed Taxonomy Evaluation Results}
\label{sec:detailed-taxonomy-results}

\subsubsection{Annotator \(\kappa\) including \texttt{Qwen2.5-14b-Instruct}}
\label{sec:pairwise-cohen-w-qwen14b}

See Table~\ref{tab:kappa-consolidated-4model} for annotator \(\kappa\) of all open-source models tested.

\begin{table}[htbp]
\centering
\footnotesize	
\begin{tabular}{lcccccccc}
\toprule
\multirow{2}{*}{\textbf{Category}} & \multicolumn{2}{c}{\textbf{DeepSeek-V3}} & \multicolumn{2}{c}{\textbf{Qwen 2.5-72B-Inst.}} & \multicolumn{2}{c}{\textbf{Qwen 2.5-32B-Inst.}} & \multicolumn{2}{c}{\textbf{Qwen 2.5-14B-Inst.}}\\
 & Random & STEM & Random & STEM & Random & STEM & Random & STEM\\ \midrule
Knowledge Domain     & 0.69{\tiny$\pm$\,0.02} & 0.64{\tiny$\pm$\,0.03} & 0.46{\tiny$\pm$\,0.02} & 0.39{\tiny$\pm$\,0.03} & 0.62{\tiny$\pm$\,0.02} & 0.68{\tiny$\pm$\,0.02} & 0.20{\tiny$\pm$\,0.01} & 0.24{\tiny$\pm$\,0.02} \\
Cognitive Process    & 0.76{\tiny$\pm$\,0.01} & 0.76{\tiny$\pm$\,0.02} & 0.67{\tiny$\pm$\,0.02} & 0.70{\tiny$\pm$\,0.02} & 0.73{\tiny$\pm$\,0.01} & 0.79{\tiny$\pm$\,0.02} & 0.30{\tiny$\pm$\,0.01} & 0.40{\tiny$\pm$\,0.02} \\
Document Type V1     & 0.90{\tiny$\pm$\,0.01} & 0.91{\tiny$\pm$\,0.01} & 0.88{\tiny$\pm$\,0.01} & 0.91{\tiny$\pm$\,0.01} & 0.86{\tiny$\pm$\,0.01} & 0.89{\tiny$\pm$\,0.01} & 0.60{\tiny$\pm$\,0.01} & 0.62{\tiny$\pm$\,0.02} \\
FDC (level~1)        & 0.92{\tiny$\pm$\,0.01} & 0.95{\tiny$\pm$\,0.01} & 0.90{\tiny$\pm$\,0.01} & 0.92{\tiny$\pm$\,0.01} & 0.88{\tiny$\pm$\,0.01} & 0.92{\tiny$\pm$\,0.01} & 0.88{\tiny$\pm$\,0.01} & 0.91{\tiny$\pm$\,0.01} \\
FDC (level~2)        & 0.86{\tiny$\pm$\,0.01} & 0.87{\tiny$\pm$\,0.01} & 0.83{\tiny$\pm$\,0.01} & 0.81{\tiny$\pm$\,0.01} & 0.81{\tiny$\pm$\,0.01} & 0.81{\tiny$\pm$\,0.01} & 0.78{\tiny$\pm$\,0.01} & 0.76{\tiny$\pm$\,0.01} \\
FDC (level~3)        & 0.71{\tiny$\pm$\,0.01} & 0.70{\tiny$\pm$\,0.01} & 0.67{\tiny$\pm$\,0.01} & 0.63{\tiny$\pm$\,0.01} & 0.64{\tiny$\pm$\,0.01} & 0.60{\tiny$\pm$\,0.01} & 0.57{\tiny$\pm$\,0.01} & 0.55{\tiny$\pm$\,0.01} \\
Extraction Artifacts & 0.81{\tiny$\pm$\,0.02} & 0.86{\tiny$\pm$\,0.03} & 0.57{\tiny$\pm$\,0.02} & 0.53{\tiny$\pm$\,0.03} & 0.74{\tiny$\pm$\,0.01} & 0.65{\tiny$\pm$\,0.03} & 0.27{\tiny$\pm$\,0.02} & 0.30{\tiny$\pm$\,0.02} \\
Missing Content      & 0.83{\tiny$\pm$\,0.01} & 0.85{\tiny$\pm$\,0.02} & 0.63{\tiny$\pm$\,0.01} & 0.65{\tiny$\pm$\,0.02} & 0.66{\tiny$\pm$\,0.02} & 0.65{\tiny$\pm$\,0.02} & 0.43{\tiny$\pm$\,0.01} & 0.45{\tiny$\pm$\,0.02} \\
Document Type V2     & 0.89{\tiny$\pm$\,0.01} & 0.89{\tiny$\pm$\,0.01} & 0.80{\tiny$\pm$\,0.01} & 0.80{\tiny$\pm$\,0.01} & 0.85{\tiny$\pm$\,0.01} & 0.83{\tiny$\pm$\,0.01} & 0.71{\tiny$\pm$\,0.01} & 0.69{\tiny$\pm$\,0.02} \\
Education Level      & 0.89{\tiny$\pm$\,0.01} & 0.86{\tiny$\pm$\,0.02} & 0.82{\tiny$\pm$\,0.01} & 0.81{\tiny$\pm$\,0.02} & 0.88{\tiny$\pm$\,0.01} & 0.85{\tiny$\pm$\,0.02} & 0.69{\tiny$\pm$\,0.02} & 0.71{\tiny$\pm$\,0.02} \\
Reasoning Depth      & 0.75{\tiny$\pm$\,0.01} & 0.72{\tiny$\pm$\,0.02} & 0.70{\tiny$\pm$\,0.01} & 0.67{\tiny$\pm$\,0.02} & 0.67{\tiny$\pm$\,0.02} & 0.67{\tiny$\pm$\,0.02} & 0.47{\tiny$\pm$\,0.01} & 0.50{\tiny$\pm$\,0.02} \\
Technical Corr.      & 0.60{\tiny$\pm$\,0.02} & 0.61{\tiny$\pm$\,0.02} & 0.52{\tiny$\pm$\,0.01} & 0.60{\tiny$\pm$\,0.02} & 0.52{\tiny$\pm$\,0.01} & 0.51{\tiny$\pm$\,0.02} & 0.30{\tiny$\pm$\,0.01} & 0.25{\tiny$\pm$\,0.02} \\ \midrule
\textbf{Overall mean} & 0.80 & 0.80 & 0.70 & 0.70 & 0.74 & 0.74 & 0.52 & 0.53\\ \bottomrule
\end{tabular}
\vspace{0.5\baselineskip}
\caption{Annotator \(\kappa\) (\(\pm\) s.e.) between each candidate model and the two gold annotators on the random (\(n = 2{,}017\)) and STEM (\(n = 871\)) evaluation sets.}
\label{tab:kappa-consolidated-4model}
\end{table}

\subsubsection{Inter-Category NMI}

\begin{figure}[ht!]
  \centering
  \begin{subfigure}[b]{0.45\textwidth}
    \centering
    \includegraphics[width=\linewidth]{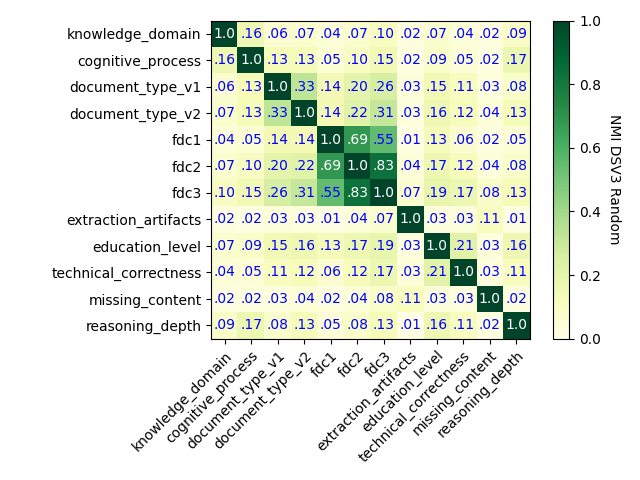}
    \caption{DeepSeek-V3 random NMI}
    \label{fig:dsv3-nmi-random}
  \end{subfigure}
  \hfill
  \begin{subfigure}[b]{0.45\textwidth}
    \centering
    \includegraphics[width=\linewidth]{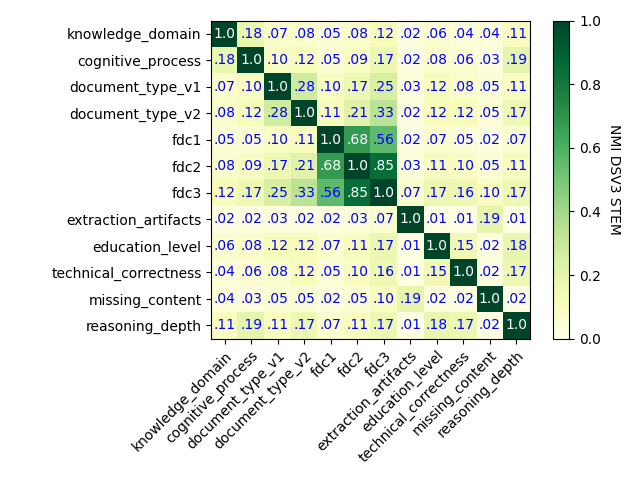}
    \caption{DeepSeek-V3 STEM NMI}
    \label{fig:dsv3-nmi-stem}
  \end{subfigure}
  \centering
  \begin{subfigure}[b]{0.45\textwidth}
    \centering
    \includegraphics[width=\linewidth]{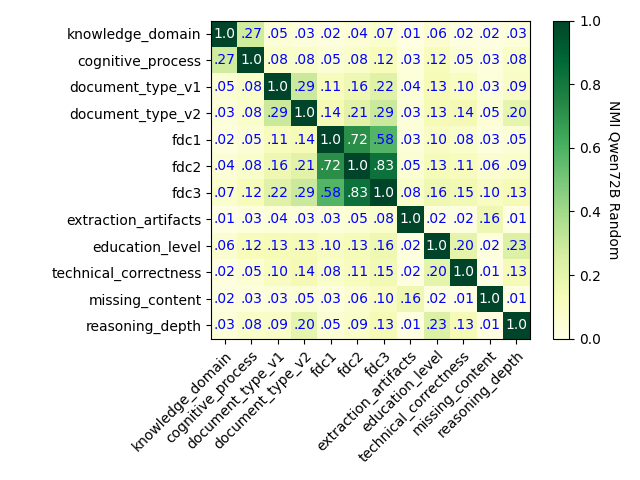}
    \caption{Qwen2.5-72b-Instructrandom NMI}
    \label{fig:qwen72b-nmi-random}
  \end{subfigure}
  \hfill
  \begin{subfigure}[b]{0.45\textwidth}
    \centering
    \includegraphics[width=\linewidth]{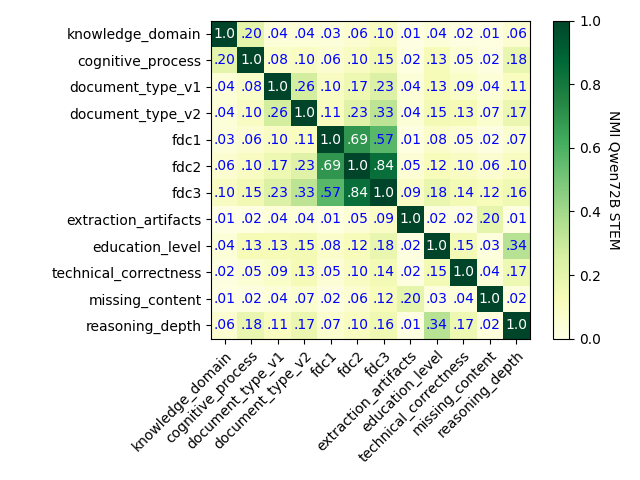}
    \caption{Qwen2.5-72b-Instruct STEM NMI}
    \label{fig:qwen72b-nmi-stem}
  \end{subfigure}
  \caption{NMI heatmaps on random and STEM evaluation sets for \texttt{DeepSeek-V3} and \texttt{Qwen2.5-72b-Instruct}}
  
  \label{fig:os-nmi-heatmaps}
\end{figure}

\paragraph{Additional NMI Heatmaps}
\label{sec:os-nmi-heatmaps}
Additional inter-category NMI heatmaps can be found in Figure~\ref{fig:os-nmi-heatmaps}.

\FloatBarrier
\subsection{\EMODEL \ Training Details \& Ablations}
\label{sec:qwen500-ft-appendix-details}

\subsubsection{Comparison of prefill and generation tokens of \texttt{Qwen2.5-32b-Instruct} vs \EMODEL}
\label{sec:qwen32b-vs-qwen0p5b}

Given we annotated the 104.6M sample of Common Crawl in two passes as explained in Apprendix~\ref{sec:qwen32b-annotation-appendix}, we decide to use the Prompt 1 (Appendix~\ref{qwen32b-prompt-1}) when calculating performance deltas from \texttt{Qwen2.5-32b-Instruct} to \EMODEL. This is a lower bound on performance impact given Prompt 1 just annotates a document for 8 categories (not all 12). However, \EMODEL \ annotates a document with all 12 categories. 

\paragraph{Average Input \& Output Tokens.} 
We sample $1{,}024{,}83$ documents annotated by both \texttt{Qwen2.5-32b-Instruct} and \EMODEL. We then calculate and report the average number of tokens in the prompt and generated output. We ignore the cached prefix of \EMODEL \ given it is only \(11\) tokens before the document is provided (these tokens come from applying the chat template and the short system prompt). The average prompt and generated tokens can be found in Table~\ref{tab:token-stats}.

\begin{table}[htbp]
\centering
\begin{tabular}{lccccc}
\toprule
\textbf{Prompt} & \textbf{Output} & \textbf{Shared prefix} & \textbf{Avg.\ prompt tok.} & \textbf{Avg.\ gen.\ tok.} \\
\midrule
Prompt 1   & Prompt 1 Output & 2104 & 3037 & 791 \\
No Prompt  & Final Output    & 0 &  934 &  51 \\
\bottomrule
\end{tabular}
\vspace{0.5\baselineskip}
\caption{Average token statistics across $1{,}024{,}836$ documents. Prompt 1 Output is produced by \texttt{Qwen2.5-32B-Instruct}; Final Output by \EMODEL.}
\label{tab:token-stats}
\end{table}

\subsubsection{Training Hyperparameters}
\label{sec:qwen500-ft-final-hparams}

The final hyperparameters used to fine-tuned \texttt{Qwen2.5-0.5b-Instruct} can be found in Table~\ref{tab:final-qwen500m-hparams}. 

\begin{table}[htbp]
\centering
\begin{tabular}{lcc}
\toprule
\textbf{Hyperparameter} & \textbf{Value} & \textbf{Notes} \\
\midrule
Optimizer type                    & AdamW &  \\
$\beta_{1}, \beta_{2}$            & 0.9, 0.95   &  \\
Weight decay                      & 0.1   & \\
Global batch size                 & 2M tokens & \\
Peak learning rate                & $1\times10^{-4}$ &  \\
LR warm-up                        & 2B tokens & linear to peak LR \\
Cosine decay phase                & $1\times10^{-4}\!\rightarrow\!1\times10^{-5}$ & follows warm-up \\
Linear anneal phase               & $1\times10^{-5}\rightarrow 0$  & final 2B tokens \\
Total fine-tuning tokens          & 82 B   & synthetic labels \\
Sequence Length                   & 16,384  &  \\
\bottomrule
\end{tabular}
\vspace{0.5\baselineskip}
\caption{Hyper-parameters used to fine-tune \texttt{Qwen2.5-0.5b-Instruct}.}
\label{tab:final-qwen500m-hparams}
\end{table}

\subsubsection{\EMODEL \ Fine-Tuning Ablations}
\label{sec:qwen500-ft-all-ablations}

\paragraph{Fine-tuning prompt / generation format ablation.}

We ablate the affects on annotator \(\kappa\) of context distillation and shortening generation tokens when fine-tuning \texttt{Qwen2.5-0.5b-Instruct}. We compare training on Prompt 1 (Appendix~\ref{qwen32b-prompt-1}) with its standard generation format against no prompt and a highly condensed output format (Algorithm~\ref{lst:condensed-output-template}).\footnote{We used an abridged version of the condensed output format that only outputs the 8 categories in Prompt 1.} Both models are trained for 200B tokens, have a learning rate of \(2\times10^{-6}\), and a max sequence length of 8192. Other than that all hyper-parameters match Table~\ref{tab:final-qwen500m-hparams}. Unfortunately, one run crashed at around 80B tokens seen, so we take the latest shared checkpoint of both models for comparison.\footnote{The model with no prompt, condensed generation format saw more examples because more examples can be packed into a given sequence.} We see that annotator \(\kappa\) of the two variants are very similar (Table~\ref{tab:kappa-io-ablation}).

\begin{table}[htbp]
\centering
\begin{tabular}{lcccc}
\toprule
\multirow{2}{*}{\textbf{Category}} &
\multicolumn{2}{c}{\textbf{Prompt 1 Input \& Format}} &
\multicolumn{2}{c}{\textbf{No Prompt, Condensed Generation}} \\
 & Random & STEM & Random & STEM \\ \midrule
Knowledge Domain                & 0.47{\scriptsize$\pm$\,0.03} & 0.52{\scriptsize$\pm$\,0.03} & 0.54{\scriptsize$\pm$\,0.02} & 0.58{\scriptsize$\pm$\,0.03}\\
Cognitive Process               & 0.56{\scriptsize$\pm$\,0.02} & 0.62{\scriptsize$\pm$\,0.03} & 0.68{\scriptsize$\pm$\,0.02} & 0.74{\scriptsize$\pm$\,0.02}\\
Document Type~V1                & 0.82{\scriptsize$\pm$\,0.01} & 0.84{\scriptsize$\pm$\,0.01} & 0.82{\scriptsize$\pm$\,0.01} & 0.83{\scriptsize$\pm$\,0.01}\\
Free Decimal Corr.\ (level~1)   & 0.83{\scriptsize$\pm$\,0.01} & 0.86{\scriptsize$\pm$\,0.01} & 0.86{\scriptsize$\pm$\,0.01} & 0.91{\scriptsize$\pm$\,0.01}\\
Free Decimal Corr.\ (level~2)   & 0.74{\scriptsize$\pm$\,0.01} & 0.65{\scriptsize$\pm$\,0.01} & 0.79{\scriptsize$\pm$\,0.01} & 0.78{\scriptsize$\pm$\,0.01}\\
Free Decimal Corr.\ (level~3)   & 0.54{\scriptsize$\pm$\,0.01} & 0.48{\scriptsize$\pm$\,0.02} & 0.62{\scriptsize$\pm$\,0.01} & 0.58{\scriptsize$\pm$\,0.01}\\
Extraction Artifacts            & 0.31{\scriptsize$\pm$\,0.03} & 0.48{\scriptsize$\pm$\,0.04} & 0.20{\scriptsize$\pm$\,0.03} & 0.21{\scriptsize$\pm$\,0.04}\\
Missing Content                 & 0.64{\scriptsize$\pm$\,0.02} & 0.74{\scriptsize$\pm$\,0.02} & 0.50{\scriptsize$\pm$\,0.02} & 0.55{\scriptsize$\pm$\,0.03}\\ \midrule
\textbf{Overall mean}           & 0.62 & 0.65 & 0.63 & 0.65\\
\bottomrule
\end{tabular}
\vspace{0.5\baselineskip}
\caption{Annotator $\kappa$ ($\pm$ standard error) against gold annotators on the random ($n=2{,}017$) and STEM ($n=871$) evaluation sets for prompt / generation format ablation.}
\label{tab:kappa-io-ablation}
\end{table}

\paragraph{Fine-tuning learning rate ablation.} Having selected no prompt and the condensed generation format (Algorithm~\ref{lst:condensed-output-template}), we run ablations to determine the best learning rate to use during training. For these experiments, we fine-tune \texttt{Qwen2.5-0.5b-Instruct} for 12B tokens and test the following learning rates: \(2\times10^{-6},\;1\times10^{-5},\;1\times10^{-4},\;\text{and }1\times10^{-3}\). All other hyper-parameters match Table~\ref{tab:final-qwen500m-hparams}. We can see that performance drops off at a learning rate of \(1\times10^{-3}\). We select \(1\times10^{-4}\) given its the highest learning rate before the drop. The results for this ablation can be found in Table~\ref{tab:kappa-lr-ablation}.

\begin{table}[htbp]
\centering
\small
\setlength{\tabcolsep}{4.5pt}
\begin{tabular}{lcccccccc}
\toprule
\multirow{2}{*}{\textbf{Category}} &
\multicolumn{2}{c}{$2\times10^{-6}$} &
\multicolumn{2}{c}{$1\times10^{-5}$} &
\multicolumn{2}{c}{$1\times10^{-4}$} &
\multicolumn{2}{c}{$1\times10^{-3}$} \\[-1pt]
 & Rand. & STEM & Rand. & STEM & Rand. & STEM & Rand. & STEM \\ \midrule
Knowledge Domain                & 0.63{\tiny$\pm$0.02} & 0.61{\tiny$\pm$0.03} & 0.63{\tiny$\pm$0.02} & 0.65{\tiny$\pm$0.03} & 0.64{\tiny$\pm$0.02} & 0.64{\tiny$\pm$0.03} & 0.58{\tiny$\pm$0.02} & 0.61{\tiny$\pm$0.03}\\
Cognitive Process               & 0.68{\tiny$\pm$0.02} & 0.76{\tiny$\pm$0.02} & 0.69{\tiny$\pm$0.02} & 0.76{\tiny$\pm$0.02} & 0.69{\tiny$\pm$0.02} & 0.75{\tiny$\pm$0.02} & 0.67{\tiny$\pm$0.02} & 0.73{\tiny$\pm$0.02}\\
Document Type~V1                & 0.82{\tiny$\pm$0.01} & 0.84{\tiny$\pm$0.01} & 0.84{\tiny$\pm$0.01} & 0.86{\tiny$\pm$0.01} & 0.85{\tiny$\pm$0.01} & 0.86{\tiny$\pm$0.01} & 0.83{\tiny$\pm$0.01} & 0.85{\tiny$\pm$0.01}\\
Free Decimal Corr.\ (level~1)   & 0.87{\tiny$\pm$0.01} & 0.89{\tiny$\pm$0.01} & 0.88{\tiny$\pm$0.01} & 0.92{\tiny$\pm$0.01} & 0.88{\tiny$\pm$0.01} & 0.92{\tiny$\pm$0.01} & 0.86{\tiny$\pm$0.01} & 0.87{\tiny$\pm$0.01}\\
Free Decimal Corr.\ (level~2)   & 0.79{\tiny$\pm$0.01} & 0.70{\tiny$\pm$0.01} & 0.80{\tiny$\pm$0.01} & 0.78{\tiny$\pm$0.01} & 0.81{\tiny$\pm$0.01} & 0.80{\tiny$\pm$0.01} & 0.78{\tiny$\pm$0.01} & 0.73{\tiny$\pm$0.01}\\
Free Decimal Corr.\ (level~3)   & 0.61{\tiny$\pm$0.01} & 0.52{\tiny$\pm$0.02} & 0.64{\tiny$\pm$0.01} & 0.58{\tiny$\pm$0.01} & 0.64{\tiny$\pm$0.01} & 0.62{\tiny$\pm$0.01} & 0.62{\tiny$\pm$0.01} & 0.54{\tiny$\pm$0.01}\\
Extraction Artifacts            & 0.28{\tiny$\pm$0.02} & 0.39{\tiny$\pm$0.04} & 0.26{\tiny$\pm$0.02} & 0.43{\tiny$\pm$0.03} & 0.28{\tiny$\pm$0.02} & 0.42{\tiny$\pm$0.04} & 0.23{\tiny$\pm$0.03} & 0.38{\tiny$\pm$0.03}\\
Missing Content                 & 0.52{\tiny$\pm$0.02} & 0.60{\tiny$\pm$0.02} & 0.50{\tiny$\pm$0.02} & 0.60{\tiny$\pm$0.02} & 0.50{\tiny$\pm$0.01} & 0.60{\tiny$\pm$0.02} & 0.50{\tiny$\pm$0.02} & 0.59{\tiny$\pm$0.03}\\
Document Type~V2                & 0.86{\tiny$\pm$0.01} & 0.86{\tiny$\pm$0.01} & 0.88{\tiny$\pm$0.01} & 0.87{\tiny$\pm$0.01} & 0.88{\tiny$\pm$0.01} & 0.87{\tiny$\pm$0.01} & 0.87{\tiny$\pm$0.01} & 0.83{\tiny$\pm$0.01}\\
Educational Level               & 0.75{\tiny$\pm$0.03} & 0.87{\tiny$\pm$0.02} & 0.77{\tiny$\pm$0.02} & 0.86{\tiny$\pm$0.02} & 0.78{\tiny$\pm$0.03} & 0.85{\tiny$\pm$0.02} & 0.74{\tiny$\pm$0.03} & 0.83{\tiny$\pm$0.02}\\
Reasoning Depth                 & 0.86{\tiny$\pm$0.02} & 0.75{\tiny$\pm$0.02} & 0.86{\tiny$\pm$0.02} & 0.75{\tiny$\pm$0.02} & 0.86{\tiny$\pm$0.02} & 0.75{\tiny$\pm$0.03} & 0.86{\tiny$\pm$0.02} & 0.73{\tiny$\pm$0.03}\\
Technical Correctness           & 0.70{\tiny$\pm$0.02} & 0.74{\tiny$\pm$0.03} & 0.71{\tiny$\pm$0.01} & 0.74{\tiny$\pm$0.03} & 0.72{\tiny$\pm$0.02} & 0.73{\tiny$\pm$0.02} & 0.71{\tiny$\pm$0.01} & 0.70{\tiny$\pm$0.03}\\ \midrule
\textbf{Overall mean}           & 0.70 & 0.71 & 0.71 & 0.73 & 0.71 & 0.74 & 0.69 & 0.70\\
\bottomrule
\end{tabular}
\vspace{0.5\baselineskip}
\caption{Annotator $\kappa$ (mean \(\pm\) standard error) of \texttt{Qwen2.5-0.5b-Instruct} finetuned for 12B tokens with four learning rates. Each pair of columns shows Random ($n{=}2{,}017$) and STEM ($n{=}871$) subsets; headers indicate the learning rate.}
\label{tab:kappa-lr-ablation}
\end{table}

\paragraph{Fine-tuning token budget ablation.} Having selected a learning rate of \(1\times10^{-4}\), we experiment with the token budget used for fine-tuning \texttt{Qwen2.5-0.5b-Instruct}. All hyper-parameters are set to the values in Table~\ref{tab:final-qwen500m-hparams} and we vary the token budget as follows: 12B, 22B, 42B, 82B. Annotator \(\kappa\) does not change as we fine-tune for longer token budgets. Given we trained all 4 models, we select the model trained on the most tokens for \EMODEL. However, it is clear that similar performance can be achieved with a smaller token budget. This has important implications for determining the number of synthetic annotations needed from a powerful LLM to adequately fine-tune a much smaller LM for classification / data labeling tasks like the taxonomic annotation. Results of this ablation can be found in Table~\ref{tab:kappa-token-budget}.

\begin{table}[t]
\centering
\small
\setlength{\tabcolsep}{4.5pt}
\begin{tabular}{lcccccccc}
\toprule
\multirow{2}{*}{\textbf{Category}} &
\multicolumn{2}{c}{\textbf{12 B tokens}} &
\multicolumn{2}{c}{\textbf{22 B tokens}} &
\multicolumn{2}{c}{\textbf{42 B tokens}} &
\multicolumn{2}{c}{\textbf{82 B tokens}} \\[-1pt]
 & Rand. & STEM & Rand. & STEM & Rand. & STEM & Rand. & STEM \\ \midrule
Knowledge Domain             & 0.63{\tiny$\pm$\,0.02} & 0.64{\tiny$\pm$\,0.03} & 0.62{\tiny$\pm$\,0.02} & 0.65{\tiny$\pm$\,0.03} & 0.61{\tiny$\pm$\,0.02} & 0.66{\tiny$\pm$\,0.02} & 0.63{\tiny$\pm$\,0.02} & 0.65{\tiny$\pm$\,0.03} \\
Cognitive Process            & 0.69{\tiny$\pm$\,0.01} & 0.75{\tiny$\pm$\,0.02} & 0.70{\tiny$\pm$\,0.01} & 0.76{\tiny$\pm$\,0.02} & 0.69{\tiny$\pm$\,0.01} & 0.76{\tiny$\pm$\,0.02} & 0.70{\tiny$\pm$\,0.02} & 0.76{\tiny$\pm$\,0.02} \\
Document Type V1             & 0.85{\tiny$\pm$\,0.01} & 0.87{\tiny$\pm$\,0.01} & 0.84{\tiny$\pm$\,0.01} & 0.86{\tiny$\pm$\,0.01} & 0.85{\tiny$\pm$\,0.01} & 0.87{\tiny$\pm$\,0.01} & 0.83{\tiny$\pm$\,0.01} & 0.86{\tiny$\pm$\,0.01} \\
Free Decimal Corr. (level~1) & 0.88{\tiny$\pm$\,0.01} & 0.92{\tiny$\pm$\,0.01} & 0.88{\tiny$\pm$\,0.01} & 0.93{\tiny$\pm$\,0.01} & 0.88{\tiny$\pm$\,0.01} & 0.93{\tiny$\pm$\,0.01} & 0.88{\tiny$\pm$\,0.01} & 0.93{\tiny$\pm$\,0.01} \\
Free Decimal Corr. (level~2) & 0.81{\tiny$\pm$\,0.01} & 0.80{\tiny$\pm$\,0.01} & 0.81{\tiny$\pm$\,0.01} & 0.80{\tiny$\pm$\,0.01} & 0.81{\tiny$\pm$\,0.01} & 0.80{\tiny$\pm$\,0.01} & 0.81{\tiny$\pm$\,0.01} & 0.80{\tiny$\pm$\,0.01} \\
Free Decimal Corr. (level~3) & 0.64{\tiny$\pm$\,0.01} & 0.62{\tiny$\pm$\,0.01} & 0.64{\tiny$\pm$\,0.01} & 0.61{\tiny$\pm$\,0.01} & 0.65{\tiny$\pm$\,0.01} & 0.61{\tiny$\pm$\,0.01} & 0.63{\tiny$\pm$\,0.01} & 0.61{\tiny$\pm$\,0.01} \\
Extraction Artifacts         & 0.28{\tiny$\pm$\,0.02} & 0.42{\tiny$\pm$\,0.03} & 0.28{\tiny$\pm$\,0.02} & 0.39{\tiny$\pm$\,0.03} & 0.28{\tiny$\pm$\,0.02} & 0.39{\tiny$\pm$\,0.03} & 0.27{\tiny$\pm$\,0.02} & 0.37{\tiny$\pm$\,0.03} \\
Missing Content              & 0.50{\tiny$\pm$\,0.01} & 0.60{\tiny$\pm$\,0.02} & 0.50{\tiny$\pm$\,0.01} & 0.59{\tiny$\pm$\,0.02} & 0.50{\tiny$\pm$\,0.01} & 0.58{\tiny$\pm$\,0.02} & 0.48{\tiny$\pm$\,0.01} & 0.57{\tiny$\pm$\,0.02} \\
Document Type V2             & 0.88{\tiny$\pm$\,0.01} & 0.87{\tiny$\pm$\,0.01} & 0.88{\tiny$\pm$\,0.01} & 0.87{\tiny$\pm$\,0.01} & 0.88{\tiny$\pm$\,0.01} & 0.86{\tiny$\pm$\,0.01} & 0.88{\tiny$\pm$\,0.01} & 0.86{\tiny$\pm$\,0.01} \\
Education Level              & 0.78{\tiny$\pm$\,0.02} & 0.85{\tiny$\pm$\,0.02} & 0.78{\tiny$\pm$\,0.02} & 0.85{\tiny$\pm$\,0.02} & 0.80{\tiny$\pm$\,0.03} & 0.86{\tiny$\pm$\,0.02} & 0.79{\tiny$\pm$\,0.02} & 0.86{\tiny$\pm$\,0.02} \\
Reasoning Depth              & 0.86{\tiny$\pm$\,0.02} & 0.74{\tiny$\pm$\,0.03} & 0.87{\tiny$\pm$\,0.01} & 0.76{\tiny$\pm$\,0.03} & 0.86{\tiny$\pm$\,0.01} & 0.76{\tiny$\pm$\,0.02} & 0.87{\tiny$\pm$\,0.01} & 0.76{\tiny$\pm$\,0.02} \\
Technical Correctness        & 0.72{\tiny$\pm$\,0.01} & 0.73{\tiny$\pm$\,0.03} & 0.72{\tiny$\pm$\,0.01} & 0.73{\tiny$\pm$\,0.02} & 0.72{\tiny$\pm$\,0.02} & 0.75{\tiny$\pm$\,0.03} & 0.72{\tiny$\pm$\,0.01} & 0.75{\tiny$\pm$\,0.02} \\ \midrule
\textbf{Overall mean}        & 0.71 & 0.74 & 0.71 & 0.73 & 0.71 & 0.74 & 0.71 & 0.73\\
\bottomrule
\end{tabular}
\vspace{0.5\baselineskip}
\caption{Annotator $\kappa$ (mean {\tiny$\pm$} s.e.) for \texttt{Qwen2.5-0.5b-Instruct} fine-tuned on 12, 22, 42, and 82 billion training tokens. Each pair of columns shows Random ($n{=}2{,}017$) and STEM ($n{=}871$) subsets.}
\label{tab:kappa-token-budget}
\end{table}

\subsection{\texttt{Qwen2.5-32b-Instruct} Large Scale Annotation}
\label{sec:qwen32b-annotation-appendix}

For the purposes of fine-tuning \EMODEL \ and getting a large sample for experimentation, we label 104.6M documents randomly sampled from our processed Common Crawl (Appendix~\ref{sec:cc-processing-pipeline}) with \texttt{Qwen2.5-32b-Instruct}. This annotation was done in two phases given we added additional categories to the taxonomy before fine-tuning. The two prompts used for annotation can be found in Appendix~\ref{qwen32b-prompt-1} and Appendix~\ref{qwen32b-prompt-2}. Both phases used the system prompt found in Appendix~\ref{qwen32b-sys-prompt}.

\subsubsection{DeepSeek-V3 \texttt{SGLang} Image}
\label{sec:dsv3-old-image}

At the time we ran this large scale annotation, the \texttt{SGLang} image for AMD MI300x was much slower. The old image mentioned in Table~\ref{tab:performance-ablation} refers to \texttt{v0.4.1.post3} and the newer image benchmarked refers to \texttt{v0.4.3.post4}.

\subsection{Prompts, Generation Formats, \& Examples}

\subsubsection{Document Sampling}

When running inference with \texttt{Qwen2.5-32b-Instruct} and \EMODEL, we subsample documents with more than 30,000 characters. We do this to improve inference speed. We use the function in Algorithm~\ref{lst:doc-chunking} to take the beginning, random sample of the middle, and end of any document with over 30,000 characters. Assuming there are 4 bytes per token, the maximum document should be about 7500 tokens. We decide not to tokenize when sub-sampling to avoid the CPU-cost of tokenizing all document before inference. The training data for \EMODEL\ is prepared using subsampling to ensure that the training and inference distributions match.

\begin{lstlisting}[float,basicstyle=\footnotesize\ttfamily,caption={Function to subsample documents with more than 30,000 characters}, label={lst:doc-chunking}]
def chunk_text(text, max_char_per_doc):
    if len(text) <= max_char_per_doc:
        return text
        
    chunk_size = max_char_per_doc // 3
    start = text[:chunk_size]
    
    # Calculate valid range for middle section
    middle_start = chunk_size 
    middle_end = len(text) - chunk_size 
    
    # Randomly select middle point within valid range
    mid_point = random.randint(middle_start + chunk_size//2, middle_end - chunk_size//2)
    
    middle = text[mid_point - chunk_size//2:mid_point + chunk_size//2]
    end = text[-chunk_size:]
    return f"[beginning]\n{start}\n[middle]\n{middle}\n[end]\n{end}"

\end{lstlisting}

\subsubsection{\EMODEL \ Generation Templates}
\label{sec:qwen500-ft-gen-template}

During fine-tuning we update the generation to a highly-condensed format. We do so by programatically extracting the response codes from Generations 1 and 2. The generation template for \EMODEL\  can be found in Algorithm~\ref{lst:condensed-output-template}.

\begin{lstlisting}[float,basicstyle=\footnotesize\ttfamily, caption={Template for \EMODEL \ Condensed Model Output Format}, label={lst:condensed-output-template}]
{FDC primary classification},{FDC secondary classification or skip}
{Bloom cognitive process primary (1-6)},{Bloom cognitive process secondary (1-6) or skip}
{Bloom knowledge domain primary (1-4)},{Bloom knowledge domain secondary (1-4) or skip}
{Document type v1 primary (1-17)},{Document type v1 secondary (1-17) or skip}
{Extraction artifacts primary (0-4)},{Extraction artifacts secondary (0-4) or skip}
{Missing content primary (0-6)},{Missing content secondary (0-6) or skip}
{Document type v2 primary (1-25)},{Document type v2 secondary (1-25) or skip}
{Reasoning depth primary (1-6)},{Reasoning depth secondary (1-6) or skip}
{Technical correctness primary (1-6)},{Technical correctness secondary (1-6) or skip}
{Educational level primary (1-5)},{Educational level secondary (1-5) or skip}
\end{lstlisting}

\subsubsection{Prompts}
\label{sec:prompts}

\paragraph{\EMODEL \ System Prompt}\label{qwen500-ft-sys-prompt} \href{https://github.com/Essential-AI/eai-taxonomy/blob/main/src/eai_taxonomy/annotation/prompts/finetuning_system_prompt.txt}{GitHub: \EMODEL \ System Prompt}

\paragraph{\texttt{Qwen2.5-32b-Instruct} System Prompt}\label{qwen32b-sys-prompt} \href{https://github.com/Essential-AI/eai-taxonomy/blob/main/src/eai_taxonomy/annotation/prompts/annotation_system_prompt.txt}{GitHub: System Prompt}

\paragraph{\texttt{Qwen2.5-32b-Instruct} Prompt 1}\label{qwen32b-prompt-1} \href{https://github.com/Essential-AI/eai-taxonomy/blob/main/src/eai_taxonomy/annotation/prompts/annotation_pass_1.txt}{GitHub: Prompt 1}

\paragraph{\texttt{Qwen2.5-32b-Instruct} Prompt 2}\label{qwen32b-prompt-2} \href{https://github.com/Essential-AI/eai-taxonomy/blob/main/src/eai_taxonomy/annotation/prompts/annotation_pass_2.txt}{GitHub: Prompt 2}



\end{document}